\documentclass{article}
\usepackage[final]{neurips_2024}

\usepackage[table]{xcolor}
\usepackage[backref]{hyperref}

\usepackage{basic}
\usepackage{math}
\usepackage{paper}
\disablehyphenation

\usepackage[titles]{tocloft}
\newtheorem{theorem}{Theorem}
\newtheorem{lemma}[theorem]{Lemma}
\newtheorem{proposition}[theorem]{Proposition}

\theoremstyle{definition}
\newtheorem{definition}{Definition}
\newtheorem{example}{Example}

\tikzstyle{shapes}=[draw, line width=2pt, line join=round]
\tikzstyle{shape1}=[shapes, circle, minimum size=.8cm]
\tikzstyle{shape2}=[shapes, regular polygon, regular polygon sides=3, minimum size=.9cm, yshift=-.11cm]
\tikzstyle{shape3}=[shapes, minimum size=.7cm]

\newcommand{\shade}{\cellcolor{black!10}}
\newcommand{\shadeinv}{\cellcolor{black!30}\color{white}}

\newcommand{\cars}{\textbf{3D Cars} \citep{3dcars}}
\newcommand{\dsprites}{\textbf{dSprites} \citep{dsprites}}
\newcommand{\shapes}{\textbf{3D Shapes} \citep{3dshapes}}
\newcommand{\mpi}{\textbf{MPI3D} \citep{mpi3d}}

\title{Enriching Disentanglement:\\ From Logical Definitions to Quantitative Metrics}
\author{%
Yivan Zhang\\
The University of Tokyo, RIKEN AIP\\
Tokyo, Japan\\
\texttt{yivanzhang@ms.k.u-tokyo.ac.jp}\\
\And
Masashi Sugiyama\\
RIKEN AIP, The University of Tokyo\\
Tokyo, Japan\\
\texttt{sugi@k.u-tokyo.ac.jp}\\
}

\begin{document}

\maketitle

\begin{abstract}
Disentangling the explanatory factors in complex data is a promising approach for generalizable and data-efficient representation learning.
While a variety of quantitative metrics for learning and evaluating disentangled representations have been proposed, it remains unclear what properties these metrics truly quantify.
In this work, we establish algebraic relationships between logical definitions and quantitative metrics to derive theoretically grounded disentanglement metrics.
Concretely, we introduce a compositional approach for converting a higher-order predicate into a real-valued quantity by replacing (i) equality with a strict premetric, (ii) the Heyting algebra of binary truth values with a quantale of continuous values, and (iii) quantifiers with aggregators.
The metrics induced by logical definitions have strong theoretical guarantees, and some of them are easily differentiable and can be used as learning objectives directly.
Finally, we empirically demonstrate the effectiveness of the proposed metrics by isolating different aspects of disentangled representations.

\end{abstract}

\section{Introduction}
In \emph{supervised learning}, we usually use a real-valued cost function $\ell: Y \times Y \to \R_{\geq 0}$ to measure how close an output $f(x)$ of a function $f: X \to Y$ is to a target label $y$, i.e., $\ell(f(x), y)$, to quantify the cost of inaccurate prediction.
Then, we can use the \emph{total cost} over a collection of input-output pairs to measure the performance of this function.
From a functional perspective, this construction induces a quantitative metric $L: [X, Y] \times [X, Y] \to \R_{\geq 0}$ between functions:\footnote{$[X, Y]$ denotes the set of all functions from a set $X$ to a set $Y$.}
\begin{equation}
\label{eq:total_cost}
\textstyle
L(f, g)
\defeq
\sum_{x \in X}
\ell(f(x), g(x)),
\end{equation}
where $g: X \to Y$ is a \say{ground-truth function} that maps each input $x$ to its target label $y$.
This metric can be used as both \emph{learning objective} and \emph{evaluation metric} for the learning model $f: X \to Y$.
What does $L(f, g)$ quantify?
It quantifies the extent to which two functions $f$ and $g$ are \emph{equal}:
\begin{equation}
\label{eq:function_equality}
(f =_{[X, Y]} g)
\defeq
\forall x \in X.\;
f(x) =_Y g(x).
\footnotemark
\end{equation}
\footnotetext{In this paper, the domains of equality predicates are explicitly subscripted.}%
Considering the equality as a predicate, we can observe a parallel between
\begin{itemize}
\item (binary-valued) equality $=_Y: Y \times Y \to \truth$ and (real-valued) cost $\ell: Y \times Y \to \R_{\geq 0}$,\footnote{$\truth$ denotes the set of binary truth values: \emph{true} $\lT$ and \emph{false} $\lF$.}
\item universal quantifier (\say{for all}) $\forall x \in X$ and summation $\sum_{x \in X}$, and
\item function equality $=_{[X, Y]}: [X, Y] \times [X, Y] \to \truth$ and total cost $L: [X, Y] \times [X, Y] \to \R_{\geq 0}$.
\end{itemize}

We would like to ask: \emph{Is it possible to measure and optimize other properties in the same way?}

\clearpage


In \emph{representation learning} \citep{bengio2013representation}, measuring and optimizing the performance of a learning model becomes a non-trivial task.
The quality of a model cannot always be measured by how close it is to a fixed ground truth.
Instead, we often need to consider the properties of the model architecture or learned representation itself, such as \emph{convexity} \citep{amos2017input}, \emph{uniformity} \citep{wang2020understanding}, \emph{invariance} \citep{kvinge2022in}, and \emph{equivariance} \citep{lee2019set, brehmer2023geometric}.
A proper comprehension of what constitutes good representations and how to assess their quality is important for designing suitable models, learning objectives, and evaluation metrics.


\paragraph{Disentangled representation learning: definitions, metrics, and methods}

\emph{Disentanglement} is an important property in representation learning, which intuitively means that different explanatory factors in data should be encoded separately \citep{bengio2013representation}.
However, disentanglement has no universally agreed-upon formal definition \citep{higgins2018towards, suter2019robustly, shu2020weakly, fumero2021learning}, and it is typically viewed not as a single property but rather as a combination of several requirements \citep{ridgeway2018learning, eastwood2018framework, do2020theory, tokui2022disentanglement}.
While many metrics for measuring disentanglement have been proposed \citep{carbonneau2022measuring}, it remains unclear what properties these metrics truly quantify and how they can be optimized directly.
Often, a new evaluation metric is introduced along with a new learning method, but it is usually unproven that the method can optimize the new metric \citep{higgins2017betavae, kim2018disentangling, chen2018isolating, li2020progressive}.
This lack of theoretical understanding makes it difficult to design learning models that can effectively learn disentangled representations.


\paragraph{A logical and algebraic approach to defining and measuring disentangled representations}

Recently, \citet{zhang2023category} proposed a general and abstract definition of disentanglement, shedding light on the common structures underlying the algebraic, statistical, and topological definitions of disentanglement.
It was shown that the abstract concept of \emph{product} \citep{maclane1978categories} underlies an essential property of disentanglement called \emph{modularity} \citep{ridgeway2018learning}, and other properties of learning models, such as \emph{informativeness} \citep{eastwood2018framework}, can also be defined abstractly using only the composition and identity of morphisms.
Following this algebraic approach, we aim to derive theoretically grounded quantitative metrics of disentanglement from the logical definitions of the desired properties, extending the parallel between \cref{eq:function_equality,eq:total_cost}.


\paragraph{Contributions}

In this paper, we focus on logically defined properties of disentangled representation learning, such as modularity and informativeness (\cref{sec:definitions}).
We introduce a compositional approach to converting a \emph{higher-order equational predicate} into a \emph{real-valued quantity} (\cref{tab:conversion}), which serves as a quantitative metric of the extent to which a function satisfies the predicate (\cref{sec:enrichment}).
Our analysis on the relationship between the logical definitions and the induced quantitative metrics provides theoretical guarantee on the properties of the optimal functions (\cref{thm:main}).
Then, we demonstrate the usefulness of this conversion method by deriving quantitative metrics for measuring properties of disentangled representations, and we analyze these metrics in terms of computation, optimization, and differentiability (\cref{sec:metrics}).
Lastly, we compare the derived metrics with several existing ones in a fully controlled experiment and demonstrate that the proposed metrics are able to isolate different aspects of disentangled representations (\cref{sec:experiments}).

\section{Logical definitions of disentangled representations}
\label{sec:definitions}
In this section, let us first take a closer look at the logical definitions of two properties of disentangled representation learning --- informativeness \citep{eastwood2018framework} and modularity \citep{ridgeway2018learning}, which are arguably more important than other properties \citep{carbonneau2022measuring}.
We limit our discussion to sets and functions, but the generalization to other morphisms, such as equivariant, stochastic, or continuous functions, is straightforward.


\subsection{Informativeness: injectivity or retractability of a learning model}

Being informative, expressive, faithful, or useful is a basic requirement for learned representations \citep{bengio2013representation}.
We want a representation learning model to preserve explanatory factors in data that are informative to the downstream tasks.
For functions, this criterion could be formulated as follows: If two factors $y$ and $y'$ are different, then their representations $m(y)$ and $m(y')$ extracted by a function $m: Y \to Z$ should be different too.
This means that the function $m$ should be \emph{injective}:
\begin{definition}[Injective function]
\label{def:injective}
A function $m: Y \to Z$ is \emph{injective} if
\begin{equation}
\label{eq:injective}
p_\text{injective}(m: Y \to Z)
\defeq
\forall y \in Y.\;
\forall y' \in Y.\;
(m(y) =_Z m(y')) \limp (y =_Y y').
\end{equation}
\end{definition}
Alternatively, because injective functions are precisely functions with \emph{retractions} (left inverses) \citep[Chapter 2]{lawvere2003sets}, we can measure the \emph{retractability} instead:
\begin{definition}[Retractable function]
\label{def:retractable}
A function $m: Y \to Z$ has a \emph{retraction} $h: Z \to Y$ if
\begin{equation}
\label{eq:retractable}
p_\text{retractable}(m: Y \to Z)
\defeq
\exists h: Z \to Y.\;
h \compL m =_{[Y, Y]} \id_Y.
\end{equation}
\end{definition}

Note that these properties are \emph{predicates} $p_\text{injective}, p_\text{retractable}: [Y, Z] \to \truth$ on the set $[Y, Z]$ of all functions from $Y$ to $Z$.
Analogous to using the total cost in \cref{eq:total_cost} to measure function equality in \cref{eq:function_equality}, if we want to measure the \emph{injectivity} in \cref{eq:injective} or \emph{retractability} in \cref{eq:retractable}, we need to find quantitative counterparts of the \textbf{implication} $\limp$, \textbf{universal quantifier} $\forall$, and \textbf{existential quantifier} $\exists$ used in their logical definitions.
Generally, it is desirable to extend the parallel between \cref{eq:total_cost,eq:function_equality} to other predicates by finding quantitative operations corresponding to logical connectives and quantifiers.
This correspondence allows us to construct and analyze quantitative metrics for machine learning models in a \emph{compositional} manner \citep{boole1854investigation}.


\subsection{Modularity: product structure preserved by a learning model}

\emph{Modularity} \citep{ridgeway2018learning} is an essential property of disentangled representation learning, which means that the explanatory factors in data, such as the color and shape of an object, are separated into independent components in the learned representation \citep{bengio2013representation}.


\begin{wrapfigure}{r}{.6\linewidth}
\centering
\vspace{-1.5em}
\begin{adjustbox}{width=\linewidth}
\begin{tikzpicture}
\node (X) [matrix] {
  \node (x) [draw, shape=ellipse, minimum width=3cm, minimum height=3.5cm, line width=1pt] {};
  \node (x1) [shape1, fill=\red , outer sep=1ex] at (-.2,  1.2) {};
  \node (x2) [shape1, fill=\blue, outer sep=1ex] at ( .3,   .3) {};
  \node (x3) [shape2, fill=\red , outer sep=1ex] at (  0, -1.2) {};
  \node (x4) [shape2, fill=\blue, outer sep=1ex] at (-.3,  -.3) {};
  \\
};
\node (Y) [matrix, left=3cm of X, yshift=.3cm, anchor=center] {
  \node [draw, rectangle, minimum width=3.2cm, minimum height=3.2cm, line width=1pt] {};
  \draw (-1.6, 0) [dashed] to (1.6, 0);
  \draw (0, -1.6) [dashed] to (0, 1.6);
  \node (y1) [text width=1cm, align=center, font=\small] at (-.8,  .8) {red \\ circle};
  \node (y2) [text width=1cm, align=center, font=\small] at ( .8,  .8) {blue\\ circle};
  \node (y3) [text width=1cm, align=center, font=\small] at (-.8, -.8) {red \\ triangle};
  \node (y4) [text width=1cm, align=center, font=\small] at ( .8, -.8) {blue\\ triangle};
  \node at (-.8, 2) {red};
  \node at ( .8, 2) {blue};
  \node [anchor=east] at (-1.6, .8) {circle};
  \node [anchor=east] at (-1.6, -.8) {triangle};
  \\
};
\node (Z) [matrix, right=2.5cm of X, yshift=.3cm, anchor=center] {
  \node [draw, rectangle, minimum width=3.2cm, minimum height=3.2cm, line width=1pt] {};
  \draw (-1.6, 0) [dashed] to (1.6, 0);
  \draw (0, -1.6) [dashed] to (0, 1.6);
  \node (z1) at ( .8,  .8) {$(1, 0)$};
  \node (z2) at (-.8,  .8) {$(0, 0)$};
  \node (z3) at ( .8, -.8) {$(1, 1)$};
  \node (z4) at (-.8, -.8) {$(0, 1)$};
  \node at (-.8, 2) {$0$};
  \node at ( .8, 2) {$1$};
  \node at (2,  .8) {$0$};
  \node at (2, -.8) {$1$};
  \\
};
\draw [|->] (y1.north east) -- (x1);
\draw [|->] (y2.east)  -- (x2);
\draw [|->] (y3.south east) -- (x3);
\draw [|->] (y4.east)  -- (x4);
\draw [|->] (x1) -- (z1.north west);
\draw [|->] (x2) -- (z2.west);
\draw [|->] (x3) -- (z3.south west);
\draw [|->] (x4) -- (z4.west);
\end{tikzpicture}
\end{adjustbox}

\hspace{1.4em}
\begin{tikzcd}[column sep=1em, row sep=.5em]
\text{factors}
&
\text{observations}
&
\text{codes}
\\
Y_\text{color} \times Y_\text{shape}
\arrow[r, "g"]
\arrow[rr, bend right=30, looseness=.5, "m \defeq f \compL g = m_\text{color} \times m_\text{shape}"']
&
X
\arrow[r, "f"]
&
Z_\text{color} \times Z_\text{shape}
\end{tikzcd}
\caption{Disentangled representation learning}
\label{fig:disentangled_representation_learning}
\vspace{-2em}
\end{wrapfigure}

As shown in \cref{fig:disentangled_representation_learning}, modularity can be defined as follows.
We assume that data with multiple explanatory factors (e.g., color and shape) is generated via a function $g: Y \to X$ \uline{from a product} $Y \defeq Y_1 \times Y_2$ of \emph{factors}.
An encoder $f: X \to Z$ is a function \uline{to a product} $Z \defeq Z_1 \times Z_2$ of \emph{codes}.
Then, an encoder is said to be \emph{modular} if it can \textbf{reconstruct the product structure}, such that the composition $m \defeq f \compL g: Y \to Z$ of the generator $g$ and the encoder $f$ is a \uline{product function}.\footnote{For two functions $f: A \to C$ and $g: B \to D$, their \emph{product} $f \times g: A \times B \to C \times D$ applies these two functions \say{in parallel} by mapping a pair $(a, b)$ in $A \times B$ to a pair $(f(a), g(b))$ in $C \times D$.}


Formally, being a product function is also a property that can be represented as a predicate:
\begin{definition}[Product function]
\label{def:product}
Let $Y \defeq Y_1 \times Y_2$ and $Z \defeq Z_1 \times Z_2$ be products of sets.
A function $m: Y \to Z$ is a \emph{product function} if
\begin{equation}
p_\text{product}(m: Y \to Z)
\defeq
\exists m_{1,1}: Y_1 \to Z_1.\;
\exists m_{2,2}: Y_2 \to Z_2.\;
m =_{[Y, Z]} m_{1,1} \times m_{2,2}.
\end{equation}
\end{definition}

\begin{example}
\label{ex:main}
Let us compare the following two functions from $Y \defeq \set{0, 1}^2$ to $Z \defeq \R^2$:\\
\begin{minipage}{.35\linewidth}
\begin{equation}
\label{eq:m}
m
\defeq
\begin{cases}
(0, 0) \mapsto (1, 2)\\
(0, 1) \mapsto (3, 4)\\
(1, 0) \mapsto (5, 6)\\
(1, 1) \mapsto (7, 8)\\
\end{cases}\hspace{-1em}
\end{equation}
\end{minipage}%
\hfill
\begin{minipage}{.65\linewidth}
\begin{equation}
\label{eq:m'}
m'
\defeq
\begin{cases}
(0, 0) \mapsto (a, c)\\
(0, 1) \mapsto (a, d)\\
(1, 0) \mapsto (b, c)\\
(1, 1) \mapsto (b, d)\\
\end{cases}\hspace{-1em}
=
\underbrace{
\begin{cases}
0 \mapsto a\\
1 \mapsto b\\
\end{cases}\hspace{-1em}
}_{m_{1,1}}
\times
\underbrace{
\begin{cases}
0 \mapsto c\\
1 \mapsto d\\
\end{cases}\hspace{-1em}
}_{m_{2,2}}
\end{equation}
\end{minipage}\\
where $a$, $b$, $c$, and $d$ are arbitrary real numbers.
According to \cref{def:product}, only $m' = m_{1,1} \times m_{2,2}$ is a product function, whose first/second output depends only on the first/second input.
\end{example}

In \cref{ex:main}, although $m$ is not a product function, we want to address the following questions:
\begin{itemize}
\item (Metric) Can we quantify the extent to which it resembles a product function?
\item (Approximation) Can we find a product function that is closest to it?
\item (Differentiability) Can we make it slightly closer to a product function?
\end{itemize}

Answers to these questions will be given in the following sections.

\section{Enrichment: from logic to metric}
\label{sec:enrichment}
In \cref{app:preliminaries,app:theory}, we describe in detail the theory of converting a higher-order predicate into a real-valued quantity.
In this section, we only introduce the conversion procedure using concrete examples and present the theoretical results.
A summary of the conversion is given in \cref{tab:conversion}.

\begin{wraptable}{r}{.5\linewidth}
\centering
\vspace{-1em}
\caption{From logic to metric}
\label{tab:conversion}
\begin{adjustbox}{width=\linewidth}
\begin{tabular}{l@{}c@{\hspace{5pt}}l@{}c}
\toprule
\multicolumn{2}{c}{Logic} & \multicolumn{2}{c}{Metric} \\
\midrule
truth values & $\truth$ & real values & $\quant$ \\
equality & $=$ & strict premetric & $d$ \\
conjunction & $\lcon$ & addition & $+$ \\
disjunction & $\ldis$ & minimum & $\min$ \\
implication & $\limp$ & subtraction$^*$ & $\monus$ \\
universal quantifier & $\forall$ & aggregator$^{**}$ & $\qforall$ \\
existential quantifier & $\exists$ & infimum & $\inf$ \\
\bottomrule
\end{tabular}
\end{adjustbox}
\raggedright
\footnotesize{%
$^*$ truncated subtraction: $b \monus a \defeq \max\set{b - a, 0}$
\\
$^{**}$ e.g., maximum, sum, mean, and mean square}
\end{wraptable}

First of all, let us clarify the terms predicate and quantity.
In the realm of classical logic, a \emph{predicate} $p: A \to \truth$ on a set $A$ is a function from the set $A$ to the set $\truth$ of binary truth values.
For example, the predicates $p_\text{injective}$, $p_\text{retractable}$, and $p_\text{product}$ in \cref{def:injective,def:retractable,def:product} are functions from the set $[Y, Z]$ of functions to the set $\truth$.
They are \emph{logical definitions} of some properties of functions.
On the other hand, in this work, a \emph{quantity} $q: A \to \quant$ on a set $A$ is defined as a function to the set $\quant$ of extended non-negative real numbers.
The quantities associated with a predicate will serve as \emph{quantitative metrics} for the property defined by the predicate.


\subsection{From equality predicate to strict premetric}

In this work, a predicate of central importance is the \emph{equality predicate} $=_A: A \times A \to \truth$ \citep{mazur2008one}.
A quantity associated with the equality predicate should be a strict premetric:
\begin{definition}[Strict premetric]
\label{def:strict_premetric}
A \emph{strict premetric} on a set $A$ is a function $d_A: A \times A \to \quant$ that
\begin{equation}
\forall a \in A.\;
\forall a' \in A.\;
(d_A(a, a') = 0) \leqv (a =_A a').
\end{equation}
\end{definition}


\subsection{From logical operation to quantitative operation}

\begin{figure}
\centering
\begin{subfigure}[t]{0.24\linewidth}
\centering
\includegraphics[height=3.6cm]{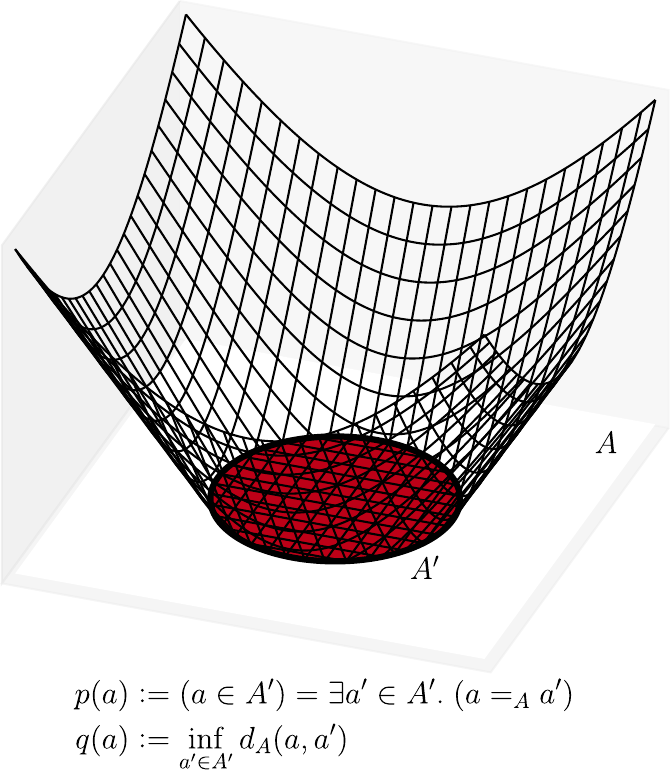}
\caption{predicate and quantity}
\end{subfigure}
\hfill
\begin{subfigure}[t]{0.24\linewidth}
\centering
\includegraphics[height=3.6cm]{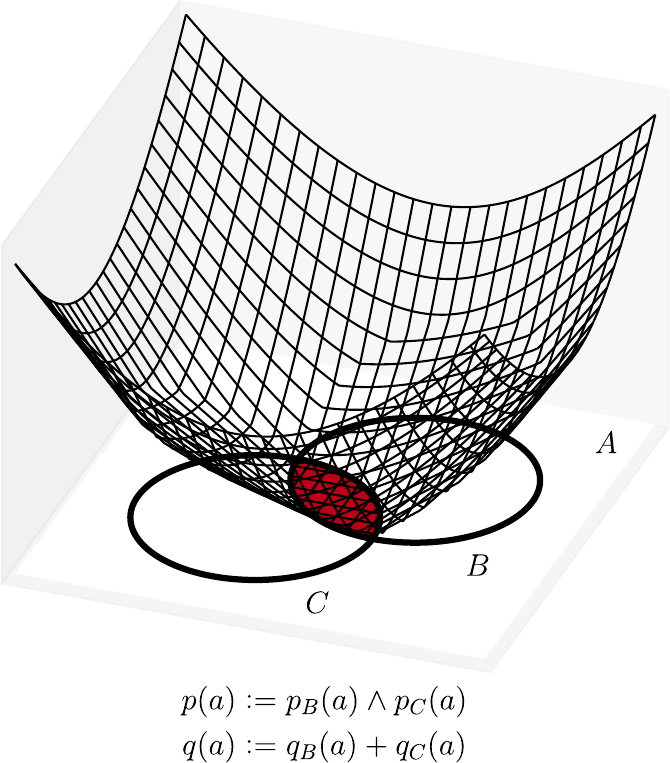}
\caption{conjunction}
\end{subfigure}
\hfill
\begin{subfigure}[t]{0.24\linewidth}
\centering
\includegraphics[height=3.6cm]{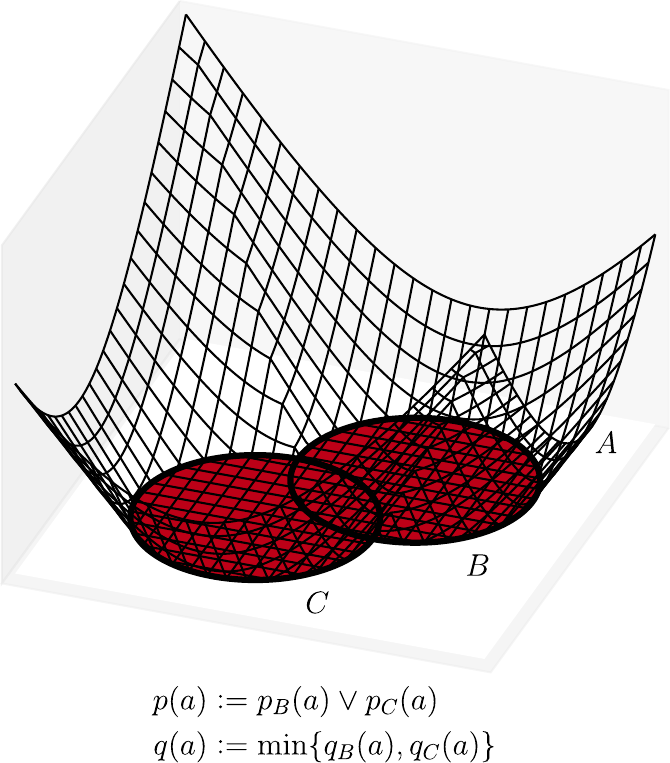}
\caption{disjunction}
\end{subfigure}
\hfill
\begin{subfigure}[t]{0.24\linewidth}
\centering
\includegraphics[height=3.6cm]{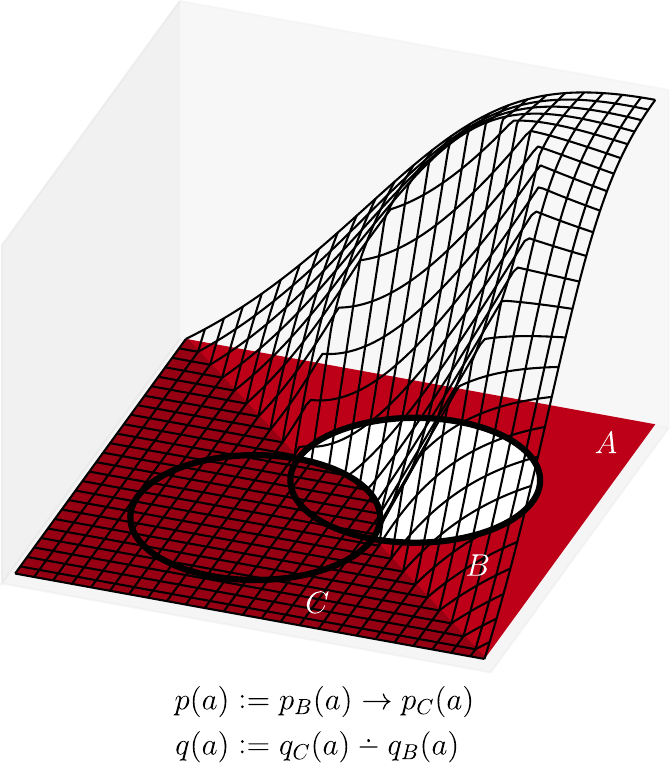}
\caption{implication}
\end{subfigure}
\caption{From predicates and logical operations to quantities and quantitative operations}
\label{fig:venn}
\end{figure}

Next, let us have a look at the logical connectives and quantifiers used in the definitions of properties.
The product of sets and functions plays a significant role in this work.
Two functions $f, g: C \to A \times B$ to a product are equal if and only if all their component functions are equal:
\begin{equation}
\label{eq:pairing_equality}
\hspace{-.5em}
(f =_{[C, A \times B]} g)
\defeq
(f_1 =_{[C, A]} g_1) \lcon (f_2 =_{[C, B]} g_2).
\footnotemark
\end{equation}
\footnotetext{For a function $f: C \to A \times B$ to a product $A \times B$, its \emph{component functions} $f_1: C \to A \defeq p_1 \compL f$ and $f_2: C \to A \defeq p_2 \compL f$ are denoted by numeric subscripts, where $p_1: A \times B \to A \defeq (a, b) \mapsto a$ and $p_2: A \times B \to B \defeq (a, b) \mapsto b$ are \emph{projection functions}.}%
Note that the \textbf{conjunction} $\lcon: \truth \times \truth \to \truth$, a logical connective, is used in \cref{eq:pairing_equality}.
To obtain a corresponding quantity, we replace it with the \emph{addition} $+: \quant \times \quant \to \quant$:
\begin{equation}
\label{eq:pairing_premetric}
d_{[C, A \times B]}(f, g)
\defeq
d_{[C, A]}(f_1, g_1) + d_{[C, B]}(f_2, g_2).
\end{equation}
The \textbf{universal quantifier} on a set $A$ is a specific (second-order) predicate $\forall_A: \truth^A \to \truth$ on the set $\truth^A$ of predicates.
We can replace it with the \emph{supremum} $\sup: \quant^A \to \quant$.
We can also choose a function from the (i) \emph{maximum}, (ii) \emph{sum}, (iii) \emph{mean}, and (iv) \emph{mean square} when the set $A$ is finite.
More generally, we can replace it with a quantity $\qforall_A: \quant^A \to \quant$ on the set $\quant^A$ of quantities that satisfies some conditions, which we refer to as a (universal) \emph{aggregator}.
Intuitively, a universal aggregator should output $0$ if and only if all inputs are $0$.
Therefore, the median, mode, and range are non-examples.
Different choices of aggregators yield metrics with different characteristics in computation and optimization.
For example, the function equality predicate
\begin{equation}
(f =_{[A, B]} g)
\defeq
\forall a \in A.\;
f(a) =_B g(a)
\end{equation}
converts to a quantity whose aggregator $\qforall$ is not limited to the sum (cf.~\cref{eq:function_equality,eq:total_cost}):
\begin{equation}
\label{eq:function_premetric}
d_{[A, B]}(f, g)
\defeq
\qforall_{a \in A}
d_B(f(a), g(a)).
\end{equation}
Dually, we also need to consider the \textbf{disjunction} $\ldis$ and the \textbf{existential quantifier} $\exists$.
We replace them with the \emph{minimum} and the \emph{infimum}, respectively.
Lastly, we replace the \textbf{implication} $a \limp b$ with the (truncated) \emph{subtraction} $b \monus a \defeq \max\set{b - a, 0}$.
These operations are illustrated in \cref{fig:venn}.


\subsection{From compound predicate to compound quantity}

Following \cref{tab:conversion}, we can convert any \emph{compound predicate} defined using equational predicates and logical operations into a corresponding \emph{compound quantity} defined using strict premetrics and quantitative operations.
Our main result on their relationship is as follows:
\begin{theorem}
\label{thm:main}
Let $p: A \to \truth$ be a predicate on a set $A$, and let $q: A \to \quant$ be a quantity converted from $p$ according to \cref{tab:conversion}.
Then, for any $a \in A$, $q(a) = 0$ implies $p(a) = \lT$.
Conversely, for any $a \in A$, $p(a) = \lT$ implies $q(a) = 0$ if and only if $p$ does not contain the implication.
\end{theorem}

The implication is special because we must sacrifice logical equivalence for the sake of continuity, which is necessary for gradient-based optimization.
We will explore this through a concrete example regarding injectivity in \cref{ssec:informativeness} and discuss it in detail in \cref{app:theory,app:discussions}.


\subsection{(Sub)homomorphism from metric to logic}

Finally, for readers interested in the theoretical background, we briefly introduce the following algebraic concepts and a proof sketch underlying \cref{tab:conversion,thm:main}.

\begin{definition}[Zero predicate]
The \emph{zero predicate} $\zeta: \quant \to \truth \defeq x \mapsto (x = 0)$ is a function that maps $0$ to true $\lT$ and any positive value to false $\lF$.
\end{definition}

\begin{definition}[(Sub)homomorphism from a quantity to a predicate]
Let $A$ be a set.
A quantity $q: A \to \quant$ on the set $A$ is \emph{homomorphic} to a predicate $p: A \to \truth$ via the zero predicate $\zeta: \quant \to \truth$ if $\zeta \compL q = p$, and is \emph{subhomomorphic} to $p$ if $\zeta \compL q \limp p$.%
\footnote{We use the infix notation, so $\zeta \compL q = p$ means that $\forall a \in A.\; (q(a) = 0) \leqv p(a)$, and $\zeta \compL q \limp p$ means that $\forall a \in A.\; (q(a) = 0) \limp p(a)$.}
\end{definition}

\begin{definition}[(Sub)homomorphism from a quantitative operation to a logical operation]
Let $n \in \N$ be a natural number.
An $n$-ary quantitative operation $\alpha: \quant^n \to \quant$ is \emph{homomorphic} to a logical operation $\beta: \truth^n \to \truth$ via the zero predicate $\zeta: \quant \to \truth$ if $\zeta \compL \alpha = \beta \compL \zeta^n$, and is \emph{subhomomorphic} to $\beta$ if $\zeta \compL \alpha \limp \beta \compL \zeta^n$.%
\footnote{$\zeta^n: \quant^n \to \truth^n \defeq (q_1, \dots, q_n) \mapsto (q_1 = 0, \dots, q_n = 0)$ is the $n$-fold \emph{product} of the zero predicate $\zeta: \quant \to \truth$.}
\end{definition}

\begin{definition}[(Sub)homomorphism from an aggregator to a quantifier]
Let $A$ be a set.
An aggregator $\alpha_A: \quant^A \to \quant$ on the set $A$ is \emph{homomorphic} to a quantifier $\beta_A: \truth^A \to \truth$ via the zero predicate $\zeta: \quant \to \truth$ if $\zeta \compL \alpha_A = \beta_A \compL \zeta^A$, and is \emph{subhomomorphic} to $\beta_A$ if $\zeta \compL \alpha_A \limp \beta_A \compL \zeta^A$.%
\footnote{$\zeta^A: \quant^A \to \truth^A \defeq \zeta \compL (-)$ is the \emph{postcomposition} with the zero predicate $\zeta: \quant \to \truth$ that maps a quantity $q: A \to \quant$ to the predicate $\zeta \compL q: A \to \truth$.}
\end{definition}

Homomorphic quantities, quantitative operations, and aggregators can be illustrated as follows:
\begin{equation}
\begin{tikzcd}[column sep=2em, row sep=2em]
A
\arrow[d, "q"']
\arrow[r, "\id_A"]
&
A
\arrow[d, "p"]
\\
\quant
\arrow[r, "\zeta"]
&
\truth
\end{tikzcd}
\qquad
\begin{tikzcd}[column sep=2em, row sep=2em]
\quant^n
\arrow[d, "\alpha"']
\arrow[r, "\zeta^n"]
&
\truth^n
\arrow[d, "\beta"]
\\
\quant
\arrow[r, "\zeta"]
&
\truth
\end{tikzcd}
\qquad
\begin{tikzcd}[column sep=2em, row sep=2em]
\quant^A
\arrow[d, "\alpha_A"']
\arrow[r, "\zeta^A"]
&
\truth^A
\arrow[d, "\beta_A"]
\\
\quant
\arrow[r, "\zeta"]
&
\truth
\end{tikzcd}
\end{equation}

Based on these algebraic concepts, we can say that strict premetrics are homomorphic to equality predicates, addition is homomorphic to conjunction (since the sum is zero if and only if both addends are zero), minimum is homomorphic to disjunction, truncated subtraction is \emph{subhomomorphic} to implication, and universal aggregators are homomorphic to the universal quantifier.

\cref{thm:main} means that any compound quantity is (sub)homomorphic to the corresponding compound predicate if each component (quantities, quantitative operations, aggregators) is (sub)homomorphic to the corresponding component (predicates, logical operations, quantifiers).
For implication, we use the truncated subtraction, which is only subhomomorphic, since there is no \emph{continuous} operation that is homomorphic to implication (see also \cref{app:discussions}).

More abstractly and concisely, we can say that we replace the \textbf{Heyting algebra} of truth values $\truth$ with a \emph{quantale} of extended non-negative real numbers $\quant$, and we replace the quantifiers $\forall$ and $\exists$ with aggregators $\qforall$ and $\inf$ (see also \cref{app:theory}).
In this way, we can derive quantitative metrics for any logically defined properties of learning models \emph{compositionally}.

\section{Quantitative metrics of disentangled representations}
\label{sec:metrics}
In this section, we demonstrate how to apply the conversion method introduced above to derive quantitative metrics for measuring the modularity (\cref{def:product}) and informativeness (\cref{def:injective,def:retractable}) of disentangled representations.

In \cref{ssec:product_approximation,ssec:constancy}, we introduce modularity metrics based on two approaches and discuss their differences in terms of computation and optimization.
We point out that the main obstacle lies in the optimization step, resulting from the existential quantifiers in the definition.
Then, we show that we can derive easily computable and differentiable metrics from a logically equivalent definition.
In \cref{ssec:informativeness}, we introduce informativeness metrics and present a result of \cref{thm:main}.


\subsection{Modularity metrics via product approximation}
\label{ssec:product_approximation}

We begin with \emph{modularity}, which is an essential property of disentangled representation learning.
Recall that modularity can be defined using the \emph{product function} (\cref{def:product}).
For easier reference, we provide the following diagram, which shows the domains and codomains of the functions involved in the upcoming discussion:
\begin{equation}
\begin{tikzcd}[column sep=3em, row sep=3em]
\begin{array}{c}
Y_1 \\ y_1
\end{array}
\arrow[d, "m_{1,1}"']
&
\begin{array}{c}
Y_1 \times Y_2 \\ (y_1, y_2)
\end{array}
\arrow[d, "m" description]
\arrow[l, "p_1"']
\arrow[r, "p_2"]
\arrow[ld, "m_1"']
\arrow[rd, "m_2"]
&
\begin{array}{c}
Y_2 \\ y_2
\end{array}
\arrow[d, "m_{2,2}"]
\\
\begin{array}{c}
Z_1 \\ m_1(y_1, y_2) \\ = m_{1,1}(y_1)
\end{array}
&
\begin{array}{c}
Z_1 \times Z_2 \\ m(y_1, y_2) \\ \;
\end{array}
\arrow[l, "p_1"]
\arrow[r, "p_2"']
&
\begin{array}{c}
Z_2 \\ m_2(y_1, y_2) \\ = m_{2,2}(y_2)
\end{array}
\end{tikzcd}
\end{equation}

From \cref{def:product}, we can derive the following metric:

\begin{definition}[Product approximation]
Let $m: Y \to Z$ be a function from a product $Y \defeq Y_1 \times Y_2$ of sets to another product $Z \defeq Z_1 \times Z_2$ of sets.
The extent to which $m$ resembles a \emph{product function} can be measured by a distance between $m$ and its best product function approximation:
\begin{equation}
q_\text{product}(m: Y \to Z)
\defeq
\inf_{m_{1,1} \in [Y_1, Z_1]}
\inf_{m_{2,2} \in [Y_2, Z_2]}
d_{[Y, Z]}(m, m_{1,1} \times m_{2,2}).
\end{equation}
\end{definition}

The derivation of $q_\text{product}$ from $p_\text{product}$ follows the conversion described in \cref{tab:conversion}: replacing the equality $=_{[Y, Z]}: [Y, Z] \times [Y, Z] \to \truth$ with a strict premetric $d_{[Y, Z]}: [Y, Z] \times [Y, Z] \to \quant$ and the existential quantifiers $\exists$ with the infimum operators $\inf$.

This modularity metric can be interpreted as a distance from a point $m \in [Y, Z]$ to a subset $\set{m_{1,1} \times m_{2,2} \given m_{1,1} \in [Y_1, Z_1], m_{2,2} \in [Y_2, Z_2]} \subset [Y, Z]$ of product functions (cf.~the Hausdorff distance \citep{lawvere1986taking, tuzhilin2016invented}).
Following from \cref{thm:main}, $q_\text{product}(m) = 0$ if and only if $p_\text{product}(m) = \lT$.
This means that the minimizers of this metric are precisely product functions.

However, we still face two obstacles: the product operation and the minimization problem.
For the product operation, we can employ \cref{eq:pairing_premetric,eq:function_premetric} to rewrite $q_\text{product}$ into a more computable form:
\begin{proposition}
\label{prop:product}
The quantity $q_\textnormal{product}(m: Y \to Z)$ equals
\begin{equation}
\qforall_{y_1 \in Y_1}
\qforall_{y_2 \in Y_2}
d_{Z_1}(m_1(y_1, y_2), m_{1,1}^*(y_1))
+
\qforall_{y_2 \in Y_2}
\qforall_{y_1 \in Y_1}
d_{Z_2}(m_2(y_1, y_2), m_{2,2}^*(y_2)),
\end{equation}
where the functions $m_{1,1}^*: Y_1 \to Z_1$ and $m_{2,2}^*: Y_2 \to Z_2$ are given by
\begin{align}
\label{eq:m11}
m_{1,1}^*: Y_1 \to Z_1
&\defeq
y_1 \mapsto \arginf_{z_1 \in Z_1} \qforall_{y_2 \in Y_2} d_{Z_1}(m_1(y_1, y_2), z_1),
\\
\label{eq:m22}
m_{2,2}^*: Y_2 \to Z_2
&\defeq
y_2 \mapsto \arginf_{z_2 \in Z_2} \qforall_{y_1 \in Y_1} d_{Z_2}(m_2(y_1, y_2), z_2).
\end{align}
\end{proposition}

A detailed derivation can be found in \cref{app:proofs}.
Note that we can obtain the optimal product function approximation $m_{1,1}^* \times m_{2,2}^*$ explicitly via \cref{eq:m11,eq:m22}.
Intuitively, we need to find an approximation of a (multi)set of codes with one factor fixed and other factors varying, and then we use the aggregation of all the approximation errors as a modularity metric.

The second obstacle --- the minimization problem --- still needs to be addressed.
Since the code spaces $Z_1$ and $Z_2$ can be infinite sets, the minimization problem may not have a closed-form minimizer or even an exact solver.
Even if an exact solver exists, the solution may not be differentiable with respect to the inputs.
Let us examine some concrete examples of $q_\text{product}$ by choosing different aggregators $\qforall$ in \cref{eq:m11,eq:m22}.
In the following three examples, we assume that the code spaces $Z_1$ and $Z_2$ are Euclidean spaces equipped with the usual Euclidean distances.

\begin{example}
If the aggregator $\qforall$ is the \emph{supremum}, the best approximation is the \emph{center} of the smallest bounding sphere \citep{megiddo1983weighted}, and the approximation error is the \emph{radius}.
\end{example}

This metric has the advantage of being definable even when the factor spaces $Y_1$ and $Y_2$ are infinite sets, and it can be computed using either randomized \citep{welzl1991smallest} or exact \citep{fischer2003fast} algorithms.
However, it is not easy to calculate its gradient.
Thus, we cannot use it as a learning objective and directly optimize it using gradient-based optimization.

\begin{example}
If the aggregator $\qforall$ is the \emph{mean}, the best approximation is the \emph{(geometric) median} \citep{weiszfeld1937point}, and the approximation error is the \emph{mean absolute deviation around the median}.
\end{example}

It is known that there is no exact algorithm for obtaining the geometric median \citep{cockayne1969euclidean}, but it can be effectively approximated using convex optimization \citep{cohen2016geometric}.
The geometric median has found applications in robust estimation in the fields of statistics and machine learning \citep{meer1991robust, minsker2015geometric, pillutla2022robust, guerraoui2023byzantine}.

\begin{example}
If the aggregator $\qforall$ is the \emph{mean square}, the best approximation is the \emph{mean}, and the approximation error is the \emph{variance}.
In this case, $q_\text{product}(m)$ can be simplified to
\begin{equation}
\label{eq:variance}
\mean_{y_1 \in Y_1} \var_{y_2 \in Y_2} m_1(y_1, y_2) +
\mean_{y_2 \in Y_2} \var_{y_1 \in Y_1} m_2(y_1, y_2).
\end{equation}
\end{example}

The variance is easier to compute and differentiate than the radius of the smallest bounding sphere and the mean absolute deviation around the median, but it is also more susceptible to outliers and noise.
Further work could explore the theoretical implications of these metrics, especially in cases where only partial combinations of factors or noisy annotations are available.

Then, let us revisit our motivating example in \cref{ex:main}:
\begin{example}
Let us consider the function $m: \set{0, 1}^2 \to \R^2$ in \cref{eq:m}.
Its best product function approximation is
\begin{equation}
m^*: \set{0, 1}^2 \to \R^2
\defeq
\begin{cases}
(0, 0) \mapsto (2, 4)\\
(0, 1) \mapsto (2, 6)\\
(1, 0) \mapsto (6, 4)\\
(1, 1) \mapsto (6, 6)\\
\end{cases}\hspace{-1em}
=
\underbrace{
\begin{cases}
0 \mapsto 2\\
1 \mapsto 6\\
\end{cases}\hspace{-1em}
}_{m_{1,1}^*}
\times
\underbrace{
\begin{cases}
0 \mapsto 4\\
1 \mapsto 6\\
\end{cases}\hspace{-1em}
}_{m_{2,2}^*}
\end{equation}
because $m_{1,1}^*(0) = 2$ is the center/median/mean of the set $\set{m_1(0, 0) = 1, m_1(0, 1) = 3}$, and so on.
The modularity metric is a distance between $m$ and $m^*$.
\end{example}


\subsection{Modularity metrics via constancy}
\label{ssec:constancy}

Upon analyzing the metrics above, it becomes evident that what we need is not the best approximation itself (e.g., the mean) but rather the approximation error (e.g., the variance) --- a measure of the \emph{constancy} of a set of codes.
Following this insight, our next objective is to formulate a modularity metric that eliminates the need for an optimization step.
\citet{zhang2023category} have proved that a function is a product function if and only if the \emph{curried functions}\footnote{For a binary function $f: A \times B \to C$, its \emph{curried function} $\expt{f}: A \to [B, C]$ is a unary function such that for all $a \in A$ and $b \in B$, $f(a, b) = \expt{f}(a)(b)$ \citep{curry1980some}.} of its component functions are constant, as shown in the following example:
\begin{example}
Consider the functions $m, m': \set{0, 1}^2 \to \R^2$ in \cref{eq:m,eq:m'} and the curried functions of their second component functions $m_2, m'_2: \set{0, 1}^2 \to \R$:\\
\begin{minipage}{.49\linewidth}
\begin{equation}
m_2
=
\begin{cases}
(0, 0) \mapsto 2\\
(0, 1) \mapsto 4\\
(1, 0) \mapsto 6\\
(1, 1) \mapsto 8\\
\end{cases}\hspace{-1em}
\iso
\begin{cases}
0 \mapsto
\begin{cases}
0 \mapsto 2\\
1 \mapsto 4\\
\end{cases}\hspace{-1em}
\\
1 \mapsto
\begin{cases}
0 \mapsto 6\\
1 \mapsto 8\\
\end{cases}
\end{cases}\hspace{-2em}
\end{equation}
\end{minipage}
\hfill
\begin{minipage}{.49\linewidth}
\begin{equation}
m'_2
=
\begin{cases}
(0, 0) \mapsto c\\
(0, 1) \mapsto d\\
(1, 0) \mapsto c\\
(1, 1) \mapsto d\\
\end{cases}\hspace{-1em}
\iso
\begin{cases}
0 \mapsto
\begin{cases}
0 \mapsto c\\
1 \mapsto d\\
\end{cases}
\\
1 \mapsto
\begin{cases}
0 \mapsto c\\
1 \mapsto d\\
\end{cases}
\end{cases}\hspace{-2em}
\end{equation}
\end{minipage}\\
The curried function $\expt{m_2}: \set{0, 1} \to [\set{0, 1}, \R]$ is not constant, while $\expt{m'_2}$ is constant with value $\set{0 \mapsto c, 1 \mapsto d} \in [\set{0, 1}, \R]$ (and so is $\expt{m'_1}$), indicating that $m'$ is a product function.
\end{example}

Based on this fact, we propose an alternative approach for measuring modularity:
\begin{definition}[Constancy of curried function]
Let $m_1$ and $m_2$ be the component functions of a function $m: Y \to Z$ from a product $Y \defeq Y_1 \times Y_2$ of sets to another product $Z \defeq Z_1 \times Z_2$ of sets.
The extent to which $m$ resembles a product function can be measured by the \emph{constancy} of the curried functions of $m_1$ and $m_2$:
\begin{equation}
q_\text{const-curry}(m: Y \to Z)
\defeq
q_\text{const}(\expt{m_1}) + q_\text{const}(\expt{m_2}),
\end{equation}
where $q_\text{const}$ is a quantity for constant functions.
\end{definition}

To complete this construction, we adopt the following definition and metric of the constant function:
\begin{definition}[Constant function]
\label{def:const}
A function $f: A \to B$ is \emph{constant} if 
\begin{equation}
p_\text{const}(f: A \to B)
\defeq
\forall a \in A.\;
\forall a' \in A.\;
f(a) =_B f(a'),
\end{equation}
which can be measured by
\begin{equation}
\label{eq:constancy_quantity}
q_\text{const}(f: A \to B)
\defeq
\qforall_{a \in A \vphantom{a'}}
\qforall_{a' \in A}
d_B(f(a), f(a')).
\end{equation}
\end{definition}

This constancy metric $q_\text{const}$ only needs to compute pairwise distances between the outputs, requiring $\abs{A}^2$ times distance computation but no optimization.
Incorporating $q_\text{const}$ into $q_\text{const-curry}$, we can get the following metric:
\begin{proposition}
\label{prop:const_curry}
The quantity $q_\textnormal{const-curry}(m: Y \to Z)$ equals
\begin{equation}
\label{eq:constant_currying_quantity}
\begin{aligned}
&
\qforall_{y_1 \in Y_1 \vphantom{y_2'}}
\qforall_{y_2 \in Y_2 \vphantom{y_2'}}
\qforall_{y_2' \in Y_2}
d_{Z_1}(m_1(y_1, y_2), m_1(y_1, y_2'))
\\
+&
\qforall_{y_2 \in Y_2 \vphantom{y_1'}}
\qforall_{y_1 \in Y_1 \vphantom{y_1'}}
\qforall_{y_1' \in Y_1}
d_{Z_2}(m_2(y_1, y_2), m_2(y_1', y_2)).
\end{aligned}
\end{equation}
\end{proposition}

Here are two examples of $q_\text{const}$ and $q_\text{const-curry}$ using different aggregators $\qforall$ in \cref{eq:constancy_quantity,eq:constant_currying_quantity}.

\begin{example}
If the aggregator $\qforall$ is the \emph{maximum}, $q_\text{const}$ is the \emph{diameter} (the maximum pairwise distance) of the outputs.
In this case, $q_\text{const-curry}(m)$ can be simplified to
\begin{equation}
\label{eq:diameter}
\max_{y_1 \in Y_1}\operatorname*{diam}_{y_2 \in Y_2} m_1(y_1, y_2) +
\max_{y_2 \in Y_2}\operatorname*{diam}_{y_1 \in Y_1} m_2(y_1, y_2).
\end{equation}
\end{example}

\begin{example}
If the aggregator $\qforall$ is the \emph{mean square}, $q_\text{const}$ is the mean pairwise squared distance, which equals the \emph{variance}.
In this case, $q_\text{const-curry}(m)$ coincides with \cref{eq:variance}.
\end{example}

In summary, \cref{eq:variance,eq:diameter} are easily computable and differentiable metrics, and their minimizers are precisely product functions.
They do not contain any hyperparameters or stochastic components and thus can serve as both learning objectives and evaluation metrics.


\subsection{Informativeness metrics}
\label{ssec:informativeness}

If an encoder $f: X \to Z$ is constant, mapping everything to the same value, according to \cref{def:product}, it is perfectly modular.
However, a constant encoder is also completely useless.
In this subsection, we shift our focus to the property of \emph{informativeness} --- a measurement of usefulness.

Informativeness is not a unique requirement for disentangled representations.
Other representation learning paradigms, such as contrastive learning \citep{jaiswal2020survey,wang2020understanding} and metric learning \citep{musgrave2020metric}, also emphasize the importance of mapping dissimilar data to far-apart locations in the representation space.
While one could integrate this requirement into a single disentanglement score (e.g., \citep{higgins2017betavae,kim2018disentangling}), we argue that it is better to evaluate the usefulness of representations separately for a more fine-grained assessment \citep{carbonneau2022measuring}.

One straightforward way to measure informativeness is to measure how much we can invert the encoding process:
\begin{definition}[Retraction approximation]
Let $m: Y \to Z$ be a function.
The extent to which $m$ is \emph{retractable} can be measured by a distance between the composition of $m$ and its best retraction approximation and the identity function:
\begin{equation}
q_\text{retractable}(m: Y \to Z)
\defeq
\inf_{h \in [Z, Y]}
d_{[Y, Y]}(h \compL m, \id_Y)
=
\inf_{h \in [Z, Y]}
\qforall_{y \in Y}
d_Y(h(m(y)), y).
\end{equation}
\end{definition}

This metric $q_\text{retractable}$ is derived from \cref{def:retractable} following the conversion procedure in \cref{tab:conversion}.
This informativeness metric also involves an optimization step similar to the modularity metric $q_\text{product}$, potentially introducing randomness or higher computation costs.
Note that we may use a parameterized subset of the set $[Z, Y]$ of all functions from codes $Z$ to factors $Y$, such as the set of linear functions.
Then, the problem becomes a regression/classification problem, and the metric is the performance of the predictor.
A number of existing works adopted this approach and used the accuracy, the area under the ROC curve (AUC-ROC), or the mean squared error (MSE) to measure the informativeness \citep{ridgeway2018learning, eastwood2018framework, eastwood2023dcies}.
However, such metrics necessitate additional hyperparameter tuning and are more likely to exhibit varying behavior across different implementations \citep{carbonneau2022measuring}.

It raises the question of whether we can measure the informativeness of an encoder without approximating its retraction.
We propose to measure informativeness by directly measuring the injectivity of the encoding process:
\begin{definition}[Contraction]
Let $m: Y \to Z$ be a function.
The extent to which $m$ is \emph{injective} can be measured by how much $m$ contracts pairs of inputs:
\begin{equation}
q_\text{injective}(m: Y \to Z)
\defeq
\qforall_{y \in Y \vphantom{y'}}
\qforall_{y' \in Y}
d_Y(y, y') \monus d_Z(m(y), m(y')).
\end{equation}
\end{definition}

This metric $q_\text{injective}$ is derived from \cref{def:injective} following the conversion procedure in \cref{tab:conversion}.
According to \cref{thm:main}, we know that $q_\text{retractable}(m) = 0$ if and only if $m$ is retractable.
However, $q_\text{injective}(m) = 0$ implies the injectivity of $m$ but not the other way around:
\begin{equation}
\begin{tikzcd}[column sep=4em, row sep=2em]
(p_\text{retractable}(m) = \lT)
\arrow[r, "\substack{\text{logically}\\\text{equivalent}}"', leftrightarrow]
\arrow[d, shift left]
&
(p_\text{injective}(m) = \lT)
\arrow[d, "\mathcolor{\red}{\times}"{anchor=center}, "\text{ \cref{thm:main}}", shift left]
\\
(q_\text{retractable}(m) = 0)
\arrow[u, "\text{\cref{thm:main}}", shift left]
&
(q_\text{injective}(m) = 0)
\arrow[u, shift left]
\end{tikzcd}
\end{equation}
In other words, a minimizer of $q_\text{injective}$ is required to be \emph{non-contractive}, which is a stronger condition than being injective.
For example, let us consider the function $m: [0, 1] \to \R \defeq y \mapsto 0.01 \times y$.
Although it is injective, its outputs are less distinguishable from each other in terms of the Euclidean distance. 
Therefore, $q_\text{injective}$ still assign a non-zero value to this function.

Although not all injective functions necessarily minimize $q_\text{injective}$, according to \cref{thm:main}, we can guarantee that minimizing $q_\text{injective}$ will not lead to non-injective functions.
Moreover, $q_\text{injective}$ does not require training regressors or classifiers to approximate the retraction.
Consequently, it does not need any time-consuming hyperparameter tuning or cross-validation like existing informativeness metrics \citep{eastwood2018framework, ridgeway2018learning}.

\section{Experiments}
\label{sec:experiments}
\begin{table}[t]
\centering
\caption{Supervised disentanglement metrics}
\label{tab:main}
\begin{adjustbox}{width=\linewidth}
\begin{tabular}{l c rrrrr c rrrrr rrrrrr}
\toprule
& \multicolumn{6}{c}{Modularity} & \multicolumn{6}{c}{Informativeness} & \multicolumn{6}{c}{Existing metrics}
\\
\cmidrule(lr){2-7}
\cmidrule(lr){8-13}
\cmidrule(lr){14-19}
& & \multicolumn{3}{c}{Product approx.} & \multicolumn{2}{c}{Constancy}
& & \multicolumn{3}{c}{Retraction approx.} & \multicolumn{2}{c}{Contraction}
& \multicolumn{2}{c}{Pair} & \multicolumn{1}{c}{Info.} & \multicolumn{3}{c}{Regressor}
\\
\cmidrule(lr){3-5}
\cmidrule(lr){6-7}
\cmidrule(lr){9-11}
\cmidrule(lr){12-13}
\cmidrule(lr){14-15}
\cmidrule(lr){16-16}
\cmidrule(lr){17-19}
& & Rad. & MAD & Var. & Diam. & MPD 
& & ME & MAE & MSE & Max & Mean
& Beta$^a$ & Factor$^b$ & MIG$^c$ & Dis.$^d$ & Com.$^d$ & Info.$^d$
\\
\midrule
entanglement & \xmark & $0.44$ & $0.75$ & $0.96$ & $0.19$ & $0.82$ & \cmark & \shadeinv $0.76$ & \shadeinv $0.96$ & \shadeinv $0.99$ & $0.44$ & $0.78$ & $0.89$ & $0.83$ & $0.18$ & $0.28$ & $0.28$ & $1.00$ \\
rotation     & \xmark & $0.22$ & $0.51$ & $0.80$ & $0.05$ & $0.64$ & \cmark & \shade $1.00$ & \shade $1.00$ & \shade $1.00$ & \shade $1.00$ & \shade $1.00$ & $0.96$ & $0.34$ & $0.17$ & $0.40$ & $0.40$ & $1.00$ \\
duplicate    & \xmark & $0.24$ & $0.43$ & $0.67$ & $0.06$ & $0.56$ & \cmark & \shade $1.00$ & \shade $1.00$ & \shade $1.00$ & \shade $1.00$ & \shade $1.00$ & $1.00$ & $1.00$ & $1.00$ & $1.00$ & $0.59$ & $1.00$ \\
complement   & \xmark & $0.12$ & $0.28$ & $0.55$ & $0.01$ & $0.42$ & \cmark & \shade $1.00$ & \shade $1.00$ & \shade $1.00$ & \shade $1.00$ & \shade $1.00$ & $1.00$ & $0.00$ & $1.00$ & $1.00$ & $0.63$ & $1.00$ \\
misalignment & \xmark & $0.22$ & $0.44$ & $0.74$ & $0.05$ & $0.58$ & \cmark & \shade $1.00$ & \shade $1.00$ & \shade $1.00$ & \shade $1.00$ & \shade $1.00$ & $1.00$ & $0.00$ & $1.00$ & $1.00$ & $1.00$ & $1.00$ \\
redundancy   & \cmark & \shade $1.00$ & \shade $1.00$ & \shade $1.00$ & \shade $1.00$ & \shade $1.00$ & \cmark & \shade $1.00$ & \shade $1.00$ & \shade $1.00$ & \shade $1.00$ & \shade $1.00$ & $1.00$ & $0.33$ & $1.00$ & $1.00$ & $0.93$ & $1.00$ \\
contraction  & \cmark & \shade $1.00$ & \shade $1.00$ & \shade $1.00$ & \shade $1.00$ & \shade $1.00$ & \cmark & \shade $1.00$ & \shade $1.00$ & \shade $1.00$ & $0.18$ & $0.49$ & $1.00$ & $1.00$ & $1.00$ & $1.00$ & $1.00$ & $1.00$ \\
nonlinear    & \cmark & \shade $1.00$ & \shade $1.00$ & \shade $1.00$ & \shade $1.00$ & \shade $1.00$ & \cmark & \shadeinv $0.79$ & \shadeinv $0.93$ & \shadeinv $0.99$ & $0.65$ & $0.95$ & $1.00$ & $1.00$ & $0.88$ & $1.00$ & $1.00$ & $1.00$ \\
constant     & \cmark & \shade $1.00$ & \shade $1.00$ & \shade $1.00$ & \shade $1.00$ & \shade $1.00$ & \xmark & $0.42$ & $0.76$ & $0.90$ & $0.18$ & $0.48$ & $0.33$ & $0.33$ & $0.00$ & $0.00$ & $0.00$ & $0.00$ \\
random       & \xmark & $0.22$ & $0.48$ & $0.78$ & $0.05$ & $0.61$ & \xmark & $0.42$ & $0.76$ & $0.90$ & $0.22$ & $0.83$ & $0.34$ & $0.33$ & $0.00$ & $0.00$ & $0.00$ & $0.04$ \\
\bottomrule
\end{tabular}
\end{adjustbox}
\raggedright
\footnotesize{
$^a$ \citep{higgins2017betavae}
$^b$ \citep{kim2018disentangling}
$^c$ \citep{chen2018isolating}
$^d$ \citep{eastwood2018framework}
}
\end{table}


In this section, we empirically demonstrate the effectiveness of the proposed metrics.
Following \citet{carbonneau2022measuring}, we did not learn representations on datasets but directly defined functions $m: Y \to Z$ from factors to codes, which allows us to capture typical failure patterns.

We evaluated modularity metrics based on (i) the radius of the smallest bounding sphere, (ii) mean absolute deviation (MAD) around the median, (iii) variance, (iv) diameter, and (v) mean pairwise distance (MPD) introduced in \cref{ssec:product_approximation,ssec:constancy}.
We evaluated informativeness metrics based on retraction approximation using the maximum error (ME), mean absolute error (MAE), and mean squared error (MSE), and we calculated the contraction discussed in \cref{ssec:informativeness}.
To compare with existing metrics \citep{higgins2017betavae, kim2018disentangling, chen2018isolating, eastwood2018framework}, we transformed the results isomorphically using $e^{-x}: [0, \infty] \to [0, 1]$, meaning that $1$ is the perfect score.
The results are shown in \cref{tab:main}, and our observations are as follows.

\textbf{If a representation is given a perfect score by a proposed metric, it must satisfy the property that the metric quantifies}, which confirms our theoretical result.
In \cref{tab:main}, the\colorbox{black!10}{light}cells show that the proposed metrics can assign a perfect score when the function truly satisfies the properties, which is indicated by $\cmark$ and $\xmark$.
The\colorbox{black!30}{\color{white}dark}cells are supposed to be perfect scores, but they fall short due to the limited expressiveness of the linear models used for the approximation.
Meanwhile, some existing metrics that only provide a single score may entangle modularity and informativeness.

\textbf{The metrics derived from equivalent definitions may differ in terms of computation cost and differentiability.}
Concretely, the radius, MAD, ME, MAE, and MSE are not differentiable due to the inner optimization problem, while the variance, diameter, MPD, and max/mean contraction are differentiable.
In terms of computation, they are much faster than metrics requiring hyperparameter tuning, such as DCI \citep{eastwood2018framework}.
Further comparisons are provided in \cref{app:experiments}.

\textbf{Different metrics may rank imperfect representations differently, even though they have exactly the same minimizers.}
This difference can lead to differences in risk preferences, sensitivity to outliers, and learning dynamics when these metrics are used as learning objectives.
Illustrations and further discussion can be found in \cref{app:discussions}.

Further, the proposed metrics can be used in \emph{weakly supervised} or \emph{fine-grained} evaluation.
See \cref{app:experiments} for detailed data configuration and further experimental results.

\section{Conclusion}
In this work, we developed a systematic and rigorous method for converting logical definitions of properties of representation learning models into quantitative metrics (\cref{tab:conversion}).
We applied this method to assess two important and distinct properties of disentangled representations: modularity and informativeness.
We derived two families of metrics for each property based on their logically equivalent definitions.
We theoretically analyzed the minimizers of these metrics (\cref{thm:main}) and compared their differences in terms of computation cost and differentiability.
Future research could compare metrics derived from different aggregators and design appropriate models to optimize these metrics with minimal supervision.

\clearpage
\section*{Acknowledgments}
We thank Paolo Perrone for generously sharing insights on Markov categories enriched in divergence spaces.
We thank Ken Sakayori for reviewing an earlier version of \cref{app:preliminaries,app:theory} and providing constructive suggestions.
We thank Zhiyuan Zhan for insightful discussions on metrics, topology, measures, and optimization of function properties, and for reviewing and proofreading some parts of \cref{app:discussions}.
We thank Masahiro Negishi for valuable discussions on disentanglement metrics and weakly supervised disentanglement.
We thank Jingwen Fu for checking the algebraic concepts in \cref{sec:enrichment}.
We also thank Johannes Ackermann, Xin-Qiang Cai, and Tongtong Fang for their valuable feedback on the manuscript.

YZ was supported by JSPS KAKENHI Grant Number 22KJ0880.
MS was supported by JST CREST Grant Number JPMJCR18A2 and a grant from Apple, Inc.
Any views, opinions, findings, and conclusions or recommendations expressed in this material are those of the authors and should not be interpreted as reflecting the views, policies or position, either expressed or implied, of Apple Inc.

\addcontentsline{toc}{section}{Bibliography}
\bibliography{references,references_algebra,references_category,references_logic,references_metric,references_probability,references_asset}

\begin{thebibliography}{154}
\providecommand{\natexlab}[1]{#1}
\providecommand{\url}[1]{\texttt{#1}}
\expandafter\ifx\csname urlstyle\endcsname\relax
  \providecommand{\doi}[1]{doi: #1}\else
  \providecommand{\doi}{doi: \begingroup \urlstyle{rm}\Url}\fi

\bibitem[Ad{\'a}mek et~al.(1990)Ad{\'a}mek, Herrlich, and Strecker]{adamek1990abstract}
Ji{\v{r}}{\'\i} Ad{\'a}mek, Horst Herrlich, and George Strecker.
\newblock \emph{Abstract and Concrete Categories: The Joy of Cats}.
\newblock John Wiley and Sons, 1990.
\newblock URL \url{http://www.tac.mta.ca/tac/reprints/articles/17/tr17abs.html}.

\bibitem[Amer(1984)]{amer1984equationally}
K~Amer.
\newblock Equationally complete classes of commutative monoids with monus.
\newblock \emph{Algebra Universalis}, 18:\penalty0 129--131, 1984.
\newblock URL \url{https://doi.org/10.1007/BF01182254}.

\bibitem[Amos et~al.(2017)Amos, Xu, and Kolter]{amos2017input}
Brandon Amos, Lei Xu, and J~Zico Kolter.
\newblock Input convex neural networks.
\newblock In \emph{International Conference on Machine Learning}, 2017.
\newblock URL \url{http://proceedings.mlr.press/v70/amos17b.html}.

\bibitem[Awodey(2010)]{awodey2010category}
Steve Awodey.
\newblock \emph{Category Theory}.
\newblock Oxford University Press, 2010.
\newblock URL \url{https://doi.org/10.1093/acprof:oso/9780198568612.001.0001}.

\bibitem[Bacci et~al.(2023)Bacci, Mardare, Panangaden, and Plotkin]{bacci2023propositional}
Giorgio Bacci, Radu Mardare, Prakash Panangaden, and Gordon Plotkin.
\newblock Propositional logics for the {Lawvere} quantale.
\newblock \emph{Electronic Notes in Theoretical Informatics and Computer Science}, 3, 2023.
\newblock URL \url{https://doi.org/10.46298/entics.12292}.
\newblock \url{https://arxiv.org/abs/2302.01224}.

\bibitem[Bacci et~al.(2024)Bacci, Mardare, Panangaden, and Plotkin]{bacci2024polynomial}
Giorgio Bacci, Radu Mardare, Prakash Panangaden, and Gordon Plotkin.
\newblock Polynomial {Lawvere} logic.
\newblock \emph{arXiv preprint}, 2024.
\newblock URL \url{https://arxiv.org/abs/2402.03543}.

\bibitem[Badreddine et~al.(2022)Badreddine, Garcez, Serafini, and Spranger]{badreddine2022logic}
Samy Badreddine, Artur~d'Avila Garcez, Luciano Serafini, and Michael Spranger.
\newblock Logic tensor networks.
\newblock \emph{Artificial Intelligence}, 303:\penalty0 103649, 2022.
\newblock URL \url{https://doi.org/10.1016/j.artint.2021.103649}.
\newblock \url{https://arxiv.org/abs/2012.13635}.

\bibitem[Balabin et~al.(2024)Balabin, Voronkova, Trofimov, Burnaev, and Barannikov]{balabin2023disentanglement}
Nikita Balabin, Daria Voronkova, Ilya Trofimov, Evgeny Burnaev, and Serguei Barannikov.
\newblock Disentanglement learning via topology.
\newblock In \emph{International Conference on Machine Learning}, 2024.

\bibitem[Bao and Sugiyama(2020)]{bao2020fractional}
Han Bao and Masashi Sugiyama.
\newblock Calibrated surrogate maximization of linear-fractional utility in binary classification.
\newblock In \emph{International Conference on Artificial Intelligence and Statistics}, pages 2337--2347, 2020.
\newblock URL \url{http://proceedings.mlr.press/v108/bao20a.html}.

\bibitem[Bao et~al.(2020)Bao, Scott, and Sugiyama]{bao2020adversarial}
Han Bao, Clay Scott, and Masashi Sugiyama.
\newblock Calibrated surrogate losses for adversarially robust classification.
\newblock In \emph{Conference on Learning Theory}, pages 408--451, 2020.
\newblock URL \url{http://proceedings.mlr.press/v125/bao20a.html}.

\bibitem[Barr and Wells(1990)]{barr1990category}
Michael Barr and Charles Wells.
\newblock \emph{Category Theory for Computing Science}, volume~1.
\newblock Prentice Hall New York, 1990.
\newblock URL \url{https://www.math.mcgill.ca/triples/Barr-Wells-ctcs.pdf}.

\bibitem[Basu et~al.(2002)Basu, Banerjee, and Mooney]{basu2002semi}
Sugato Basu, Arindam Banerjee, and Raymond~J Mooney.
\newblock Semi-supervised clustering by seeding.
\newblock In \emph{International Conference on Machine Learning}, 2002.
\newblock URL \url{https://dl.acm.org/doi/10.5555/645531.656012}.

\bibitem[Behrmann et~al.(2019)Behrmann, Grathwohl, Chen, Duvenaud, and Jacobsen]{behrmann2019invertible}
Jens Behrmann, Will Grathwohl, Ricky~TQ Chen, David Duvenaud, and J{\"o}rn-Henrik Jacobsen.
\newblock Invertible residual networks.
\newblock In \emph{International Conference on Machine Learning}, 2019.
\newblock URL \url{https://proceedings.mlr.press/v97/behrmann19a.html}.

\bibitem[Ben~Yaacov(2022)]{yaacov2022expressive}
Ita{\"\i} Ben~Yaacov.
\newblock On the expressive power of quantifiers in continuous logic.
\newblock \emph{arXiv preprint}, 2022.
\newblock URL \url{https://arxiv.org/abs/2207.01863}.

\bibitem[Ben~Yaacov and Usvyatsov(2010)]{yaacov2010continuous}
Ita{\"\i} Ben~Yaacov and Alexander Usvyatsov.
\newblock Continuous first order logic and local stability.
\newblock \emph{Transactions of the American Mathematical Society}, 362\penalty0 (10):\penalty0 5213--5259, 2010.
\newblock URL \url{https://doi.org/10.1090/S0002-9947-10-04837-3}.
\newblock \url{https://arxiv.org/abs/0801.4303}.

\bibitem[Ben~Yaacov et~al.(2008)Ben~Yaacov, Berenstein, Henson, and Usvyatsov]{yaacov2008model}
Ita{\"\i} Ben~Yaacov, Alexander Berenstein, C.~Ward Henson, and Alexander Usvyatsov.
\newblock \emph{Model Theory for Metric Structures}, page 315–427.
\newblock London Mathematical Society Lecture Note Series. Cambridge University Press, 2008.
\newblock URL \url{https://doi.org/10.1017/CBO9780511735219.011}.

\bibitem[Bengio et~al.(2013)Bengio, Courville, and Vincent]{bengio2013representation}
Yoshua Bengio, Aaron Courville, and Pascal Vincent.
\newblock Representation learning: A review and new perspectives.
\newblock \emph{IEEE Transactions on Pattern Analysis and Machine Intelligence}, 35\penalty0 (8):\penalty0 1798--1828, 2013.
\newblock URL \url{https://doi.org/10.1109/TPAMI.2013.50}.
\newblock \url{https://arxiv.org/abs/1206.5538}.

\bibitem[Bergmann(2008)]{bergmann2008introduction}
Merrie Bergmann.
\newblock \emph{An Introduction to Many-Valued and Fuzzy Logic: Semantics, Algebras, and Derivation Systems}.
\newblock Cambridge University Press, 2008.
\newblock URL \url{https://doi.org/10.1017/CBO9780511801129}.

\bibitem[Bilenko et~al.(2004)Bilenko, Basu, and Mooney]{bilenko2004integrating}
Mikhail Bilenko, Sugato Basu, and Raymond~J Mooney.
\newblock Integrating constraints and metric learning in semi-supervised clustering.
\newblock In \emph{International Conference on Machine Learning}, 2004.
\newblock URL \url{https://dl.acm.org/doi/10.1145/1015330.1015360}.

\bibitem[Boole(1854)]{boole1854investigation}
George Boole.
\newblock \emph{An Investigation of the Laws of Thought: On Which Are Founded the Mathematical Theories of Logic and Probabilities}.
\newblock Cambridge University Press, 1854.
\newblock URL \url{https://doi.org/10.1017/CBO9780511693090}.

\bibitem[Bradley et~al.(2022)Bradley, Terilla, and Vlassopoulos]{bradley2022enriched}
Tai-Danae Bradley, John Terilla, and Yiannis Vlassopoulos.
\newblock An enriched category theory of language: From syntax to semantics.
\newblock \emph{La Matematica}, 1\penalty0 (2):\penalty0 551--580, 2022.
\newblock URL \url{https://doi.org/10.1007/s44007-022-00021-2}.
\newblock \url{https://arxiv.org/abs/2106.07890}.

\bibitem[Brehmer et~al.(2022)Brehmer, Haan, Lippe, and Cohen]{brehmer2022weakly}
Johann Brehmer, Pim~De Haan, Phillip Lippe, and Taco Cohen.
\newblock Weakly supervised causal representation learning.
\newblock In \emph{Neural Information Processing Systems}, 2022.
\newblock URL \url{https://openreview.net/forum?id=dz79MhQXWvg}.

\bibitem[Brehmer et~al.(2023)Brehmer, Haan, Behrends, and Cohen]{brehmer2023geometric}
Johann Brehmer, Pim~De Haan, S{\"o}nke Behrends, and Taco Cohen.
\newblock Geometric algebra transformer.
\newblock In \emph{Neural Information Processing Systems}, 2023.
\newblock URL \url{https://openreview.net/forum?id=M7r2CO4tJC}.

\bibitem[Breiman et~al.(1984)Breiman, Friedman, Olshen, and Stone]{breiman1984classification}
Leo Breiman, Jerome Friedman, Richard~A. Olshen, and Charles~J. Stone.
\newblock \emph{Classification and Regression Trees}.
\newblock Routledge, 1984.
\newblock URL \url{https://doi.org/10.1201/9781315139470}.

\bibitem[Burgess and Kim(2018)]{3dshapes}
Chris Burgess and Hyunjik Kim.
\newblock {3D} shapes dataset, 2018.
\newblock \url{https://github.com/deepmind/3d-shapes} (Apache License 2.0).

\bibitem[Capucci(2024)]{capucci2024quantifiers}
Matteo Capucci.
\newblock On quantifiers for quantitative reasoning.
\newblock \emph{arXiv preprint}, 2024.
\newblock URL \url{https://arxiv.org/abs/2406.04936}.

\bibitem[Carbonneau et~al.(2022)Carbonneau, Zaidi, Boilard, and Gagnon]{carbonneau2022measuring}
Marc-Andr{\'e} Carbonneau, Julian Zaidi, Jonathan Boilard, and Ghyslain Gagnon.
\newblock Measuring disentanglement: A review of metrics.
\newblock \emph{IEEE Transactions on Neural Networks and Learning Systems}, 2022.
\newblock URL \url{https://doi.org/10.1109/TNNLS.2022.3218982}.
\newblock \url{https://arxiv.org/abs/2012.09276}.

\bibitem[Caselles-Dupr{\'e} et~al.(2019)Caselles-Dupr{\'e}, Garcia~Ortiz, and Filliat]{caselles2019symmetry}
Hugo Caselles-Dupr{\'e}, Michael Garcia~Ortiz, and David Filliat.
\newblock Symmetry-based disentangled representation learning requires interaction with environments.
\newblock In \emph{Neural Information Processing Systems}, 2019.
\newblock URL \url{https://proceedings.neurips.cc/paper/2019/hash/36e729ec173b94133d8fa552e4029f8b-Abstract.html}.

\bibitem[Chang(1958)]{chang1958algebraic}
Chen~Chung Chang.
\newblock Algebraic analysis of many valued logics.
\newblock \emph{Transactions of the American Mathematical society}, 88\penalty0 (2):\penalty0 467--490, 1958.
\newblock URL \url{https://doi.org/10.2307/1993227}.

\bibitem[Chang and Keisler(1966)]{chang1966continuous}
Chen~Chung Chang and H~Jerome Keisler.
\newblock \emph{Continuous Model Theory}, volume~58 of \emph{Annals of Mathematics Studies}.
\newblock Princeton University Press, 1966.
\newblock URL \url{https://doi.org/10.1515/9781400882052}.

\bibitem[Chen et~al.(2018)Chen, Li, Grosse, and Duvenaud]{chen2018isolating}
Ricky~TQ Chen, Xuechen Li, Roger~B Grosse, and David~K Duvenaud.
\newblock Isolating sources of disentanglement in variational autoencoders.
\newblock In \emph{Neural Information Processing Systems}, 2018.
\newblock URL \url{https://proceedings.neurips.cc/paper/2018/hash/1ee3dfcd8a0645a25a35977997223d22-Abstract.html}.

\bibitem[Chen et~al.(2020)Chen, Dobriban, and Lee]{chen2020group}
Shuxiao Chen, Edgar Dobriban, and Jane~H Lee.
\newblock A group-theoretic framework for data augmentation.
\newblock \emph{The Journal of Machine Learning Research}, 21\penalty0 (245):\penalty0 1--71, 2020.
\newblock URL \url{http://jmlr.org/papers/v21/20-163.html}.

\bibitem[Chen et~al.(2024)Chen, Zhou, and Yan]{chen2024going}
Yiting Chen, Zhanpeng Zhou, and Junchi Yan.
\newblock Going beyond neural network feature similarity: The network feature complexity and its interpretation using category theory.
\newblock In \emph{International Conference on Learning Representations}, 2024.
\newblock URL \url{https://openreview.net/forum?id=4bSQ3lsfEV}.

\bibitem[Cho and Jacobs(2019)]{cho2019disintegration}
Kenta Cho and Bart Jacobs.
\newblock Disintegration and {Bayesian} inversion via string diagrams.
\newblock \emph{Mathematical Structures in Computer Science}, 29\penalty0 (7):\penalty0 938--971, 2019.
\newblock URL \url{https://doi.org/10.1017/S0960129518000488}.
\newblock \url{https://arxiv.org/abs/1709.00322}.

\bibitem[Cho(2020)]{cho2020categorical}
Simon Cho.
\newblock Categorical semantics of metric spaces and continuous logic.
\newblock \emph{The Journal of Symbolic Logic}, 85\penalty0 (3):\penalty0 1044--1078, 2020.
\newblock URL \url{https://doi.org/10.1017/jsl.2020.44}.

\bibitem[Cockayne and Melzak(1969)]{cockayne1969euclidean}
Ernest~J Cockayne and Zdzislaw~A Melzak.
\newblock {Euclidean} constructibility in graph-minimization problems.
\newblock \emph{Mathematics Magazine}, 42\penalty0 (4):\penalty0 206--208, 1969.
\newblock URL \url{https://doi.org/10.1080/0025570X.1969.11975961}.

\bibitem[Cohen et~al.(2016)Cohen, Lee, Miller, Pachocki, and Sidford]{cohen2016geometric}
Michael~B Cohen, Yin~Tat Lee, Gary Miller, Jakub Pachocki, and Aaron Sidford.
\newblock Geometric median in nearly linear time.
\newblock In \emph{Proceedings of the forty-eighth annual ACM symposium on Theory of Computing}, pages 9--21, 2016.
\newblock URL \url{https://doi.org/10.1145/2897518.2897647}.

\bibitem[Cohen(2021)]{cohen2021equivariant}
Taco Cohen.
\newblock \emph{Equivariant convolutional networks}.
\newblock PhD thesis, University of Amsterdam, 2021.
\newblock URL \url{https://hdl.handle.net/11245.1/0f7014ae-ee94-430e-a5d8-37d03d8d10e6}.

\bibitem[Cohen and Welling(2014)]{cohen2014learning}
Taco Cohen and Max Welling.
\newblock Learning the irreducible representations of commutative {Lie} groups.
\newblock In \emph{International Conference on Machine Learning}, 2014.
\newblock URL \url{https://proceedings.mlr.press/v32/cohen14.html}.

\bibitem[Cohen and Welling(2015)]{cohen2015transformation}
Taco Cohen and Max Welling.
\newblock Transformation properties of learned visual representations.
\newblock In \emph{International Conference on Learning Representations}, 2015.
\newblock URL \url{http://arxiv.org/abs/1412.7659}.

\bibitem[Cohen and Welling(2016)]{cohen2016group}
Taco Cohen and Max Welling.
\newblock Group equivariant convolutional networks.
\newblock In \emph{International Conference on Machine Learning}, 2016.
\newblock URL \url{http://proceedings.mlr.press/v48/cohenc16.html}.

\bibitem[Cruttwell et~al.(2022)Cruttwell, Gavranovi{\'c}, Ghani, Wilson, and Zanasi]{cruttwell2022categorical}
Geoffrey~SH Cruttwell, Bruno Gavranovi{\'c}, Neil Ghani, Paul Wilson, and Fabio Zanasi.
\newblock Categorical foundations of gradient-based learning.
\newblock In \emph{Programming Languages and Systems}, pages 1--28. Springer International Publishing, 2022.
\newblock URL \url{https://doi.org/10.1007/978-3-030-99336-8_1}.
\newblock \url{https://arxiv.org/abs/2103.01931}.

\bibitem[Curry(1980)]{curry1980some}
Haskell~B. Curry.
\newblock Some philosophical aspects of combinatory logic.
\newblock In \emph{Studies in Logic and the Foundations of Mathematics}, volume 101, pages 85--101. Elsevier, 1980.
\newblock URL \url{https://doi.org/10.1016/S0049-237X(08)71254-0}.

\bibitem[Dagnino and Pasquali(2022)]{dagnino2022logical}
Francesco Dagnino and Fabio Pasquali.
\newblock Logical foundations of quantitative equality.
\newblock In \emph{Logic in Computer Science}, 2022.
\newblock URL \url{https://doi.org/10.1145/3531130.3533337}.

\bibitem[Daniels and Velikova(2010)]{daniels2010monotone}
Hennie Daniels and Marina Velikova.
\newblock Monotone and partially monotone neural networks.
\newblock \emph{IEEE Transactions on Neural Networks}, 21\penalty0 (6):\penalty0 906--917, 2010.
\newblock URL \url{https://doi.org/10.1109/TNN.2010.2044803}.

\bibitem[d'Avila Garcez et~al.(2002)d'Avila Garcez, Broda, and Gabbay]{davilagarcez2002neural}
Artur~S. d'Avila Garcez, Krysia~B. Broda, and Dov~M. Gabbay.
\newblock \emph{Neural-Symbolic Learning Systems}.
\newblock Springer, 2002.
\newblock URL \url{https://doi.org/10.1007/978-1-4471-0211-3}.

\bibitem[de~Haan et~al.(2020)de~Haan, Cohen, and Welling]{de2020natural}
Pim de~Haan, Taco Cohen, and Max Welling.
\newblock Natural graph networks.
\newblock In \emph{Neural Information Processing Systems}, 2020.
\newblock URL \url{https://proceedings.neurips.cc/paper/2020/hash/2517756c5a9be6ac007fe9bb7fb92611-Abstract.html}.

\bibitem[Dittadi et~al.(2021)Dittadi, Tr{\"a}uble, Locatello, Wuthrich, Agrawal, Winther, Bauer, and Sch{\"o}lkopf]{dittadi2021transfer}
Andrea Dittadi, Frederik Tr{\"a}uble, Francesco Locatello, Manuel Wuthrich, Vaibhav Agrawal, Ole Winther, Stefan Bauer, and Bernhard Sch{\"o}lkopf.
\newblock On the transfer of disentangled representations in realistic settings.
\newblock In \emph{International Conference on Learning Representations}, 2021.
\newblock URL \url{https://openreview.net/forum?id=8VXvj1QNRl1}.

\bibitem[Do and Tran(2020)]{do2020theory}
Kien Do and Truyen Tran.
\newblock Theory and evaluation metrics for learning disentangled representations.
\newblock In \emph{International Conference on Learning Representations}, 2020.
\newblock URL \url{https://openreview.net/forum?id=HJgK0h4Ywr}.

\bibitem[Dudzik(2017)]{dudzik2017quantales}
Andrew Dudzik.
\newblock \emph{Quantales and Hyperstructures: Monads, Mo'Problems}.
\newblock PhD thesis, University of California, Berkeley, 2017.
\newblock URL \url{https://arxiv.org/abs/1707.09227}.

\bibitem[Dudzik and Veli{\v{c}}kovi{\'c}(2022)]{dudzik2022graph}
Andrew~Joseph Dudzik and Petar Veli{\v{c}}kovi{\'c}.
\newblock Graph neural networks are dynamic programmers.
\newblock In \emph{Neural Information Processing Systems}, 2022.
\newblock URL \url{https://openreview.net/forum?id=wu1Za9dY1GY}.

\bibitem[Eastwood and Williams(2018)]{eastwood2018framework}
Cian Eastwood and Christopher~KI Williams.
\newblock A framework for the quantitative evaluation of disentangled representations.
\newblock In \emph{International Conference on Learning Representations}, 2018.
\newblock URL \url{https://openreview.net/forum?id=By-7dz-AZ}.

\bibitem[Eastwood et~al.(2023)Eastwood, Nicolicioiu, Von~K{\"u}gelgen, Keki{\'c}, Tr{\"a}uble, Dittadi, and Sch{\"o}lkopf]{eastwood2023dcies}
Cian Eastwood, Andrei~Liviu Nicolicioiu, Julius Von~K{\"u}gelgen, Armin Keki{\'c}, Frederik Tr{\"a}uble, Andrea Dittadi, and Bernhard Sch{\"o}lkopf.
\newblock {DCI-ES}: An extended disentanglement framework with connections to identifiability.
\newblock In \emph{International Conference on Learning Representations}, 2023.
\newblock URL \url{https://openreview.net/forum?id=462z-gLgSht}.

\bibitem[Fagin et~al.(2024)Fagin, Riegel, and Gray]{fagin2024foundations}
Ronald Fagin, Ryan Riegel, and Alexander Gray.
\newblock Foundations of reasoning with uncertainty via real-valued logics.
\newblock \emph{Proceedings of the National Academy of Sciences}, 121\penalty0 (21), 2024.
\newblock URL \url{https://doi.org/10.1073/pnas.2309905121}.

\bibitem[Figueroa and van~den Berg(2022)]{figueroa2022topos}
Daniel Figueroa and Benno van~den Berg.
\newblock A topos for continuous logic.
\newblock \emph{Theory and Applications of Categories}, 38\penalty0 (28), 2022.
\newblock URL \url{http://www.tac.mta.ca/tac/volumes/38/28/38-28abs.html}.

\bibitem[Fischer et~al.(2003)Fischer, G{\"a}rtner, and Kutz]{fischer2003fast}
Kaspar Fischer, Bernd G{\"a}rtner, and Martin Kutz.
\newblock Fast smallest-enclosing-ball computation in high dimensions.
\newblock In \emph{European Symposium on Algorithms}, pages 630--641. Springer, 2003.
\newblock URL \url{https://doi.org/10.1007/978-3-540-39658-1_57}.

\bibitem[Fong and Spivak(2019)]{fong2019invitation}
Brendan Fong and David~I Spivak.
\newblock \emph{An Invitation to Applied Category Theory: Seven Sketches in Compositionality}.
\newblock Cambridge University Press, 2019.
\newblock URL \url{https://doi.org/10.1017/9781108668804}.
\newblock \url{https://arxiv.org/abs/1803.05316}.

\bibitem[Fritz(2020)]{fritz2020synthetic}
Tobias Fritz.
\newblock A synthetic approach to {Markov} kernels, conditional independence and theorems on sufficient statistics.
\newblock \emph{Advances in Mathematics}, 370:\penalty0 107239, 2020.
\newblock URL \url{https://doi.org/10.1016/j.aim.2020.107239}.
\newblock \url{https://arxiv.org/abs/1908.07021}.

\bibitem[Fujii(2023)]{fujii2023ordered}
Soichiro Fujii.
\newblock Ordered semirings and subadditive morphisms.
\newblock \emph{arXiv preprint}, 2023.
\newblock URL \url{https://arxiv.org/abs/2311.03862}.

\bibitem[Fumero et~al.(2021)Fumero, Cosmo, Melzi, and Rodol{\`a}]{fumero2021learning}
Marco Fumero, Luca Cosmo, Simone Melzi, and Emanuele Rodol{\`a}.
\newblock Learning disentangled representations via product manifold projection.
\newblock In \emph{International Conference on Machine Learning}, 2021.
\newblock URL \url{http://proceedings.mlr.press/v139/fumero21a.html}.

\bibitem[Gavranovi{\'c} et~al.(2024)Gavranovi{\'c}, Lessard, Dudzik, von Glehn, Ara{\'u}jo, and Veli{\v{c}}kovi{\'c}]{gavranovic2024position}
Bruno Gavranovi{\'c}, Paul Lessard, Andrew~Joseph Dudzik, Tamara von Glehn, Jo{\~a}o Guilherme~Madeira Ara{\'u}jo, and Petar Veli{\v{c}}kovi{\'c}.
\newblock Position: Categorical deep learning is an algebraic theory of all architectures.
\newblock In \emph{International Conference on Machine Learning}, 2024.
\newblock URL \url{https://openreview.net/forum?id=EIcxV7T0Sy}.

\bibitem[Gondal et~al.(2019)Gondal, Wuthrich, Miladinovic, Locatello, Breidt, Volchkov, Akpo, Bachem, Sch{\"o}lkopf, and Bauer]{mpi3d}
Muhammad~Waleed Gondal, Manuel Wuthrich, Djordje Miladinovic, Francesco Locatello, Martin Breidt, Valentin Volchkov, Joel Akpo, Olivier Bachem, Bernhard Sch{\"o}lkopf, and Stefan Bauer.
\newblock On the transfer of inductive bias from simulation to the real world: a new disentanglement dataset.
\newblock In \emph{Neural Information Processing Systems}, 2019.
\newblock URL \url{https://proceedings.neurips.cc/paper/2019/hash/d97d404b6119214e4a7018391195240a-Abstract.html}.
\newblock \url{https://github.com/rr-learning/disentanglement_dataset} (Creative Commons Attribution 4.0 International License).

\bibitem[Goodfellow et~al.(2009)Goodfellow, Lee, Le, Saxe, and Ng]{goodfellow2009measuring}
Ian Goodfellow, Honglak Lee, Quoc Le, Andrew Saxe, and Andrew Ng.
\newblock Measuring invariances in deep networks.
\newblock In \emph{Neural Information Processing Systems}, 2009.
\newblock URL \url{https://proceedings.neurips.cc/paper/2009/hash/428fca9bc1921c25c5121f9da7815cde-Abstract.html}.

\bibitem[Guerraoui et~al.(2023)Guerraoui, Gupta, and Pinot]{guerraoui2023byzantine}
Rachid Guerraoui, Nirupam Gupta, and Rafael Pinot.
\newblock Byzantine machine learning: A primer.
\newblock \emph{ACM Computing Surveys}, 2023.
\newblock URL \url{https://doi.org/10.1145/3616537}.

\bibitem[H{\'a}jek(1998)]{hajek1998metamathematics}
Petr H{\'a}jek.
\newblock \emph{Metamathematics of Fuzzy Logic}, volume~4 of \emph{Trends in Logic}.
\newblock Springer, 1998.
\newblock URL \url{https://doi.org/10.1007/978-94-011-5300-3}.

\bibitem[Harris et~al.(2020)Harris, Millman, van~der Walt, Gommers, Virtanen, Cournapeau, Wieser, Taylor, Berg, Smith, Kern, Picus, Hoyer, van Kerkwijk, Brett, Haldane, del R{\'{i}}o, Wiebe, Peterson, G{\'{e}}rard-Marchant, Sheppard, Reddy, Weckesser, Abbasi, Gohlke, and Oliphant]{numpy}
Charles~R. Harris, K.~Jarrod Millman, St{\'{e}}fan~J. van~der Walt, Ralf Gommers, Pauli Virtanen, David Cournapeau, Eric Wieser, Julian Taylor, Sebastian Berg, Nathaniel~J. Smith, Robert Kern, Matti Picus, Stephan Hoyer, Marten~H. van Kerkwijk, Matthew Brett, Allan Haldane, Jaime~Fern{\'{a}}ndez del R{\'{i}}o, Mark Wiebe, Pearu Peterson, Pierre G{\'{e}}rard-Marchant, Kevin Sheppard, Tyler Reddy, Warren Weckesser, Hameer Abbasi, Christoph Gohlke, and Travis~E. Oliphant.
\newblock Array programming with {NumPy}.
\newblock \emph{Nature}, 585\penalty0 (7825):\penalty0 357--362, 2020.
\newblock URL \url{https://doi.org/10.1038/s41586-020-2649-2}.
\newblock \url{https://numpy.org}.

\bibitem[Higgins et~al.(2017)Higgins, Matthey, Pal, Burgess, Glorot, Botvinick, Mohamed, and Lerchner]{higgins2017betavae}
Irina Higgins, Loic Matthey, Arka Pal, Christopher Burgess, Xavier Glorot, Matthew Botvinick, Shakir Mohamed, and Alexander Lerchner.
\newblock {beta-VAE}: Learning basic visual concepts with a constrained variational framework.
\newblock In \emph{International Conference on Learning Representations}, 2017.
\newblock URL \url{https://openreview.net/forum?id=Sy2fzU9gl}.

\bibitem[Higgins et~al.(2018)Higgins, Amos, Pfau, Racaniere, Matthey, Rezende, and Lerchner]{higgins2018towards}
Irina Higgins, David Amos, David Pfau, Sebastien Racaniere, Loic Matthey, Danilo Rezende, and Alexander Lerchner.
\newblock Towards a definition of disentangled representations.
\newblock \emph{arXiv preprint}, 2018.
\newblock URL \url{https://arxiv.org/abs/1812.02230}.

\bibitem[Hyv{\"a}rinen and Oja(2000)]{hyvarinen2000independent}
Aapo Hyv{\"a}rinen and Erkki Oja.
\newblock Independent component analysis: Algorithms and applications.
\newblock \emph{Neural networks}, 13\penalty0 (4):\penalty0 411--430, 2000.
\newblock URL \url{https://doi.org/10.1016/S0893-6080(00)00026-5}.

\bibitem[Ishikawa et~al.(2023)Ishikawa, Teshima, Tojo, Oono, Ikeda, and Sugiyama]{ishikawa2023universal}
Isao Ishikawa, Takeshi Teshima, Koichi Tojo, Kenta Oono, Masahiro Ikeda, and Masashi Sugiyama.
\newblock Universal approximation property of invertible neural networks.
\newblock \emph{Journal of Machine Learning Research}, 24\penalty0 (287):\penalty0 1--68, 2023.
\newblock URL \url{https://www.jmlr.org/papers/v24/22-0384.html}.

\bibitem[Jaiswal et~al.(2020)Jaiswal, Babu, Zadeh, Banerjee, and Makedon]{jaiswal2020survey}
Ashish Jaiswal, Ashwin~Ramesh Babu, Mohammad~Zaki Zadeh, Debapriya Banerjee, and Fillia Makedon.
\newblock A survey on contrastive self-supervised learning.
\newblock \emph{Technologies}, 9\penalty0 (1):\penalty0 2, 2020.
\newblock URL \url{https://doi.org/10.3390/technologies9010002}.
\newblock \url{https://arxiv.org/abs/2011.00362}.

\bibitem[Johnstone(2002)]{johnstone2002sketches}
Peter Johnstone.
\newblock \emph{Sketches of an Elephant: A Topos Theory Compendium}.
\newblock Oxford University Press, 2002.

\bibitem[Kelly(1982)]{kelly1982basic}
Max Kelly.
\newblock Basic concepts of enriched category theory.
\newblock In \emph{London Mathematical Society Lecture Note Series}, volume~64. Cambridge University Press, 1982.
\newblock URL \url{http://www.tac.mta.ca/tac/reprints/articles/10/tr10.html}.

\bibitem[Kendall(1938)]{kendall1938new}
Maurice~G. Kendall.
\newblock A new measure of rank correlation.
\newblock \emph{Biometrika}, 30\penalty0 (1/2):\penalty0 81--93, 1938.
\newblock URL \url{https://doi.org/10.2307/2332226}.

\bibitem[Keurti et~al.(2023)Keurti, Pan, Besserve, Grewe, and Sch{\"o}lkopf]{keurti2023homomorphism}
Hamza Keurti, Hsiao-Ru Pan, Michel Besserve, Benjamin~F Grewe, and Bernhard Sch{\"o}lkopf.
\newblock Homomorphism {AutoEncoder} -- learning group structured representations from observed transitions.
\newblock In \emph{International Conference on Machine Learning}, 2023.
\newblock URL \url{https://proceedings.mlr.press/v202/keurti23a.html}.

\bibitem[Kim and Mnih(2018)]{kim2018disentangling}
Hyunjik Kim and Andriy Mnih.
\newblock Disentangling by factorising.
\newblock In \emph{International Conference on Machine Learning}, 2018.
\newblock URL \url{http://proceedings.mlr.press/v80/kim18b.html}.

\bibitem[Kingma and Welling(2014)]{kingma2014auto}
Diederik~P. Kingma and Max Welling.
\newblock Auto-encoding variational bayes.
\newblock In \emph{International Conference on Learning Representations}, 2014.
\newblock URL \url{https://openreview.net/forum?id=33X9fd2-9FyZd}.
\newblock \url{http://arxiv.org/abs/1312.6114}.

\bibitem[K{\"o}hler et~al.(2020)K{\"o}hler, Klein, and No{\'e}]{kohler2020equivariant}
Jonas K{\"o}hler, Leon Klein, and Frank No{\'e}.
\newblock Equivariant flows: exact likelihood generative learning for symmetric densities.
\newblock In \emph{International Conference on Machine Learning}, 2020.
\newblock URL \url{https://proceedings.mlr.press/v119/kohler20a.html}.

\bibitem[Koller and Friedman(2009)]{koller2009probabilistic}
Daphne Koller and Nir Friedman.
\newblock \emph{Probabilistic Graphical Models: Principles and Techniques}.
\newblock MIT press, 2009.
\newblock URL \url{https://mitpress.mit.edu/9780262013192/}.

\bibitem[Kullback and Leibler(1951)]{kullback1951information}
Solomon Kullback and Richard~A. Leibler.
\newblock On information and sufficiency.
\newblock \emph{The Annals of Mathematical Statistics}, 22\penalty0 (1):\penalty0 79--86, 1951.
\newblock URL \url{https://doi.org/10.1214/aoms/1177729694}.
\newblock \url{https://www.jstor.org/stable/2236703}.

\bibitem[Kvinge et~al.(2022)Kvinge, Emerson, Jorgenson, Vasquez, Doster, and Lew]{kvinge2022in}
Henry Kvinge, Tegan Emerson, Grayson Jorgenson, Scott Vasquez, Timothy Doster, and Jesse Lew.
\newblock In what ways are deep neural networks invariant and how should we measure this?
\newblock In \emph{Neural Information Processing Systems}, 2022.
\newblock URL \url{https://openreview.net/forum?id=SCD0hn3kMHw}.

\bibitem[Lawvere(1969)]{lawvere1969adjointness}
F.~William Lawvere.
\newblock Adjointness in foundations.
\newblock \emph{Dialectica}, pages 281--296, 1969.
\newblock URL \url{https://doi.org/10.1111/j.1746-8361.1969.tb01194.x}.
\newblock \url{http://www.tac.mta.ca/tac/reprints/articles/16/tr16abs.html}.

\bibitem[Lawvere(1973)]{lawvere1973metric}
F.~William Lawvere.
\newblock Metric spaces, generalized logic, and closed categories.
\newblock \emph{Rendiconti del seminario mat{\'e}matico e fisico di Milano}, 43:\penalty0 135--166, 1973.
\newblock URL \url{https://doi.org/10.1007/BF02924844}.
\newblock \url{http://www.tac.mta.ca/tac/reprints/articles/1/tr1abs.html}.

\bibitem[Lawvere(1986)]{lawvere1986taking}
F.~William Lawvere.
\newblock Taking categories seriously.
\newblock \emph{Revista colombiana de matematicas}, 20\penalty0 (3-4):\penalty0 147--178, 1986.
\newblock \url{http://www.tac.mta.ca/tac/reprints/articles/8/tr8abs.html}.

\bibitem[Lawvere and Rosebrugh(2003)]{lawvere2003sets}
F~William Lawvere and Robert Rosebrugh.
\newblock \emph{Sets for Mathematics}.
\newblock Cambridge University Press, 2003.
\newblock URL \url{https://doi.org/10.1017/CBO9780511755460}.

\bibitem[Lee et~al.(2019)Lee, Lee, Kim, Kosiorek, Choi, and Teh]{lee2019set}
Juho Lee, Yoonho Lee, Jungtaek Kim, Adam Kosiorek, Seungjin Choi, and Yee~Whye Teh.
\newblock Set transformer: A framework for attention-based permutation-invariant neural networks.
\newblock In \emph{International Conference on Machine Learning}, 2019.
\newblock URL \url{http://proceedings.mlr.press/v97/lee19d}.

\bibitem[Leinster(2010)]{leinster2010informal}
Tom Leinster.
\newblock An informal introduction to topos theory.
\newblock \emph{arXiv preprint}, 2010.
\newblock URL \url{https://arxiv.org/abs/1012.5647}.

\bibitem[Leinster(2014)]{leinster2014basic}
Tom Leinster.
\newblock \emph{Basic Category Theory}.
\newblock Cambridge University Press, 2014.
\newblock URL \url{https://doi.org/10.1017/CBO9781107360068}.
\newblock \url{https://arxiv.org/abs/1612.09375}.

\bibitem[Li et~al.(2020)Li, Murkute, Gyawali, and Wang]{li2020progressive}
Zhiyuan Li, Jaideep~Vitthal Murkute, Prashnna~Kumar Gyawali, and Linwei Wang.
\newblock Progressive learning and disentanglement of hierarchical representations.
\newblock In \emph{International Conference on Learning Representations}, 2020.
\newblock URL \url{https://openreview.net/forum?id=SJxpsxrYPS}.

\bibitem[Locatello et~al.(2019{\natexlab{a}})Locatello, Abbati, Rainforth, Bauer, Sch{\"o}lkopf, and Bachem]{locatello2019fairness}
Francesco Locatello, Gabriele Abbati, Thomas Rainforth, Stefan Bauer, Bernhard Sch{\"o}lkopf, and Olivier Bachem.
\newblock On the fairness of disentangled representations.
\newblock In \emph{Neural Information Processing Systems}, 2019{\natexlab{a}}.
\newblock URL \url{https://proceedings.neurips.cc/paper/2019/hash/1b486d7a5189ebe8d8c46afc64b0d1b4-Abstract.html}.

\bibitem[Locatello et~al.(2019{\natexlab{b}})Locatello, Bauer, Lucic, Raetsch, Gelly, Sch{\"o}lkopf, and Bachem]{locatello2019challenging}
Francesco Locatello, Stefan Bauer, Mario Lucic, Gunnar Raetsch, Sylvain Gelly, Bernhard Sch{\"o}lkopf, and Olivier Bachem.
\newblock Challenging common assumptions in the unsupervised learning of disentangled representations.
\newblock In \emph{International Conference on Machine Learning}, 2019{\natexlab{b}}.
\newblock URL \url{https://proceedings.mlr.press/v97/locatello19a.html}.

\bibitem[Locatello et~al.(2020)Locatello, Poole, R{\"a}tsch, Sch{\"o}lkopf, Bachem, and Tschannen]{locatello2020weakly}
Francesco Locatello, Ben Poole, Gunnar R{\"a}tsch, Bernhard Sch{\"o}lkopf, Olivier Bachem, and Michael Tschannen.
\newblock Weakly-supervised disentanglement without compromises.
\newblock In \emph{International Conference on Machine Learning}, 2020.
\newblock URL \url{http://proceedings.mlr.press/v119/locatello20a.html}.

\bibitem[{\L}ukasiewicz(1920)]{lukasiewicz1920logice}
Jan {\L}ukasiewicz.
\newblock {O logice trójwartościowej (On three-valued logic)}.
\newblock In \emph{Selected Works}, volume~11 of \emph{Studies in Logic}, pages 87--88. North-Holland Publishing Company, 1920.

\bibitem[{\L}ukasiewicz and Tarski(1930)]{lukasiewicz1930untersuchungen}
Jan {\L}ukasiewicz and Alfred Tarski.
\newblock {Untersuchungen {\"u}ber den Aussagenkalk{\"u}l (Investigations on the propositional calculus)}.
\newblock \emph{Comptes Rendus des Séances de la Société des Sciences et des Lettres de Varsovie, Class III}, 23, 1930.

\bibitem[Mac~Lane(1978)]{maclane1978categories}
Saunders Mac~Lane.
\newblock \emph{Categories for the Working Mathematician}.
\newblock Springer, 1978.
\newblock URL \url{https://doi.org/10.1007/978-1-4757-4721-8}.

\bibitem[Mac~Lane and Moerdijk(1994)]{maclane1994sheaves}
Saunders Mac~Lane and Ieke Moerdijk.
\newblock \emph{Sheaves in Geometry and Logic: A First Introduction to Topos Theory}.
\newblock Springer, 1994.
\newblock URL \url{https://doi.org/10.1007/978-1-4612-0927-0}.

\bibitem[Mahon et~al.(2023)Mahon, Shah, and Lukasiewicz]{mahon2023correcting}
Louis Mahon, Lei Shah, and Thomas Lukasiewicz.
\newblock Correcting flaws in common disentanglement metrics.
\newblock \emph{arXiv preprint}, 2023.
\newblock URL \url{https://arxiv.org/abs/2304.02335}.

\bibitem[Malinowski(2007)]{malinowski2007many}
Gregorz Malinowski.
\newblock Many-valued logic and its philosophy.
\newblock In \emph{Handbook of the History of Logic}, volume~8, pages 13--94. Elsevier, 2007.
\newblock URL \url{https://doi.org/10.1016/S1874-5857(07)80004-5}.

\bibitem[Manhaeve et~al.(2018)Manhaeve, Dumancic, Kimmig, Demeester, and De~Raedt]{manhaeve2018deepproblog}
Robin Manhaeve, Sebastijan Dumancic, Angelika Kimmig, Thomas Demeester, and Luc De~Raedt.
\newblock {DeepProbLog}: Neural probabilistic logic programming.
\newblock In \emph{Neural Information Processing Systems}, 2018.
\newblock URL \url{https://proceedings.neurips.cc/paper/2018/hash/dc5d637ed5e62c36ecb73b654b05ba2a-Abstract.html}.

\bibitem[Mardare et~al.(2016)Mardare, Panangaden, and Plotkin]{mardare2016quantitative}
Radu Mardare, Prakash Panangaden, and Gordon Plotkin.
\newblock Quantitative algebraic reasoning.
\newblock In \emph{Logic in Computer Science}, 2016.
\newblock URL \url{https://doi.org/10.1145/2933575.2934518}.

\bibitem[Mardare et~al.(2021)Mardare, Panangaden, and Plotkin]{mardare2021fixed}
Radu Mardare, Prakash Panangaden, and Gordon Plotkin.
\newblock Fixed-points for quantitative equational logics.
\newblock In \emph{Logic in Computer Science}, 2021.
\newblock URL \url{https://doi.org/10.1109/LICS52264.2021.9470662}.

\bibitem[Maron et~al.(2019)Maron, Ben-Hamu, Shamir, and Lipman]{maron2019invariant}
Haggai Maron, Heli Ben-Hamu, Nadav Shamir, and Yaron Lipman.
\newblock Invariant and equivariant graph networks.
\newblock In \emph{International Conference on Learning Representations}, 2019.
\newblock URL \url{https://openreview.net/forum?id=Syx72jC9tm}.

\bibitem[Matthey et~al.(2017)Matthey, Higgins, Hassabis, and Lerchner]{dsprites}
Loic Matthey, Irina Higgins, Demis Hassabis, and Alexander Lerchner.
\newblock {dSprites}: Disentanglement testing sprites dataset, 2017.
\newblock \url{https://github.com/deepmind/dsprites-dataset} (Apache License 2.0).

\bibitem[Mazur(2008)]{mazur2008one}
Barry Mazur.
\newblock When is one thing equal to some other thing?
\newblock In \emph{Proof and Other Dilemmas: Mathematics and Philosophy}, page 221–242. Mathematical Association of America, 2008.
\newblock URL \url{https://doi.org/10.5948/UPO9781614445050.015}.
\newblock \url{https://people.math.harvard.edu/~mazur/preprints/when_is_one.pdf}.

\bibitem[Meer et~al.(1991)Meer, Mintz, Rosenfeld, and Kim]{meer1991robust}
Peter Meer, Doron Mintz, Azriel Rosenfeld, and Dong~Yoon Kim.
\newblock Robust regression methods for computer vision: A review.
\newblock \emph{International journal of computer vision}, 6:\penalty0 59--70, 1991.
\newblock URL \url{https://doi.org/10.1007/BF00127126}.

\bibitem[Megiddo(1983)]{megiddo1983weighted}
Nimrod Megiddo.
\newblock The weighted {Euclidean} 1-center problem.
\newblock \emph{Mathematics of Operations Research}, 8\penalty0 (4):\penalty0 498--504, 1983.
\newblock URL \url{https://doi.org/10.1287/moor.8.4.498}.

\bibitem[Minsker(2015)]{minsker2015geometric}
Stanislav Minsker.
\newblock Geometric median and robust estimation in {Banach} spaces.
\newblock \emph{Bernoulli}, 21\penalty0 (4):\penalty0 2308--2335, 2015.
\newblock URL \url{https://doi.org/10.3150/14-BEJ645}.

\bibitem[Miyato et~al.(2022)Miyato, Koyama, and Fukumizu]{miyato2022unsupervised}
Takeru Miyato, Masanori Koyama, and Kenji Fukumizu.
\newblock Unsupervised learning of equivariant structure from sequences.
\newblock In \emph{Neural Information Processing Systems}, 2022.
\newblock URL \url{https://openreview.net/forum?id=7b7iGkuVqlZ}.

\bibitem[Montero et~al.(2022)Montero, Bowers, Costa, Ludwig, and Malhotra]{montero2022lost}
Milton~L. Montero, Jeffrey Bowers, Rui~Ponte Costa, Casimir~JH Ludwig, and Gaurav Malhotra.
\newblock Lost in latent space: Examining failures of disentangled models at combinatorial generalisation.
\newblock In \emph{Neural Information Processing Systems}, 2022.
\newblock URL \url{https://openreview.net/forum?id=7yUxTNWyQGf}.

\bibitem[Montero et~al.(2021)Montero, Ludwig, Costa, Malhotra, and Bowers]{montero2021role}
Milton~Llera Montero, Casimir~JH Ludwig, Rui~Ponte Costa, Gaurav Malhotra, and Jeffrey Bowers.
\newblock The role of disentanglement in generalisation.
\newblock In \emph{International Conference on Learning Representations}, 2021.
\newblock URL \url{https://openreview.net/forum?id=qbH974jKUVy}.

\bibitem[Mulvey(1986)]{mulvey1986ampersand}
Christopher~J. Mulvey.
\newblock \&.
\newblock \emph{Supplemento ai Rendiconti del Circolo Matem{\`a}tico di Palermo. Serie II}, 12:\penalty0 99--104, 1986.
\newblock URL \url{https://zbmath.org/0633.46065}.

\bibitem[Musgrave et~al.(2020)Musgrave, Belongie, and Lim]{musgrave2020metric}
Kevin Musgrave, Serge Belongie, and Ser-Nam Lim.
\newblock A metric learning reality check.
\newblock In \emph{European Conference on Computer Vision}, pages 681--699, 2020.
\newblock URL \url{https://doi.org/10.1007/978-3-030-58595-2_41}.

\bibitem[Navon et~al.(2023)Navon, Shamsian, Achituve, Fetaya, Chechik, and Maron]{navon2023equivariant}
Aviv Navon, Aviv Shamsian, Idan Achituve, Ethan Fetaya, Gal Chechik, and Haggai Maron.
\newblock Equivariant architectures for learning in deep weight spaces.
\newblock In \emph{International Conference on Machine Learning}, 2023.
\newblock URL \url{https://proceedings.mlr.press/v202/navon23a.html}.

\bibitem[Ni et~al.(2019)Ni, Charoenphakdee, Honda, and Sugiyama]{ni2019calibration}
Chenri Ni, Nontawat Charoenphakdee, Junya Honda, and Masashi Sugiyama.
\newblock On the calibration of multiclass classification with rejection.
\newblock In \emph{Neural Information Processing Systems}, 2019.
\newblock URL \url{https://proceedings.neurips.cc/paper/2019/hash/571d3a9420bfd9219f65b643d0003bf4-Abstract.html}.

\bibitem[Painter et~al.(2020)Painter, Prugel-Bennett, and Hare]{painter2020linear}
Matthew Painter, Adam Prugel-Bennett, and Jonathon Hare.
\newblock Linear disentangled representations and unsupervised action estimation.
\newblock In \emph{Neural Information Processing Systems}, 2020.
\newblock URL \url{https://proceedings.neurips.cc/paper/2020/hash/9a02387b02ce7de2dac4b925892f68fb-Abstract.html}.

\bibitem[Paszke et~al.(2019)Paszke, Gross, Massa, Lerer, Bradbury, Chanan, Killeen, Lin, Gimelshein, Antiga, Desmaison, Kopf, Yang, DeVito, Raison, Tejani, Chilamkurthy, Steiner, Fang, Bai, and Chintala]{pytorch}
Adam Paszke, Sam Gross, Francisco Massa, Adam Lerer, James Bradbury, Gregory Chanan, Trevor Killeen, Zeming Lin, Natalia Gimelshein, Luca Antiga, Alban Desmaison, Andreas Kopf, Edward Yang, Zachary DeVito, Martin Raison, Alykhan Tejani, Sasank Chilamkurthy, Benoit Steiner, Lu~Fang, Junjie Bai, and Soumith Chintala.
\newblock {PyTorch}: An imperative style, high-performance deep learning library.
\newblock In \emph{Neural Information Processing Systems}, 2019.
\newblock URL \url{https://proceedings.neurips.cc/paper/2019/hash/bdbca288fee7f92f2bfa9f7012727740-Abstract.html}.
\newblock \url{https://pytorch.org}.

\bibitem[Pearce-Crump(2023)]{pearce2023categorification}
Edward Pearce-Crump.
\newblock Categorification of group equivariant neural networks.
\newblock \emph{arXiv preprint}, 2023.
\newblock URL \url{https://arxiv.org/abs/2304.14144}.

\bibitem[Pedregosa et~al.(2011)Pedregosa, Varoquaux, Gramfort, Michel, Thirion, Grisel, Blondel, Prettenhofer, Weiss, Dubourg, Vanderplas, Passos, Cournapeau, Brucher, Perrot, and Duchesnay]{sklearn}
F.~Pedregosa, G.~Varoquaux, A.~Gramfort, V.~Michel, B.~Thirion, O.~Grisel, M.~Blondel, P.~Prettenhofer, R.~Weiss, V.~Dubourg, J.~Vanderplas, A.~Passos, D.~Cournapeau, M.~Brucher, M.~Perrot, and E.~Duchesnay.
\newblock {Scikit-learn}: Machine learning in {Python}.
\newblock \emph{Journal of Machine Learning Research}, 12:\penalty0 2825--2830, 2011.
\newblock URL \url{https://jmlr.org/papers/v12/pedregosa11a.html}.
\newblock \url{https://scikit-learn.org}.

\bibitem[Perrone(2023)]{perrone2023markov}
Paolo Perrone.
\newblock {Markov} categories and entropy.
\newblock \emph{IEEE Transactions on Information Theory}, 2023.
\newblock URL \url{https://doi.org/10.1109/TIT.2023.3328825}.
\newblock \url{https://arxiv.org/abs/2212.11719}.

\bibitem[Pfau et~al.(2020)Pfau, Higgins, Botev, and Racani{\`e}re]{pfau2020disentangling}
David Pfau, Irina Higgins, Alex Botev, and S{\'e}bastien Racani{\`e}re.
\newblock Disentangling by subspace diffusion.
\newblock In \emph{Neural Information Processing Systems}, 2020.
\newblock URL \url{https://proceedings.neurips.cc/paper/2020/hash/c9f029a6a1b20a8408f372351b321dd8-Abstract.html}.

\bibitem[Pillutla et~al.(2022)Pillutla, Kakade, and Harchaoui]{pillutla2022robust}
Krishna Pillutla, Sham~M. Kakade, and Zaid Harchaoui.
\newblock Robust aggregation for federated learning.
\newblock \emph{IEEE Transactions on Signal Processing}, 70:\penalty0 1142--1154, 2022.
\newblock URL \url{https://doi.org/10.1109/TSP.2022.3153135}.

\bibitem[Pin(1998)]{pin1998tropical}
Jean-Eric Pin.
\newblock Tropical semirings.
\newblock In \emph{Idempotency}, pages 50--69. Cambridge University Press, 1998.
\newblock URL \url{https://doi.org/10.1017/CBO9780511662508.004}.

\bibitem[Quessard et~al.(2020)Quessard, Barrett, and Clements]{quessard2020learning}
Robin Quessard, Thomas Barrett, and William Clements.
\newblock Learning disentangled representations and group structure of dynamical environments.
\newblock In \emph{Neural Information Processing Systems}, 2020.
\newblock URL \url{https://proceedings.neurips.cc/paper/2020/hash/e449b9317dad920c0dd5ad0a2a2d5e49-Abstract.html}.

\bibitem[Reed et~al.(2015)Reed, Zhang, Zhang, and Lee]{3dcars}
Scott~E Reed, Yi~Zhang, Yuting Zhang, and Honglak Lee.
\newblock Deep visual analogy-making.
\newblock In \emph{Neural Information Processing Systems}, 2015.
\newblock URL \url{https://proceedings.neurips.cc/paper/2015/hash/e07413354875be01a996dc560274708e-Abstract.html}.

\bibitem[Reid and Williamson(2010)]{reid2010composite}
Mark~D. Reid and Robert~C. Williamson.
\newblock Composite binary losses.
\newblock \emph{Journal of Machine Learning Research}, 11\penalty0 (83):\penalty0 2387--2422, 2010.
\newblock URL \url{http://jmlr.org/papers/v11/reid10a.html}.

\bibitem[Rezende and Mohamed(2015)]{rezende2015variational}
Danilo Rezende and Shakir Mohamed.
\newblock Variational inference with normalizing flows.
\newblock In \emph{International Conference on Machine Learning}, 2015.
\newblock URL \url{https://proceedings.mlr.press/v37/rezende15.html}.

\bibitem[Ridgeway and Mozer(2018)]{ridgeway2018learning}
Karl Ridgeway and Michael~C Mozer.
\newblock Learning deep disentangled embeddings with the f-statistic loss.
\newblock In \emph{Neural Information Processing Systems}, 2018.
\newblock URL \url{https://proceedings.neurips.cc/paper/2018/hash/2b24d495052a8ce66358eb576b8912c8-Abstract.html}.

\bibitem[Roth et~al.(2023)Roth, Ibrahim, Akata, Vincent, and Bouchacourt]{roth2023disentanglement}
Karsten Roth, Mark Ibrahim, Zeynep Akata, Pascal Vincent, and Diane Bouchacourt.
\newblock Disentanglement of correlated factors via hausdorff factorized support.
\newblock In \emph{International Conference on Learning Representations}, 2023.
\newblock URL \url{https://openreview.net/forum?id=OKcJhpQiGiX}.

\bibitem[Sch{\"o}lkopf and von K{\"u}gelgen(2022)]{scholkopf2022statistical}
Bernhard Sch{\"o}lkopf and Julius von K{\"u}gelgen.
\newblock From statistical to causal learning.
\newblock In \emph{International Congress of Mathematicians}. EMS Press, 2022.
\newblock URL \url{https://doi.org/10.4171/icm2022/173}.
\newblock \url{https://arxiv.org/abs/2204.00607}.

\bibitem[Sen et~al.(2022)Sen, de~Carvalho, Riegel, and Gray]{sen2022neuro}
Prithviraj Sen, Breno~WSR de~Carvalho, Ryan Riegel, and Alexander Gray.
\newblock Neuro-symbolic inductive logic programming with logical neural networks.
\newblock In \emph{Proceedings of the AAAI Conference on Artificial Intelligence}, 2022.
\newblock URL \url{https://doi.org/10.1609/aaai.v36i8.20795}.

\bibitem[Shiebler et~al.(2021)Shiebler, Gavranovi{\'c}, and Wilson]{shiebler2021category}
Dan Shiebler, Bruno Gavranovi{\'c}, and Paul Wilson.
\newblock Category theory in machine learning.
\newblock \emph{arXiv preprint}, 2021.
\newblock URL \url{https://arxiv.org/abs/2106.07032}.

\bibitem[Shiebler(2023)]{shiebler2023compositionality}
Daniel Shiebler.
\newblock \emph{Compositionality and Functorial Invariants in Machine Learning}.
\newblock PhD thesis, University of Oxford, 2023.
\newblock URL \url{http://doi.org/10.5287/bodleian:DE1aDx4Zw}.

\bibitem[Shu et~al.(2020)Shu, Chen, Kumar, Ermon, and Poole]{shu2020weakly}
Rui Shu, Yining Chen, Abhishek Kumar, Stefano Ermon, and Ben Poole.
\newblock Weakly supervised disentanglement with guarantees.
\newblock In \emph{International Conference on Learning Representations}, 2020.
\newblock URL \url{https://openreview.net/forum?id=HJgSwyBKvr}.

\bibitem[Sill(1997)]{sill1997monotonic}
Joseph Sill.
\newblock Monotonic networks.
\newblock In \emph{Neural Information Processing Systems}, 1997.
\newblock URL \url{https://proceedings.neurips.cc/paper/1997/hash/83adc9225e4deb67d7ce42d58fe5157c-Abstract.html}.

\bibitem[Steinwart(2007)]{steinwart2007compare}
Ingo Steinwart.
\newblock How to compare different loss functions and their risks.
\newblock \emph{Constructive Approximation}, 26:\penalty0 225--287, 2007.
\newblock URL \url{https://doi.org/10.1007/s00365-006-0662-3}.

\bibitem[Suter et~al.(2019)Suter, Miladinovic, Sch{\"o}lkopf, and Bauer]{suter2019robustly}
Raphael Suter, Djordje Miladinovic, Bernhard Sch{\"o}lkopf, and Stefan Bauer.
\newblock Robustly disentangled causal mechanisms: Validating deep representations for interventional robustness.
\newblock In \emph{International Conference on Machine Learning}, 2019.
\newblock URL \url{http://proceedings.mlr.press/v97/suter19a.html}.

\bibitem[Tokui and Sato(2022)]{tokui2022disentanglement}
Seiya Tokui and Issei Sato.
\newblock Disentanglement analysis with partial information decomposition.
\newblock In \emph{International Conference on Learning Representations}, 2022.
\newblock URL \url{https://openreview.net/forum?id=pETy-HVvGtt}.

\bibitem[Tonnaer et~al.(2022)Tonnaer, Rey, Menkovski, Holenderski, and Portegies]{tonnaer2022quantifying}
Loek Tonnaer, Luis A~P{\'e}rez Rey, Vlado Menkovski, Mike Holenderski, and Jacobus~W Portegies.
\newblock Quantifying and learning linear symmetry-based disentanglement.
\newblock In \emph{International Conference on Machine Learning}, 2022.
\newblock URL \url{https://proceedings.mlr.press/v162/tonnaer22a.html}.

\bibitem[Tr{\"a}uble et~al.(2021)Tr{\"a}uble, Creager, Kilbertus, Locatello, Dittadi, Goyal, Sch{\"o}lkopf, and Bauer]{trauble2021disentangled}
Frederik Tr{\"a}uble, Elliot Creager, Niki Kilbertus, Francesco Locatello, Andrea Dittadi, Anirudh Goyal, Bernhard Sch{\"o}lkopf, and Stefan Bauer.
\newblock On disentangled representations learned from correlated data.
\newblock In \emph{International Conference on Machine Learning}, 2021.
\newblock URL \url{https://proceedings.mlr.press/v139/trauble21a.html}.

\bibitem[Trimble(2019)]{trimble2019elementary}
Todd Trimble.
\newblock An elementary approach to elementary topos theory, 2019.
\newblock URL \url{https://ncatlab.org/toddtrimble/published/An+elementary+approach+to+elementary+topos+theory}.

\bibitem[Tuzhilin(2016)]{tuzhilin2016invented}
Alexey~A. Tuzhilin.
\newblock Who invented the {Gromov-Hausdorff} distance?
\newblock \emph{arXiv preprint}, 2016.
\newblock URL \url{https://arxiv.org/abs/1612.00728}.

\bibitem[van~der Pol et~al.(2022)van~der Pol, van Hoof, Oliehoek, and Welling]{van2022multiagent}
Elise van~der Pol, Herke van Hoof, Frans~A Oliehoek, and Max Welling.
\newblock Multi-agent {MDP} homomorphic networks.
\newblock In \emph{International Conference on Learning Representations}, 2022.
\newblock URL \url{https://openreview.net/forum?id=H7HDG--DJF0}.

\bibitem[Virtanen et~al.(2020)Virtanen, Gommers, Oliphant, Haberland, Reddy, Cournapeau, Burovski, Peterson, Weckesser, Bright, {van der Walt}, Brett, Wilson, Millman, Mayorov, Nelson, Jones, Kern, Larson, Carey, Polat, Feng, Moore, {VanderPlas}, Laxalde, Perktold, Cimrman, Henriksen, Quintero, Harris, Archibald, Ribeiro, Pedregosa, {van Mulbregt}, and {SciPy 1.0 Contributors}]{scipy}
Pauli Virtanen, Ralf Gommers, Travis~E. Oliphant, Matt Haberland, Tyler Reddy, David Cournapeau, Evgeni Burovski, Pearu Peterson, Warren Weckesser, Jonathan Bright, St{\'e}fan~J. {van der Walt}, Matthew Brett, Joshua Wilson, K.~Jarrod Millman, Nikolay Mayorov, Andrew R.~J. Nelson, Eric Jones, Robert Kern, Eric Larson, C~J Carey, {\.I}lhan Polat, Yu~Feng, Eric~W. Moore, Jake {VanderPlas}, Denis Laxalde, Josef Perktold, Robert Cimrman, Ian Henriksen, E.~A. Quintero, Charles~R. Harris, Anne~M. Archibald, Ant{\^o}nio~H. Ribeiro, Fabian Pedregosa, Paul {van Mulbregt}, and {SciPy 1.0 Contributors}.
\newblock {SciPy 1.0}: Fundamental algorithms for scientific computing in {Python}.
\newblock \emph{Nature methods}, 17\penalty0 (3):\penalty0 261--272, 2020.
\newblock URL \url{https://doi.org/10.1038/s41592-019-0686-2}.
\newblock \url{https://scipy.org}.

\bibitem[Wagstaff et~al.(2001)Wagstaff, Cardie, Rogers, and Schr\"{o}dl]{wagstaff2001constrained}
Kiri Wagstaff, Claire Cardie, Seth Rogers, and Stefan Schr\"{o}dl.
\newblock Constrained k-means clustering with background knowledge.
\newblock In \emph{International Conference on Machine Learning}, 2001.
\newblock URL \url{https://dl.acm.org/doi/10.5555/645530.655669}.

\bibitem[Wang and Isola(2020)]{wang2020understanding}
Tongzhou Wang and Phillip Isola.
\newblock Understanding contrastive representation learning through alignment and uniformity on the hypersphere.
\newblock In \emph{International conference on machine learning}, 2020.
\newblock URL \url{https://proceedings.mlr.press/v119/wang20k.html}.

\bibitem[Weiszfeld(1937)]{weiszfeld1937point}
Endre Weiszfeld.
\newblock Sur le point pour lequel la somme des distances de n points donn{\'e}s est minimum (on the point for which the sum of the distances to n given points is minimum).
\newblock \emph{Tohoku Mathematical Journal, First Series}, 43:\penalty0 355--386, 1937.
\newblock URL \url{https://doi.org/10.1007/s10479-008-0352-z}.

\bibitem[Welzl(1991)]{welzl1991smallest}
Emo Welzl.
\newblock Smallest enclosing disks (balls and ellipsoids).
\newblock In \emph{New Results and New Trends in Computer Science}, pages 359--370. Springer, 1991.
\newblock URL \url{https://doi.org/10.1007/BFb0038202}.

\bibitem[Xu et~al.(2022)Xu, Niethammer, and Raffel]{xu2022compositional}
Zhenlin Xu, Marc Niethammer, and Colin~A Raffel.
\newblock Compositional generalization in unsupervised compositional representation learning: A study on disentanglement and emergent language.
\newblock In \emph{Neural Information Processing Systems}, 2022.
\newblock URL \url{https://openreview.net/forum?id=ZEQ5Gf8DiD}.

\bibitem[Yang et~al.(2022)Yang, Ren, Wang, Zeng, and Zheng]{yang2022towards}
Tao Yang, Xuanchi Ren, Yuwang Wang, Wenjun Zeng, and Nanning Zheng.
\newblock Towards building a group-based unsupervised representation disentanglement framework.
\newblock In \emph{International Conference on Learning Representations}, 2022.
\newblock URL \url{https://openreview.net/forum?id=YgPqNctmyd}.

\bibitem[Yuan(2023)]{yuan2023power}
Yang Yuan.
\newblock On the power of foundation models.
\newblock In \emph{International Conference on Machine Learning}, 2023.
\newblock URL \url{https://proceedings.mlr.press/v202/yuan23b.html}.

\bibitem[Zaheer et~al.(2017)Zaheer, Kottur, Ravanbakhsh, Poczos, Salakhutdinov, and Smola]{zaheer2017deep}
Manzil Zaheer, Satwik Kottur, Siamak Ravanbakhsh, Barnabas Poczos, Russ~R Salakhutdinov, and Alexander~J Smola.
\newblock Deep sets.
\newblock In \emph{Neural Information Processing Systems}, 2017.
\newblock URL \url{https://proceedings.neurips.cc/paper/2017/hash/f22e4747da1aa27e363d86d40ff442fe-Abstract.html}.

\bibitem[Zhang et~al.(2021)Zhang, Moscovich, and Singer]{zhang2021product}
Sharon Zhang, Amit Moscovich, and Amit Singer.
\newblock Product manifold learning.
\newblock In \emph{International Conference on Artificial Intelligence and Statistics}, 2021.
\newblock URL \url{http://proceedings.mlr.press/v130/zhang21j.html}.

\bibitem[Zhang and Sugiyama(2023)]{zhang2023category}
Yivan Zhang and Masashi Sugiyama.
\newblock A category-theoretical meta-analysis of definitions of disentanglement.
\newblock In \emph{International Conference on Machine Learning}, 2023.
\newblock URL \url{https://proceedings.mlr.press/v202/zhang23ak.html}.

\bibitem[Zhou et~al.(2020)Zhou, Zelikman, Lu, Ng, Carlsson, and Ermon]{zhou2020evaluating}
Sharon Zhou, Eric Zelikman, Fred Lu, Andrew~Y Ng, Gunnar Carlsson, and Stefano Ermon.
\newblock Evaluating the disentanglement of deep generative models through manifold topology.
\newblock In \emph{International Conference on Learning Representations}, 2020.
\newblock URL \url{https://openreview.net/forum?id=djwS0m4Ft_A}.

\end{thebibliography}
\bibliographystyle{plainnat}

\newpage
\appendix
{
\hypersetup{linkcolor=black}
\tableofcontents
\listoffigures
\listoftables
}

\newpage
\section{Preliminaries}
\label{app:preliminaries}
In this paper, we used abstract mathematical tools such as category theory and topos theory to develop a theory of the relationship between logical definitions and quantitative metrics.
However, this level of abstraction may be unfamiliar or even intimidating to some readers, and sometimes unnecessary for machine learning practitioners.
Therefore, we have used only the most basic algebraic concepts, such as homomorphism, in the main text.
For readers interested in the mathematical background, we provide a brief introduction to the basic categorical concepts in this section.


\subsection{Basic category theory}

\emph{Category theory} is a branch of mathematics that studies mathematical structures in an abstract way, which is suitable for identifying and organizing common patterns across various fields of mathematics \citep{maclane1978categories, adamek1990abstract, awodey2010category}.
It has found applications in many fields, including computer science \citep{barr1990category}, probability theory \citep{cho2019disintegration, fritz2020synthetic, perrone2023markov}, and machine learning \citep{de2020natural, shiebler2021category, cruttwell2022categorical, dudzik2022graph, shiebler2023compositionality, yuan2023power, pearce2023categorification, chen2024going, gavranovic2024position}.

The most fundamental concept is that of a \emph{category}:
\begin{definition}
\label{def:category}
A \emph{category} $\cC = (\Obj, \Hom, \compL, \id)$ consists of
\begin{itemize}
\item a collection $\Obj$ of objects,
\item a \uline{set} $\Hom(A, B)$ of morphisms between objects,\footnote{To be more precise, a category whose morphisms are sets is called a \emph{locally small} category.}
\item a composition \uline{function} $\compL: \Hom(B, C) \times \Hom(A, B) \to \Hom(A, C)$ for each triple of objects, and
\item an identity morphism $\id_A \in \Hom(A, A)$ for each object,
\end{itemize}
subject to
\begin{itemize}
\item associativity: $(h \compL g) \compL f = h \compL (g \compL f)$ and
\item identity: $\id_B \compL f = f = f \compL \id_A$.
\end{itemize}
\end{definition}

A crucial example is the category $\cSet$ of sets and functions.
However, what is of particular interest is not the category itself but its relationships with other categories.
Building on the concepts of the \emph{functor} and \emph{natural transformation}, whose definitions are omitted here, we can develop tools to better understand the properties of a category.

Moreover, we can \emph{define} objects in terms of their relations with other objects, employing what is known as \emph{universal construction}.
For example, a \emph{terminal object} $1$ in a category is an object such that for any object $A$, there exists a unique morphism $e_A: A \to 1$ to it, which we call a \emph{terminal morphism}.
In $\cSet$, any set $\singleton$ with only one element is a terminal object.
Based on the concept of the terminal object, a \emph{global element} of an object $B$ is defined to be a morphism $b: 1 \to B$ from a terminal object.
We write $b_A: A \to B$ as an abbreviation for the \emph{constant morphism} $b \compL e_A: A \xto{e_A} 1 \xto{b} B$ with value $b: 1 \to B$.
These concepts will be used to develop our theory in \cref{app:theory}.
Other important universal constructions include the \emph{product}, \emph{pullback}, and \emph{exponential}.
The concept of the product is of great importance in disentangled representation learning \citep{zhang2023category}.

Regarding the pullback, we need to mention the following useful lemma:
\begin{lemma}[Pullback lemma]
Suppose that in the following commutative diagram, the right square is a pullback.
\begin{equation}
\begin{tikzcd}
\cdot
\arrow[r]
\arrow[d]
&
\cdot
\arrow[r]
\arrow[d]
&
\cdot
\arrow[d]
\\
\cdot
\arrow[r]
&
\cdot
\arrow[r]
&
\cdot
\end{tikzcd}
\end{equation}
Then, the left square is a pullback if and only if the outer rectangle is a pullback.
\end{lemma}

This lemma is usually left as an exercise in textbooks \citetext{e.g., \citealp[p.~72, Exercise~8]{maclane1978categories}, 
\citealp[Exercise~5.1.35]{leinster2014basic}}.
A proof can be found in \citet[Proposition~7.3]{fong2019invitation}.
We need to use this lemma to (de)compose pullbacks.
As a side note, we use the asterisk $f^*g$ to denote the pullback of $g$ along $f$.


\subsection{Elementary topos theory}

\emph{Topos theory} studies categories that, in some sense, exhibit behavior akin to the category of sets and functions \citep{lawvere2003sets}.
Topos theory has found applications in geometry, topology, and logic \citep{maclane1994sheaves, johnstone2002sketches, leinster2010informal, trimble2019elementary}.
In this work, we only explore its relation to logic.

To formally define a \emph{topos}, two essential concepts are those of the subobject and subobject classifier.
A \emph{subobject} of an object $C$ is simply a monomorphism $b: B \mono C$ to the object $C$.
The subobject classifier is defined as follows:
\begin{definition}
\label{def:subobject_classifier}
In a finitely complete category, a \emph{subobject classifier} is a universal subobject $\lT: 1 \mono \Omega$ such that for every subobject $b: B \mono C$, there exists a unique morphism $\chi_b: C \to \Omega$ such that $b$ is a pullback of $\lT$ along $\chi_b$.
The morphism $\chi_b$ is called the \emph{classifying morphism} of $b$.
\begin{equation}
\begin{tikzcd}
B
\arrow[r, unique morphism]
\arrow[d, "b"', monomorphism]
\arrow[rd, "\lrcorner", phantom, very near start]
&
1
\arrow[d, "\lT", monomorphism]
\\
C
\arrow[r, "\chi_b"', unique morphism]
&
\Omega
\end{tikzcd}
\end{equation}
\end{definition}

Alternatively, we can state that
\begin{proposition}
\label{prop:classifier_terminal}
A subobject classifier is precisely a terminal object in the category of monomorphisms and pullbacks.
\end{proposition}

Then, we can study the morphisms to the object $\Omega$:
\begin{definition}
\label{def:predicate}
In a category with a subobject classifier $\lT: 1 \mono \Omega$, a \emph{predicate} on an object $C$ is a morphism $p: C \to \Omega$.
\end{definition}

Based on \cref{def:subobject_classifier}, we can state that subobjects of an object $C$ are classified by predicates on $C$.

For example, in $\cSet$, subobjects are subsets, a function from a singleton $\singleton$ to a two-element set is a subobject classifier, which is usually denoted by $\lT: \singleton \to \truth$, a predicate on a set $C$ is a function $p: C \to \truth$, and a subset precisely corresponds to its indicator function.

Among various equivalent definitions of a topos, a concise one is as follows:
\begin{definition}
An \emph{elementary topos} is a finitely complete and cartesian closed category with a subobject classifier.
\end{definition}

Despite its concise definition, a great number of logical structures can be derived from it, which will be explored in \cref{app:theory}.


\subsection{Enriched category theory}

\emph{Enriched category theory} generalizes the concept of the category by replacing the sets of morphisms with objects in a suitable category \citep{kelly1982basic}.
It has been used to better understand a wide range of domains, from metric spaces \citep{lawvere1973metric} to language \citep{bradley2022enriched}.

Let us dive into the definition of an enriched category:
\begin{definition}
\label{def:enriched_category}
A category $\cC = (\Obj, \Hom, \compL, \id)$ enriched in a monoidal category $(\cV, \otimes, I)$ consists of
\begin{itemize}
\item a collection $\Obj$ of objects,
\item a \uline{hom-object} $\Hom(A, B) \in \Obj_\cV$ between objects,
\item a composition \uline{morphism} $\compL: \Hom(B, C) \otimes \Hom(A, B) \to \Hom(A, C)$ for each triple of objects, and
\item an identity element $\id_A: I \to \Hom(A, A)$ for each object,
\end{itemize}
subject to associativity and identity.
\end{definition}

Comparing \cref{def:category,def:enriched_category}, we can say that a (locally small) category is a category enriched in the category $\cSet$ of sets and functions.
Enrichment is a way to describe the additional structures of morphisms and the properties that need to be respected by composition.

An example is a preorder, which can be seen as a category enriched in the category of boolean values.
Another example is a \emph{Lawvere metric space} \citep{lawvere1973metric}, which is a set with a premetric that satisfies the triangle inequality.
Note that the transitivity of a preorder and the triangle inequality of a Lawvere metric are described by their composition morphisms, respectively.

In this work, we use enrichment to describe the association of a set of morphisms with additional operations, such as a strict premetric and aggregators.

\section{Theory}
\label{app:theory}
In this section, we detail the theory of converting logical definitions into their corresponding quantitative metrics based on elementary topos theory and enriched category theory.


\subsection{Subobject quantifier and quantizer}
\label{ssec:subobject_quantifier}

Since our main goal is to develop a multi-valued (possibly continuous and differentiable) quantification of properties defined by a certain type of logic, we begin with a category $\cE$ that has sufficient structures to allow the desired logical operations and build the quantification upon these structures.

Firstly, recall that a subobject classifier $\Omega$, if exists, is the \emph{representing object} of the subobject functor $\Sub_\cE$, such that a subobject $b: B \mono C$ corresponds to a unique classifying morphism $\chi_b: C \to \Omega$, and an \emph{external operation} on the set $\Sub_\cE(C)$ of subobjects that is natural in the object $C$ corresponds to an \emph{internal operation}.

For example, the intersection
\begin{equation}
\cap_C: \Sub_\cE(C) \times \Sub_\cE(C) \to \Sub_\cE(C)
\end{equation}
corresponds a natural transformation
\begin{equation}
\Hom_\cE(-, \Omega) \times \Hom_\cE(-, \Omega) \nat \Hom_\cE(-, \Omega),
\end{equation}
which, because the hom-functor $\Hom_\cE$ preserves limits, is isomorphic to a natural transformation between hom-functors
\begin{equation}
\Hom_\cE(-, \Omega \times \Omega) \nat \Hom_\cE(-, \Omega),
\end{equation}
which, by the Yoneda lemma, is isomorphic to an internal operation on the subobject classifier $\Omega$:
\begin{equation}
\lcon: \Omega \times \Omega \to \Omega.
\end{equation}

Note that the subobject classifier, the classifying morphisms, and those internal operations are determined uniquely up to isomorphism.
However, in order to obtain a multi-valued quantification, the requirement for uniqueness might be too restrictive.
Thus, we propose to study a weakened concept instead:
\begin{definition}
\label{def:subobject_quantifier}
In a finitely complete category, a \emph{subobject quantifier} is a subobject $\omicron: 1 \mono \Psi$ such that for every subobject $b: B \mono C$, there exists at least one morphism $\phi_b: C \to \Psi$ such that $b$ is a pullback of $\omicron$ along $\phi_b$.
The morphism $\phi_b$ is called a \emph{quantifying morphism} of $b$.
If the category has a subobject classifier $\lT: 1 \mono \Omega$, the \emph{quantizer} $\kappa: \Psi \to \Omega$ of the subobject quantifier $\omicron$ is the classifying morphism of $\omicron$.
\begin{equation}
\begin{tikzcd}
B
\arrow[r, unique morphism]
\arrow[d, "b"', monomorphism]
\arrow[rd, "\lrcorner", phantom, very near start]
&
1
\arrow[r, unique morphism]
\arrow[d, "\omicron"', monomorphism]
\arrow[rd, "\lrcorner", phantom, very near start]
&
1
\arrow[d, "\lT", monomorphism]
\\
C
\arrow[r, "\phi_b"']
\arrow[rr, "\chi_b"', unique morphism, bend right=40]
&
\Psi
\arrow[r, "\kappa"', unique morphism]
&
\Omega
\end{tikzcd}
\end{equation}
\end{definition}

More succinctly, we can state that (cf. \cref{prop:classifier_terminal})
\begin{proposition}
\label{prop:quantifier_weakly_terminal}
A subobject quantifier is a weakly terminal object in the category of monomorphisms and pullbacks.
\end{proposition}

Thus, there is a unique morphism $\chi_b: C \to \Omega$ classifying a subobject $b: B \mono C$ of an object $C$, but there could be multiple morphisms $\phi_b: C \to \Psi$ quantifying this subobject.

It is provable that the domain of a terminal object $\lT$ in the category of monomorphisms and pullbacks (\cref{prop:classifier_terminal}) must be a terminal object $1$ in the category $\cE$, but not all weakly terminal objects in the category of monomorphisms and pullbacks (\cref{prop:quantifier_weakly_terminal}) are monomorphisms out of a terminal object.
We choose \cref{def:subobject_quantifier} because we want only one global element $\omicron: 1 \mono \Psi$ to be designated to the truth value $\lT: 1 \mono \Omega$.

Here, we give two examples to motivate this definition of subobject quantifier.
One is related to \emph{three-valued logic} \citetext{\citealp{bergmann2008introduction}, \citealp[Exercise~2.34]{fong2019invitation}}:
\begin{example}
In $\cSet$, the function
\begin{equation}
\texttt{yes}: \singleton \to \set{\texttt{no}, \texttt{maybe}, \texttt{yes}},
\end{equation}
which maps the element $*$ in a singleton set $\singleton$ (a terminal object in $\cSet$) to an element $\texttt{yes}$ in a three-element set $\set{\texttt{no}, \texttt{maybe}, \texttt{yes}}$ is a subobject quantifier.
For any subset $B$ of a set $C$, a quantifying morphism $\phi_b: C \to \Psi$ is a function that maps all elements in the subset $B$ to $\texttt{yes}$ and all other elements to either $\texttt{maybe}$ or $\texttt{no}$.
\end{example}

The other is related to \emph{metric spaces} \citep{lawvere1973metric} and will be our running example in the following subsections:
\begin{example}
In $\cSet$, the function $0: \singleton \to \quant$ selecting the number $0$ out of the set $\quant$ of extended non-negative real numbers is a subobject quantifier.
The quantizer is a function
\begin{equation}
\kappa: \quant \to \truth
\defeq
n \mapsto
\begin{cases}
\lT & n = 0,\\
\lF & n > 0,\\
\end{cases}
\end{equation}
which maps $0$ to $\lT$ and any non-zero number to $\lF$.
\end{example}

Intuitively, with a subobject quantifier, there is only one way to be true, but there may be many ways to be false.
In $\cSet$, a quantizer $\kappa$ maps multiple \say{degrees of truth} from a potentially large, even infinite set $\Psi$ to a smaller set $\Omega$ of truth values, hence the name.

Next, we define the counterpart of the concept of predicate (\cref{def:predicate}):
\begin{definition}
In a category with a subobject quantifier $\omicron: 1 \mono \Psi$, a \emph{quantity} on an object $C$ is a morphism $q: C \to \Psi$.
\end{definition}
Since we weakened the requirement for uniqueness, there is no one-to-one correspondence between subobjects and quantities.
However, they are still related as follows:
\begin{lemma}
\label{lem:quantity_pullback}
In a category with a subobject classifier $\lT: 1 \mono \Omega$, a subobject quantifier $\omicron: 1 \mono \Psi$, and a quantizer $\kappa: \Psi \to \Omega$, a quantity $q: C \to \Psi$ on an object $C$ is a quantifying morphism of a subobject $q^*\omicron$ of the object $C$, which is isomorphic to a subobject $(\kappa \compL q)^*\lT$.
\end{lemma}
\begin{pf}
$q^*\omicron$ and $(\kappa \compL q)^*\lT$ are both subobjects of $C$ because pullbacks preserve subobjects.
Their isomorphism follows from the pullback lemma.
\end{pf}

\begin{lemma}
\label{lem:quantity_subobject}
In a category with a subobject classifier $\lT: 1 \mono \Omega$, a subobject quantifier $\omicron: 1 \mono \Psi$, and a quantizer $\kappa: \Psi \to \Omega$, a quantity $q: C \to \Psi$ on an object $C$ is a quantifying morphism of a subobject $b: B \mono C$ if and only if $\kappa \compL q = \chi_b$, where $\chi_b$ is the classifying morphism of the subobject $b$.
\end{lemma}
\begin{pf}
Necessity follows from the pullback lemma and the uniqueness of the classifying morphism;
sufficiency follows from \cref{lem:quantity_pullback}.
\end{pf}

The following relationship between a quantity and the quantizer is also useful:
\begin{lemma}
\label{lem:quantity}
In a category with a subobject classifier $\lT: 1 \mono \Omega$, a subobject quantifier $\omicron: 1 \mono \Psi$, and a quantizer $\kappa: \Psi \to \Omega$, for any quantity $q: C \to \Psi$ on an object $C$, $q = \omicron_C$ if and only if $\kappa \compL q = \kappa \compL \omicron_C = \lT_C$, where $\omicron_C$ is the constant morphism $\omicron \compL e_C: C \xto{e_C} 1 \xto{\omicron} \Psi$ with value $\omicron: 1 \to \Psi$.
\end{lemma}
\begin{pf}
This is due to the universal property of pullback $\omicron$ of $\lT$ along $\kappa$.
\end{pf}

We can see that the hom-functor on the quantizer $\Hom_\cE(-, \kappa): \Hom_\cE(-, \Psi) \nat \Hom_\cE(-, \Omega)$ is a natural transformation which maps the quantities $\Hom_\cE(C, \Psi)$ on an object $C$ to the predicates $\Hom_\cE(C, \Omega)$ on $C$, which are precisely subobjects of $C$.


\subsection{Equality and premetric}

Next, we take a closer look at a concrete and important predicate --- equality --- and its corresponding quantities.
\begin{definition}
In a category with a subobject classifier $\lT: 1 \to \Omega$, the \emph{equality predicate} $=_C: C \times C \to \Omega$ on an object $C$ is the classifying morphism of the diagonal morphism $\Delta_C: C \mono C \times C \defeq \angles{\id_C, \id_C}$.
\end{definition}

By \cref{lem:quantity_subobject}, a quantity $d_C: C \times C \to \Psi$ is a quantifying morphism of $\Delta_C$ if and only if $\kappa \compL d_C = {=_C}$, depicted in the following diagram:
\begin{equation}
\begin{tikzcd}
C
\arrow[r, unique morphism]
\arrow[d, "\Delta_C"', monomorphism]
\arrow[rd, "\lrcorner", phantom, very near start]
&
1
\arrow[r, unique morphism]
\arrow[d, "\omicron"', monomorphism]
\arrow[rd, "\lrcorner", phantom, very near start]
&
1
\arrow[d, "\lT", monomorphism]
\\
C \times C
\arrow[r, "d_C"']
\arrow[rr, "=_C"', unique morphism, bend right=40]
&
\Psi
\arrow[r, "\kappa"', unique morphism]
&
\Omega
\end{tikzcd}
\end{equation}

In $\cSet$, we have the following definitions:
\begin{definition}
A \emph{premetric} on a set $C$ is a binary function $d_C: C \times C \to \quant$ such that
\begin{equation}
\forall c \in C.\;
d_C(c, c) = 0.
\end{equation}
Or equivalently, $d_C \compL \Delta_C = 0_C$, depicted in the following diagram:
\begin{equation}
\begin{tikzcd}
C
\arrow[r]
\arrow[d, "\Delta_C"', monomorphism]
&
\singleton
\arrow[d, "0", monomorphism]
\\
C \times C
\arrow[r, "d_C"']
&
{\quant}
\end{tikzcd}
\end{equation}
\end{definition}

\begin{definition}
A \emph{strict premetric} on a set $C$ is a premetric $d_C: C \times C \to \quant$ such that
\begin{equation}
\forall c_1 \in C.\;
\forall c_2 \in C.\;
(d_C(c_1, c_2) = 0) \limp (c_1 =_C c_2).
\end{equation}
Or equivalently, $\Delta_C$ is a pullback of $0$ along $d_C$:
\begin{equation}
\begin{tikzcd}
C
\arrow[r]
\arrow[d, "\Delta_C"', monomorphism]
\arrow[rd, "\lrcorner", phantom, very near start]
&
\singleton
\arrow[d, "0", monomorphism]
\\
C \times C
\arrow[r, "d_C"']
&
{\quant}
\end{tikzcd}
\end{equation}
\end{definition}

In other words, strict premetrics are precisely quantifying morphisms of the diagonal morphism $\Delta_C$ in the category $\cSet$ with $0: \singleton \to \quant$ as a subobject quantifier.

Note that the symmetry
\begin{equation}
\forall c_1 \in C.\;
\forall c_2 \in C.\;
d_C(c_1, c_2) = d_C(c_2, c_1)
\end{equation}
and the triangle inequality
\begin{equation}
\forall c_1 \in C.\;
\forall c_2 \in C.\;
\forall c_3 \in C.\;
d_C(c_1, c_2) + d_C(c_2, c_3) \geq d_C(c_1, c_3),
\end{equation}
which make $d_C$ a \emph{metric}, are not required.
The addition $+$ and the order $\geq$ on the set $\quant$ are not needed to define a strict premetric.
However, they are necessary for defining other operations and properties, which will be discussed in the next subsection.


\subsection{Preorder}
\label{ssec:preorder}

There is a preorder $\subseteq_C$ of inclusion defined on the set $\Sub_\cE(C)$ of subobjects of an object $C$: for two subobjects $a: A \mono C$ and $b: B \mono C$, $a \subseteq_C b$ if and only if there exists a morphism $f: A \to B$ such that $a = b \compL f$:
\begin{equation}
\begin{tikzcd}[column sep=1em]
A
\arrow[rr, "f"]
\arrow[rd, "a"', monomorphism]
&&
B
\arrow[ld, "b", monomorphism]
\\
&
C
\end{tikzcd}
\end{equation}
This preorder on the subobjects $\Sub_\cE(C)$ induces a preorder on the predicates $\Hom_\cE(C, \Omega)$ via the isomorphism.
We can generalize this construction and define a preorder on other hom-sets:

\begin{definition}
In a category $\cE$ with pullbacks, the inclusion preorder $\subseteq_C$ on the set $\Sub_\cE(C)$ of subobjects of an object $C$ and a subobject $m: S \mono T$ induce a preorder $\preceq_C^m$ on the hom-set $\Hom_\cE(C, T)$ via pullback of $m$: for any two morphisms $f_1, f_2: C \to T$, $f_1 \preceq_C^m f_2$ if and only if $f_1^*m \subseteq_C f_2^*m$.
\begin{equation}
\begin{tikzcd}
f^*S
\arrow[r]
\arrow[d, "f^*m"', monomorphism]
\arrow[rd, "\lrcorner", phantom, very near start]
&
S
\arrow[d, "m", monomorphism]
\\
C
\arrow[r, "f"']
&
T
\end{tikzcd}
\end{equation}
\end{definition}

Based on this definition, we can explore the preorders on any hom-sets.
From now, we assume that $\cE$ is a category with necessary structures that we need.

Then, the preorder of predicates is $\preceq_C^\lT$ on $\Hom_\cE(C, \Omega)$, and the preorder of quantities is $\preceq_C^\omicron$ on $\Hom_\cE(C, \Psi)$.
By \cref{lem:quantity_pullback}, we know that for two quantities $q_1, q_2: C \to \Psi$, $q_1 \preceq_C^\omicron q_2$ if and only if $(\kappa \compL q_1) \preceq_C^\lT (\kappa \compL q_2)$, which means that $\Hom_\cE(C, \kappa)$ is an order-preserving function from $(\Hom_\cE(C, \Psi), \preceq_C^\omicron)$ to $(\Hom_\cE(C, \Omega), \preceq_C^\lT)$.

Next, we will explore the structures of $\Hom_\cE(C, \Omega)$ and $\Hom_\cE(C, \Psi)$.
To begin with, $\lT_C$ is a top in $\Hom_\cE(C, \Omega)$, and $\omicron_C$ is a top in $\Hom_\cE(C, \Psi)$, because $\id_C$ is a top in $\Sub_\cE(C)$.
\begin{equation}
\begin{tikzcd}
C
\arrow[r, unique morphism]
\arrow[d, "\id_C"', monomorphism]
\arrow[rd, "\lrcorner", phantom, very near start]
&
1
\arrow[r, unique morphism]
\arrow[d, "\omicron"', monomorphism]
\arrow[rd, "\lrcorner", phantom, very near start]
&
1
\arrow[d, "\lT", monomorphism]
\\
C
\arrow[r, "\omicron_C"']
\arrow[rr, "\lT_C"', unique morphism, bend right=40]
&
\Psi
\arrow[r, "\kappa"', unique morphism]
&
\Omega
\end{tikzcd}
\end{equation}

The inclusion preorder on $\Sub_\cE(C)$ also has a bottom --- the initial morphism $i_C: 0 \mono C$ from an initial object $0$ in the category $\cE$ to the object $C$.
Then, its classifying morphism $\lF_C: C \to \Omega$ is a bottom in $\Hom_\cE(C, \Omega)$.
It can be proven that $\lF_C$ is a constant morphism with value $\lF: 1 \to \Omega$, which is the classifying morphism of the initial/terminal morphism $0 \mono 1$.
Any quantifying morphism $\psi_C: C \to \Psi$ of $i_C$ is a bottom in $\Hom_\cE(C, \Psi)$, but it is not necessarily a constant morphism.
\begin{equation}
\begin{tikzcd}
0
\arrow[r, unique morphism]
\arrow[d, "i_C"', monomorphism]
\arrow[rd, "\lrcorner", phantom, very near start]
&
1
\arrow[r, unique morphism]
\arrow[d, "\omicron"', monomorphism]
\arrow[rd, "\lrcorner", phantom, very near start]
&
1
\arrow[d, "\lT", monomorphism]
\\
C
\arrow[r, "\psi_C"']
\arrow[rr, "\lF_C"', unique morphism, bend right=40]
&
\Psi
\arrow[r, "\kappa"', unique morphism]
&
\Omega
\end{tikzcd}
\end{equation}

The preorder on the global elements plays a special role:
\begin{example}
In $\cSet$, $\preceq_1^\lT$, also denoted by $\lproves$, is a preorder on the set $\truth$ with only one non-identity relation $\lF \lproves \lT$.
\end{example}

\begin{example}
In $\cSet$, $\preceq_1^0$ is a preorder on the set $\quant$ where $n \preceq_1^0 0$ for any number $n$, and $m \preceq_1^0 n$ and $n \preceq_1^0 m$ for any positive numbers $m$ and $n$.
\end{example}

By definition, $(\truth, \lproves)$ and $(\quant, \preceq_1^0)$ are equivalent.
However, we can consider a \emph{suborder} of $(\quant, \preceq_1^0)$, e.g., the usual \say{greater than or equal to} $\geq$ total order, to further differentiate positive numbers.
Note that $0$ remains the top in this suborder $(\quant, \geq)$.


\subsection{Operation: algebra over the product endofunctor}
\label{ssec:operation}

In an elementary topos $\cE$, the subobjects $\Sub_\cE(C)$ of an object $C$ not only forms a preorder by inclusion but also are equipped with certain \emph{set operations} (e.g., intersection and disjoint union), which are defined in terms of the universal properties of their corresponding \emph{order operations} (e.g., meet and join).
Further, these operations are reflected in the structures of the subobject classifier (e.g., conjunction and disjunction).

Here, we establish a link between the structures of the subobject classifier and those of a subobject quantifier.
Our primary result is as follows:
\begin{theorem}
\label{thm:operation}
Consider a category with a subobject classifier $\lT: 1 \mono \Omega$, a subobject quantifier $\omicron: 1 \mono \Psi$, and a quantizer $\kappa: \Psi \to \Omega$.

Let $n$ be a natural number.
Let $\beta: \Omega^n \to \Omega$ be an $n$-ary logical operation on $\Omega$, and let $\alpha: \Psi^n \to \Psi$ be an $n$-ary quantitative operation on $\Psi$.

For $i \in \set{1, \dots, n}$, let $p_i: C \to \Omega$ be a predicate on an object $C$, and let $q_i: C \to \Psi$ be a quantity on the object $C$ such that $p_i = \kappa \compL q_i$.
Let $p = \angles{p_1, \dots, p_n}$ be the tupling of the predicates, and let $q = \angles{q_1, \dots, q_n}$ be the tupling of the quantities.
Let $b: B \mono C \defeq (\beta \compL p)^*\lT$ be the subobject classified by $\beta \compL p$, and let $a: A \mono C \defeq (\alpha \compL q)^*\omicron$ be the subobject quantified by $\alpha \compL q$.
\begin{equation}
\begin{tikzcd}[column sep=0, row sep=2em]
&[2em]
A
\arrow[rr]
\arrow[dd, "a"', monomorphism, pos=.3]
\arrow[rrdd, "\lrcorner", phantom, pos=.05]
&[2.5em]&[1em]
\alpha^*1
\arrow[rr]
\arrow[dd, "\alpha^*\omicron"', pos=.3]
\arrow[rrdd, "\lrcorner", phantom, pos=.05]
&[3.5em]&[1.5em]
1
\arrow[dd, "\omicron", pos=.3]
\arrow[ld, leftrightarrow]
\\
B
\arrow[rr, crossing over]
\arrow[dd, "b"', monomorphism, pos=.3]
\arrow[rrdd, "\lrcorner", phantom, pos=.05]
&&
\beta^*1
\arrow[rr, crossing over]
\arrow[rrdd, "\lrcorner", phantom, pos=.05]
&&
1
\\
&
C
\arrow[rr, "q"']
\arrow[ld, "\id_C", leftrightarrow]
&&
\Psi^n
\arrow[rr, "\alpha"']
\arrow[ld, "\kappa^n"]
&&
\Psi
\arrow[ld, "\kappa"]
\\
C
\arrow[rr, "p"']
&&
\Omega^n
\arrow[rr, "\beta"']
\arrow[from=uu, "\beta^*\lT"', pos=.3, crossing over]
&&
\Omega
\arrow[from=uu, "\lT", pos=.3, crossing over]
\end{tikzcd}
\end{equation}
Then, we have
\begin{enumerate}[label=(\roman*)]
\item \label{item:tupling_quantization}
$\kappa^n \compL q = p$
\item \label{item:subhomomorphic}
$(\beta \compL \kappa^n \compL \alpha^*\omicron = \lT_{\alpha^*1}) \limp ((\beta \compL p) \compL a = \lT_A)$
\item \label{item:homomorphic_implies_subhomomorphic}
$(\beta \compL \kappa^n = \kappa \compL \alpha) \limp (\beta \compL \kappa^n \compL \alpha^*\omicron = \lT_{\alpha^*1})$
\item \label{item:tupling_predicate_quantity}
$(\beta \compL \kappa^n = \kappa \compL \alpha) \limp (\kappa \compL (\alpha \compL q) = \beta \compL p)$
\item \label{item:homomorphic}
$(\beta \compL \kappa^n = \kappa \compL \alpha) \limp ((\alpha \compL q) \compL b = \omicron_B)$
\end{enumerate}
\end{theorem}

For convenience, we call an $n$-ary operation $\alpha: \Psi^n \to \Psi$ (as an algebra over the product endofunctor $(-)^n$) \emph{homomorphic} to $\beta: \Omega^n \to \Omega$ via a morphism $\kappa: \Psi \to \Omega$ if
\begin{equation}
\beta \compL \kappa^n = \kappa \compL \alpha
\end{equation}
and \emph{subhomomorphic} to $\beta: \Omega^n \to \Omega$ if it satisfies the condition
\begin{equation}
\beta \compL \kappa^n \compL \alpha^*\omicron = \lT_{\alpha^*1}.
\end{equation}

\begin{pf}
\ref{item:tupling_quantization} follows from the property of tupling and product: $\kappa^n \compL q = \angles{\kappa \compL q_1, \dots, \kappa \compL q_n} = \angles{p_1, \dots, p_n} = p$.

\ref{item:subhomomorphic} states that if $\alpha$ is subhomomorphic to $\beta$, then $\beta \compL p$ is the classifying morphism of the subobject quantified by $\alpha \compL q$.
\begin{flalign}
& \beta \compL p \compL a
\\
={}& \beta \compL \kappa^n \compL q \compL a
& \text{\ref{item:tupling_quantization}}
\\
={}& \beta \compL \kappa^n \compL \alpha^*\omicron \compL (\alpha^*\omicron)^*q 
& \text{(pullback)}
\\
={}& \lT_{\alpha^*1} \compL (\alpha^*\omicron)^*q 
& \text{(subhomomorphism)}
\\
={}& \lT_A 
& \text{(composition)}
\end{flalign}

\ref{item:homomorphic_implies_subhomomorphic} means that $\alpha$ being homomorphic to $\beta$ is a stronger condition than being merely subhomomorphic to $\beta$.
\begin{flalign}
& \beta \compL \kappa^n \compL \alpha^*\omicron
\\
={}& \kappa \compL \alpha \compL \alpha^*\omicron
& \text{(homomorphism)}
\\
={}& \kappa \compL \omicron \compL \omicron^*\alpha
& \text{(pullback)}
\\
={}& \lT_{\alpha^*1}
& \text{(composition)}
\end{flalign}

\ref{item:tupling_predicate_quantity} shows the relationship between the predicate $\beta \compL p$ and the quantity $\alpha \compL q$ when $\alpha$ is homomorphic to $\beta$.
\begin{flalign}
& \kappa \compL \alpha \compL q
\\
={}& \beta \compL \kappa^n \compL q
& \text{(homomorphism)}
\\
={}& \beta \compL p
& \text{\ref{item:tupling_quantization}}
\end{flalign}

\ref{item:homomorphic} means that if $\alpha$ is homomorphic to $\beta$, then $\alpha \compL q$ is a quantifying morphism of $b$.
\begin{flalign}
& \kappa \compL \alpha \compL q \compL b
\\
={}& \beta \compL p \compL b
& \text{\ref{item:tupling_predicate_quantity}}
\\
={}& \lT_B
& \text{(pullback)}
\end{flalign}
$\alpha \compL q \compL b = \omicron_B$ follows from \cref{lem:quantity}.
\end{pf}

In summary, if $\alpha$ is subhomomorphic to $\beta$, then $a$ is included in $b$ (\ref{item:subhomomorphic}); if $\alpha$ is homomorphic to $\beta$, then $a$ and $b$ are isomorphic (\ref{item:homomorphic_implies_subhomomorphic} and \ref{item:homomorphic}).
We consider this weaker condition because subhomomorphic but non-homomorphic operations may exhibit favorable properties in other aspects, such as continuity.
We will discuss several concrete examples in the following subsections.


\begin{figure}[t]
\centering
\includegraphics[width=.5\linewidth]{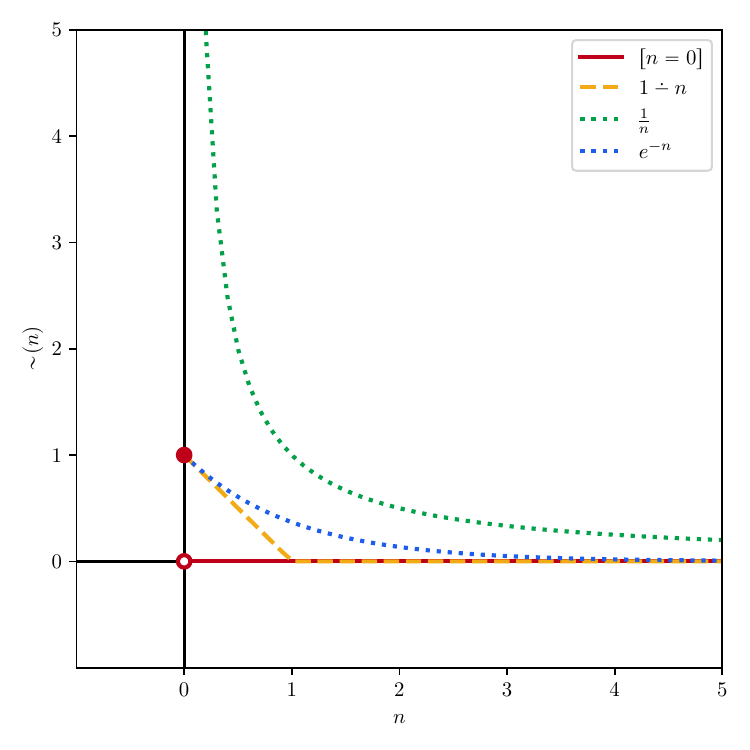}
\vspace{-1em}
\caption{Negation}
\label{fig:negation}
\end{figure}


\subsection{Negation}
\label{ssec:negation}

First, let us take a closer look at a unary logical operation --- \emph{negation} $\lnot: \Omega \to \Omega$, which is defined as the classifying morphism of $\lF: 1 \to \Omega$.
Recall that $\lF$ is the classifying morphism of $0 \mono 1$.

Let us consider a unary quantitative operation $\qnot: \Psi \to \Psi$.
If $\qnot$ is homomorphic to $\lnot$ via the quantizer $\kappa$, it means that $\lnot \compL \kappa = \kappa \compL \qnot$, or the following diagram commutes:
\begin{equation}
\begin{tikzcd}
\Psi
\arrow[r, "\qnot"]
\arrow[d, "\kappa"']
&
\Psi
\arrow[d, "\kappa"]
\\
\Omega
\arrow[r, "\lnot"']
&
\Omega
\end{tikzcd}
\end{equation}

Let us consider the set $\quant$ in $\cSet$.
A quantitative operation $\qnot: \quant \to \quant$ \emph{homomorphic} to the negation $\lnot: \truth \to \truth$ is a function 
\begin{equation}
\qnot(n)
\defeq
[n = 0] \times n_0
=
\begin{cases}
n_0 & n = 0,\\
0 & n > 0,\\
\end{cases}
\end{equation}
which maps $0$ to a non-zero number $n_0$ and any non-zero number to $0$.\footnote{$[-]: \truth \to \quant \defeq \begin{cases}\lF \mapsto 0,\\ \lT \mapsto 1.\end{cases}$}
However, this function is discontinuous at $0$.
On the other hand, if we consider a quantitative operation $\qnot$ \emph{subhomomorphic} to the negation $\lnot$, then the only requirement is that for all $n$, $\qnot(n) = 0$ implies $n > 0$, or, by contraposition, $\qnot(0) > 0$.
Continuous choices include the \emph{hinge function} $n \mapsto 1 \monus n = \max\set{1 - n, 0}$, \emph{reciprocal function} $n \mapsto \frac1n$, and \emph{exponential decay function} $n \mapsto e^{-n}$ (see \cref{fig:negation}).
Note that the latter two are actually homomorphic to the constant false $\lF$ because their outputs are always non-zero.
The hinge function $1 \monus q$, as discussed later, can be seen as derived from the implication $p \limp \lF$.

In this way, if we have a quantity $q$ for a predicate $p$, we can obtain a quantity $\qnot q$ for the negation $\lnot p$ of the predicate as well.
If the quantitative operation $\qnot$ is subhomomorphic but not homomorphic to the logical operation $\lnot$, we can guarantee that for any $x$, $\qnot q(x) = 0$ implies $\lnot p(x) = \lT$, but not vice versa.


\begin{figure}[t]
\centering
\begin{minipage}[t]{0.49\linewidth}
\includegraphics[width=\linewidth]{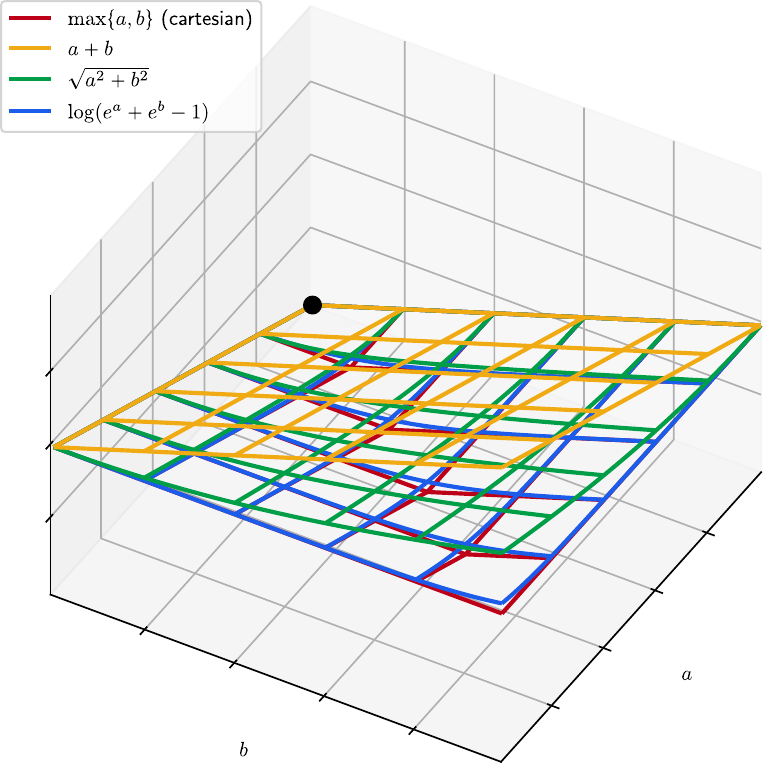}
\caption{Conjunction}
\label{fig:conjunction}
\end{minipage}
\hfill
\begin{minipage}[t]{0.49\linewidth}
\includegraphics[width=\linewidth]{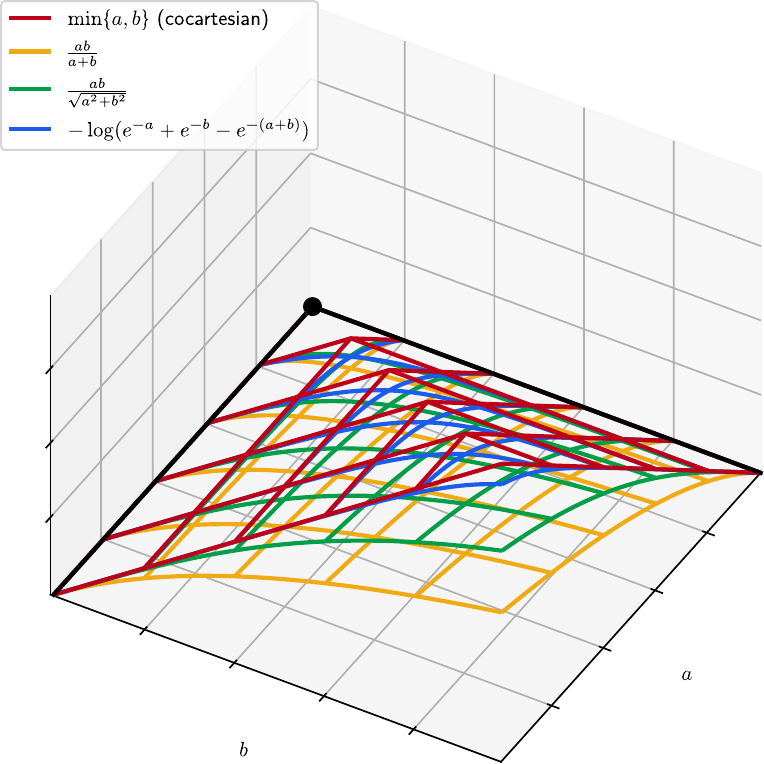}
\caption{Disjunction}
\label{fig:disjunction}
\end{minipage}
\end{figure}


\begin{figure}[t]
\centering
\begin{minipage}[t]{0.49\linewidth}
\includegraphics[width=\linewidth]{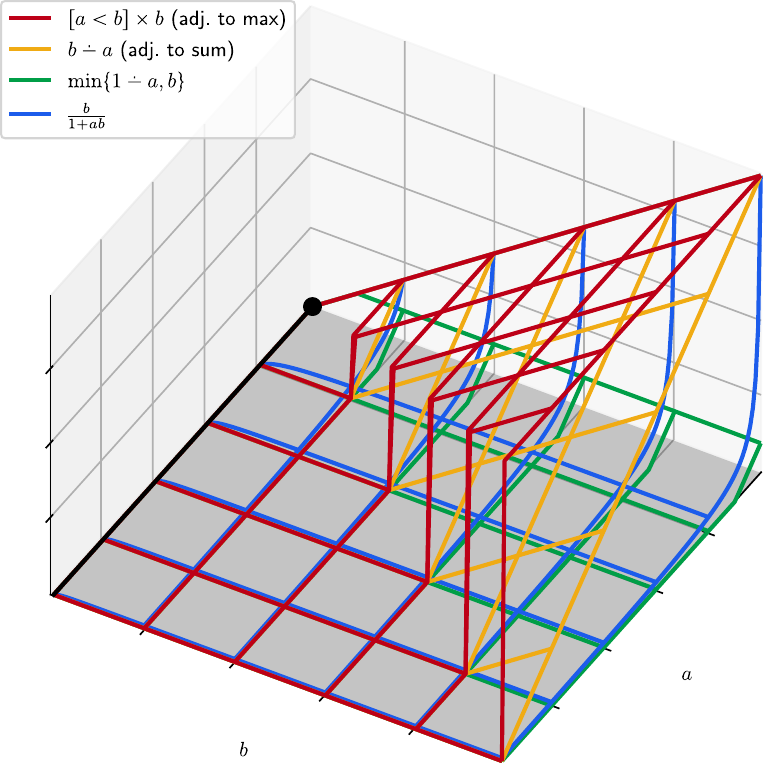}
\caption{Implication}
\label{fig:implication}
\end{minipage}
\hfill
\begin{minipage}[t]{0.49\linewidth}
\includegraphics[width=\linewidth]{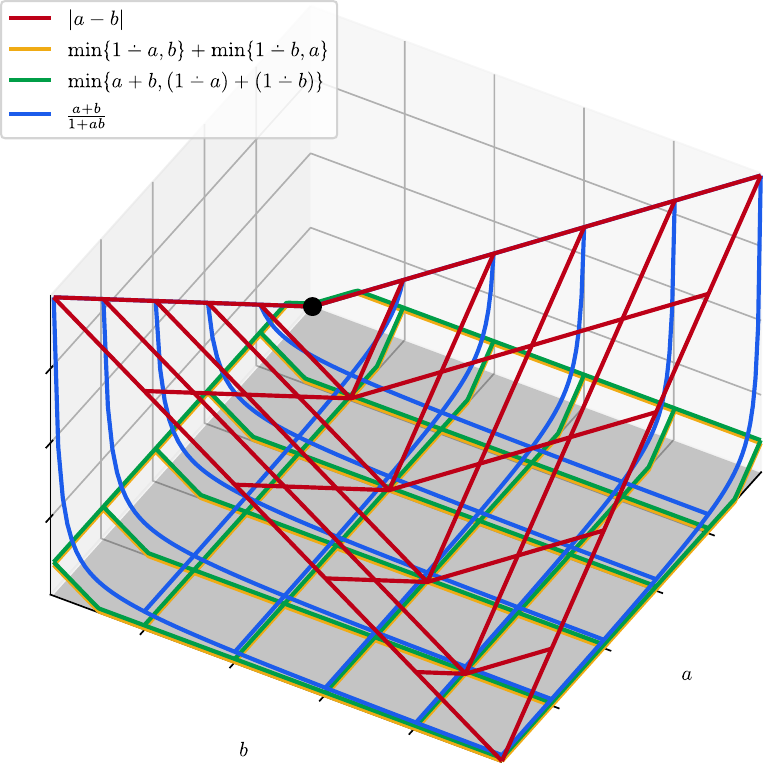}
\caption{Equivalence}
\label{fig:equivalence}
\end{minipage}
\end{figure}


\subsection{Conjunction}
\label{ssec:conjunction}

Next, let us move on to an important binary logical operation --- \emph{conjunction} $\lcon: \Omega \times \Omega \to \Omega$, which is defined as the classifying morphism of $\angles{\lT, \lT}: 1 \mono \Omega \times \Omega$.
Similarly, we consider a binary quantitative operation $\qcon: \Psi \times \Psi \to \Psi$ homomorphic to the conjunction $\lcon$ via the quantizer $\kappa$:
\begin{equation}
\begin{tikzcd}
\Psi \times \Psi
\arrow[r, "\qcon"]
\arrow[d, "\kappa \times \kappa"']
&
\Psi
\arrow[d, "\kappa"]
\\
\Omega \times \Omega
\arrow[r, "\lcon"']
&
\Omega
\end{tikzcd}
\end{equation}

By abuse of notation, the conjunction $\lcon$ also denotes a binary operation on the set $\Hom_\cE(C, \Omega)$ of predicates, such that for any two predicates $p_1, p_2 \in \Hom_\cE(C, \Omega)$,
\begin{equation}
p_1 \lcon p_2 \defeq \lcon \compL \angles{p_1, p_2}.
\end{equation}
The same goes for the quantitative operation $\qcon$.

For the conjunction, we can prove a stronger result:
\begin{theorem}
\label{thm:conjunction}
Consider a category with a subobject classifier $\lT: 1 \mono \Omega$, a subobject quantifier $\omicron: 1 \mono \Psi$, and a quantizer $\kappa: \Psi \to \Omega$.

Let the conjunction $\lcon: \Omega \times \Omega \to \Omega$ be the classifying morphism of $\angles{\lT, \lT}: 1 \mono \Omega \times \Omega$, and let $\qcon: \Psi \times \Psi \to \Psi$ be a binary operation on $\Psi$ homomorphic to the conjunction $\lcon$ via the quantizer $\kappa$.

Let $p_1, p_2: C \to \Omega$ be two predicates on an object $C$, and let $q_1, q_2: C \to \Psi$ be two quantities on the object $C$ such that $p_1 = \kappa \compL q_1$ and $p_2 = \kappa \compL q_2$.
Let $p = \angles{p_1, p_2}$ be the pairing of the predicates, and let $q = \angles{q_1, q_2}$ be the pairing of the quantities.
Let $b: B \mono C \defeq (p_1 \lcon p_2)^*\lT$ be the subobject classified by $p_1 \lcon p_2$.
\begin{equation}
\label{diag:conjunction}
\begin{tikzcd}[column sep=4em]
B
\arrow[r]
\arrow[d, "b"', monomorphism]
\arrow[rd, "\lrcorner", phantom, very near start]
&
1
\arrow[r]
\arrow[d, "\angles{\omicron, \omicron}"', monomorphism]
\arrow[rd, "\lrcorner", phantom, very near start]
&
1
\arrow[d, "\omicron", monomorphism]
\\
C
\arrow[r, "\angles{q_1, q_2}"']
\arrow[d, "\id_C"']
&
\Psi \times \Psi
\arrow[r, "\qcon"']
\arrow[d, "\kappa \times \kappa"']
&
\Psi
\arrow[d, "\kappa"]
\\
C
\arrow[r, "\angles{p_1, p_2}"']
&
\Omega \times \Omega
\arrow[r, "\lcon"']
&
\Omega
\end{tikzcd}
\end{equation}
Then, $\qcon$ is a quantifying morphism of $\angles{\omicron, \omicron}$, and $b$ is a pullback of $\angles{\omicron, \omicron}$ along $\angles{q_1, q_2}$.
\end{theorem}
\begin{pf}
If $q_1 \qcon q_2 = \omicron_C$, then $(\kappa \compL q_1) \lcon (\kappa \compL q_2) = \lT_C$, which leads to $\angles{\kappa \compL q_1, \kappa \compL q_2} = \angles{\lT, \lT} \compL e_C = \angles{\lT_C, \lT_C}$ due to the universal property of pullback.
Then, we have $\kappa \compL q_1 = \kappa \compL q_2 = \lT_C$ due to the universal property of pairing, and consequently $q_1 = q_2 = \omicron_C$ according to \cref{lem:quantity}, i.e., $\angles{q_1, q_2} = \angles{\omicron, \omicron} \compL e_C$.
Therefore, $\angles{\omicron, \omicron}$ is a pullback of $\omicron$ along $\qcon$.
According to \cref{thm:operation}, $b$ is a pullback of $\omicron$ along $q_1 \qcon q_2$.
Then, following the pullback lemma, $b$ is a pullback of $\angles{\omicron, \omicron}$ along $\angles{q_1, q_2}$.
\end{pf}

In other words, the pullback square symbols in \cref{diag:conjunction} are unambiguous --- the top row, the left column, the right column, the top-left square, and the top-right square are all pullbacks.

Note that for any quantities $a, b, c: C \to \Psi$, we have
\begin{align}
\kappa \compL ((a \qcon b) \qcon c) &= \kappa \compL (a \qcon (b \qcon c)),
\\
\kappa \compL (a \qcon b) &= \kappa \compL (b \qcon a),
\\
\kappa \compL (\omicron_C \qcon a) &= \kappa \compL a,
\end{align}
due to the associativity, commutativity, and unitality of the conjunction, but the quantitative operation $\qcon$ is not required to satisfy these properties, i.e., it is possible that
\begin{align}
(a \qcon b) \qcon c & \neq a \qcon (b \qcon c),
\\
a \qcon b & \neq b \qcon a,
\\
\omicron_C \qcon a & \neq a.
\end{align}
However, in the following examples, we mainly consider quantitative operations $\qcon$ such that these properties are satisfied.
In such cases, $(\Hom_\cE(C, \Psi), \qcon, \omicron_C)$ forms a commutative monoid, and consequently $\Hom_\cE(C, \kappa)$ is a monoid homomorphism from it to $(\Hom_\cE(C, \Omega), \lcon, \lT_C)$.

In \cref{ssec:preorder}, we introduced preorder structures on the predicates $\Hom_\cE(C, \Omega)$ and the quantities $\Hom_\cE(C, \Psi)$.
The conjunction $\lcon$ is a commutative monoidal structure compatible with the preorder $(\Hom_\cE(C, \Omega), \preceq_C^\lT)$ because it is the meet operation.
For the set $\Hom_\cE(C, \Psi)$ of quantities, we can choose the quantitative operation $\qcon$ to be the meet operation as well.
Alternatively, we can only require it to be compatible with the preorder, in the sense that the monoid product is order-preserving: for any quantities $q_1, q_1', q_2, q_2' \in \Hom_\cE(C, \Psi)$, if $q_1 \preceq_C^\omicron q_1'$ and $q_2 \preceq_C^\omicron q_2'$, then $q_1 \qcon q_2 \preceq_C^\omicron q_1' \qcon q_2'$.
In other words, we require $(\Hom_\cE(C, \Psi), \preceq_C^\omicron, \qcon, \omicron_C)$ to be a \emph{symmetric monoidal preorder} \citep[Definition~2.2]{fong2019invitation}.

\begin{example}
Let us consider two commutative monoidal structures on the preorder $(\quant, \geq)$.
The max operation $\max: \quant \times \quant \to \quant$ is the meet, i.e., cartesian product, while the addition $+: \quant \times \quant \to \quant$ is a monoidal product.
They are both semicartesian because the top $0$ is the unit.
\end{example}

Note that $(\quant, +, 0)$, $(\inter, \times, 1)$, and $([1, \infty], \times, 1)$ are isomorphic to each other with the following isomorphisms:
\begin{equation}
\begin{tikzcd}[column sep=-1em, row sep=4em]
&
(\quant, +, 0)
\arrow[ld, "\exp(-x)"' sloped, shift left]
\arrow[rd, "\exp(x)" sloped, shift left]
\\
(\inter, \times, 1)
\arrow[rr, "\frac1x", shift left]
\arrow[ru, "-\log(x)" sloped, shift left]
&&
([1, \infty], \times, 1)
\arrow[ll, "\frac1x", shift left]
\arrow[lu, "\log(x)"' sloped, shift left]
\end{tikzcd}
\end{equation}
We can also induce a monoidal structure on a set if it is isomorphic to a monoid:
\begin{lemma}
\label{lem:induced_monoid}
Let $f: A \toot B :g$ be a pair of bijections between sets $A$ and $B$.
If $(B, \otimes_B, I_B)$ is a monoid, then $(A, \otimes_A \defeq g \compL \otimes_B \compL (f \times f), I_A \defeq g \compL I_B)$ is also a monoid.
\end{lemma}

\begin{pf}
Associativity:
\begin{flalign}
& (a \otimes_A b) \otimes_A c &
\\
={}& g((f(a) \otimes_B f(b)) \otimes_B f(c))
\\
={}& g(f(a) \otimes_B (f(b) \otimes_B f(c)))
\\
={}& a \otimes_A (b \otimes_A c)
\end{flalign}
Left unitality:
\begin{flalign}
& I_A \otimes_A a &
\\
={}& g(I_B) \otimes_A a
\\
={}& g(f(g(I_B)) \otimes_B f(a))
\\
={}& g(I_B \otimes_B f(a))
\\
={}& g(f(a))
\\
={}& a
\end{flalign}
Right unitality can be proven similarly.
\end{pf}

In this way, we can obtain a richer choice of monoidal structures on the set $\quant$ beyond the addition (see \cref{fig:conjunction}):
\begin{example}[Semicartesian monoidal product]
\small
\begin{align}
e^x - 1: (\quant, \qcon, 0) & \toot (\quant, +, 0) : \log(1 + x) &&
a \qcon b \defeq \log(e^a + e^b - 1)
\\
x^2: (\quant, \qcon, 0) & \toot (\quant, +, 0) : \sqrt{x} &&
a \qcon b \defeq \sqrt{a^2 + b^2}
\\
\sqrt{x}: (\quant, \qcon, 0) & \toot (\quant, +, 0) : x^2 &&
a \qcon b \defeq a + b + 2\sqrt{ab}
\\
x + 1: (\quant, \qcon, 0) & \toot ([1, \infty], \times , 1) : x - 1 &&
a \qcon b \defeq a + b + ab
\end{align}
\end{example}

In summary, the max operation on $\quant$ can be regarded as a \emph{continuous conjunction}, whereas a semicartesian monoidal product, such as the addition, can be viewed as a \emph{soft max}.


\subsection{Disjunction}
\label{ssec:disjunction}

Dually, the \emph{disjunction} $\ldis: \Omega \times \Omega \to \Omega$ reflects the join of the inclusion preorder of subobjects.
Similarly, we want to find a quantitative operation $\qdis: \Psi \times \Psi \to \Psi$ homomorphic to the disjunction $\ldis$ via the quantizer $\kappa$:
\begin{equation}
\begin{tikzcd}
\Psi \times \Psi
\arrow[r, "\qdis"]
\arrow[d, "\kappa \times \kappa"']
&
\Psi
\arrow[d, "\kappa"]
\\
\Omega \times \Omega
\arrow[r, "\ldis"']
&
\Omega
\end{tikzcd}
\end{equation}

Using the same technique as in \cref{lem:induced_monoid}, we can obtain several monoidal structures on the set $\quant$ homomorphic to the disjunction (see \cref{fig:disjunction}):
\begin{example}[Semicocartesian monoidal product]
\small
\begin{align}
\frac1x: (\quant, \qdis, \infty) & \toot (\quant, +, 0) : \frac1x &&
a \qdis b \defeq \frac{ab}{a + b}
\\[1ex]
\frac1{x^2}: (\quant, \qdis, \infty) & \toot (\quant, +, 0) : \frac1{\sqrt{x}} &&
a \qdis b \defeq \frac{ab}{\sqrt{a^2 + b^2}}
\\[1ex]
1 - e^{-x}: (\quant, \qdis, \infty) & \toot (\inter, \times, 1) : -\log(1 - x) &&
a \qdis b \defeq -\log(e^{-a} + e^{-b} - e^{-(a + b)})
\\[1ex]
\tanh: (\quant, \qdis, \infty) & \toot (\inter, \times, 1) : \arctanh &&
a \qdis b \defeq \arctanh(\tanh(a)\tanh(b))
\end{align}
\end{example}

However, we usually choose the quantitative operation $\qdis$ to be the join operation $\min$ of the preorder $(\quant, \geq)$.
In this way, $\qcon$ distributes over $\qdis$, i.e., for any quantities $a, b, c: C \to \Psi$, we have
\begin{equation}
a \qcon (b \qdis c) = (a \qcon b) \qdis (a \qcon c).
\end{equation}

A typical example is the min-plus semiring $(\quant, \min, +)$ \citep{pin1998tropical}.


\subsection{Implication}
\label{ssec:implication}

Finally, let us construct a quantitative counterpart of the \emph{implication} $\limp: \Omega \times \Omega \to \Omega$, which is right adjoint to the conjunction $\lcon$.
For global elements $a, b, c: 1 \to \Omega$, this means that
\begin{equation}
c \lcon a \lproves b \text{ if and only if } c \lproves a \limp b.
\end{equation}

There are two ways to construct a quantitative operation $\qimp: \Psi \times \Psi \to \Psi$ corresponding to the implication $\limp$.
One way is to find a quantitative operation $\qimp$ homomorphic to the implication $\limp$ via the quantizer $\kappa$:
\begin{equation}
\begin{tikzcd}
\Psi \times \Psi
\arrow[r, "\qimp"]
\arrow[d, "\kappa \times \kappa"']
&
\Psi
\arrow[d, "\kappa"]
\\
\Omega \times \Omega
\arrow[r, "\limp"']
&
\Omega
\end{tikzcd}
\end{equation}

\begin{example}
\label{eg:implication_homo}
For the set $\quant$, a quantitative operation $\qimp: \quant \times \quant \to \quant$ homomorphic to the implication $\limp: \truth \times \truth \to \truth$ is a function
\begin{equation}
(a, b)
\mapsto
[a = 0] \times [b > 0] \times f(b)
=
\begin{cases}
f(b) & a = 0 \text{ and } b > 0,\\
0 & \text{otherwise},\\
\end{cases}
\end{equation}
where $f: (0, \infty] \to (0, \infty]$ is an arbitrary function to non-zero numbers.
Note that this function is discontinuous at the line $a = 0$ and $b > 0$ (cf. \cref{ssec:negation}).
\end{example}

The other way is to find a quantitative operation $\qimp$ right adjoint to an operation $\qcon$ homomorphic to the conjunction $\lcon$, i.e., the \emph{internal hom} of the \emph{monoidal closed preorder} \citep[Definition~2.79]{fong2019invitation}.

\begin{example}
\label{eg:implication_max}
For the meet-semilattice $(\quant, \geq, \max, 0)$, the function
\begin{equation}
(a, b)
\mapsto
[a < b] \times b
=
\begin{cases}
0 & a \geq b,\\
b & a < b,\\
\end{cases}
\end{equation}
is right adjoint to the max because
\begin{equation}
\max\set{c, a} \geq b
\text{ if and only if }
c \geq
\begin{cases}
0 & a \geq b,\\
b & a < b.\\
\end{cases}
\end{equation}
While this function is subhomomorphic to the implication, it is still discontinuous at the line $a = b$.
\end{example}

\begin{example}
\label{eg:implication_sum}
For the monoidal preorder $(\quant, \geq, +, 0)$, the truncated subtraction $\monus$ (a.k.a.~\emph{monus} \citep{amer1984equationally})
\begin{equation}
b \monus a
\defeq
\max\set{b - a, 0}
=
\begin{cases}
0 & a \geq b,\\
b - a & a < b,\\
\end{cases}
\end{equation}
is right adjoint to the addition because
\begin{equation}
c + a \geq b
\text{ if and only if }
c \geq b \monus a.
\end{equation}
The truncated subtraction is continuous and subhomomorphic to the implication.
Note that the hinge function $n \mapsto \max\set{1 - n, 0} = 1 \monus n$ for the negation can be interpreted as the quantitative operation derived from $\lnot n = n \limp \lF$, where $1$ is homomorphic to the constant false $\lF$.
\end{example}

Similarly to \cref{lem:induced_monoid}, if two symmetric monoidal preorders are isomorphic and one of them is closed, we can induce that the other is also closed:
\begin{lemma}
Let $f: A \toot B :g$ be a pair of isomorphisms between symmetric monoidal preorders $(A, \preceq_A, \otimes_A, I_A)$ and $(B, \preceq_B, \otimes_B, I_B)$.
If $(B, \preceq_B, \otimes_B, I_B, \ihom_B)$ is closed,
then $(A, \preceq_A, \otimes_A, I_A, \ihom_A \defeq g \compL {\ihom_B} \compL (f \times f))$ is also closed.
\end{lemma}
\begin{pf}
For all $a, b, c \in A$, we have
\begin{flalign}
& c \preceq_A (a \ihom_A b) &
\\
\equiv
& c \preceq_A g(f(a) \ihom_B f(b))
\\
\equiv
& f(c) \preceq_B f(a) \ihom_B f(b)
\\
\equiv
& f(c) \otimes_B f(a) \preceq_B f(b)
\\
\equiv
& c \otimes_A a \preceq_A b
\end{flalign}
This means that $\ihom_A$ is right adjoint to $\otimes_A$.
\end{pf}

In this way, we can find the internal homs corresponding to the monoidal products discussed in \cref{ssec:conjunction} (see \cref{fig:implication}).

As a side note, we can use the quantitative operations discussed above to define quantitative operations for other logical connectives.
For example, the logical equivalence $a \leqv b$ can be represented as $(a \limp b) \lcon (b \limp a)$ (bi-implication), $(\lnot a \ldis b) \lcon (\lnot b \ldis a)$ (conjunctive normal form (CNF)), or $(a \lcon b) \ldis (\lnot a \lcon \lnot b)$ (disjunctive normal form (DNF)), and its quantitative operations can be defined accordingly.
Some examples are shown in \cref{fig:equivalence}.


\subsection{Heyting algebra, quantale, and ordered semiring}

Now, having constructed the logical operations and their corresponding quantitative operations, we can compare the structures of the subobject classifier with those of a subobject quantifier.

It is known that the global elements of the subobject classifier with the logical operations form a Heyting algebra $(\Omega, \lproves, \lT, \lF, \lcon, \ldis, \limp)$:
\begin{definition}
A \emph{Heyting algebra} is a cartesian closed bounded lattice.
\end{definition}
In categorical terms, $\lT$ is the terminal object, $\lF$ is the initial object, $\lcon$ is the product, $\ldis$ is the coproduct, and $\limp$ is the exponential in the preorder $(\Omega, \lproves)$ (as a thin category).

We weakened the requirements to construct the algebraic structures of the subobject quantifier, which usually forms what is called a quantale $(\Psi, \preceq, 1, 0, \qcon, \qdis, \qimp)$ \citep{mulvey1986ampersand, dudzik2017quantales}:
\begin{definition}
A (unital) \emph{quantale} is a monoidal closed suplattice.
\end{definition}
This means that we can consider a preorder $(\Psi, \preceq)$ on the subobject quantifier as a thin category, where $1$ is the terminal object, $0$ is the initial object, $\qcon$ is a monoidal product and not necessarily the product, $\qdis$ is still the coproduct, and $\qimp$ is the internal hom right adjoint to the monoidal product $\qcon$.
An example is the Lawvere quantale $(\quant, \geq, 0, \infty, +, \min, \monus)$ \citep{lawvere1973metric, bacci2023propositional}.
Then, the quantizer $\kappa: \Psi \to \Omega$ is a homomorphism preserving some or all the structures.

Note that due to the adjoint functor theorem and the fact that left adjoints preserve colimits, the product distributes over the coproduct in a Heyting algebra, and the monoidal product distributes over the coproduct in a quantale.
We can further relax the requirement for $\qdis$ to be the coproduct and instead consider a monoidal product (\cref{ssec:disjunction}).
If we still require the distributivity for the two monoidal structures $\qcon$ and $\qdis$, the algebraic structure is an \emph{ordered semiring} \citep{fujii2023ordered}.
Further investigation is left for future work.


\subsection{Quantifier: algebra over the exponentiation endofunctor}

Up to this point, our focus has been on $n$-ary operations in propositional logic (\cref{ssec:operation}).
Next, we introduce the universal quantification $\forall$ and existential quantification $\exists$ used in predicate logic and their quantitative counterparts.

Externally, the universal quantification and the existential quantification are right and left adjoint to the pullback of projection, respectively \citep{lawvere1969adjointness}.
Internally, the universal quantifier $\forall_D: \Omega^D \to \Omega$ and existential quantifier $\exists_D: \Omega^D \to \Omega$ are given by morphisms from the power object $\Omega^D$ of an object $D$ to the subobject classifier $\Omega$.

Recall that the power object $\Omega^D$ is also an exponential object into the subobject classifier $\Omega$, so the universal quantifier $\forall_D$ and existential quantifier $\exists_D$ can also be viewed as algebras over the exponentiation endofunctor $(-)^D$ of exponentiation on the subobject classifier $\Omega$.
Then, we can consider algebras on the subobject quantifier $\Psi$ homomorphic to them, which serve as quantitative counterparts of these quantifiers.

Our main result is as follows (cf.~\cref{thm:operation}):
\begin{theorem}
\label{thm:quantifier}
Consider an elementary topos with a subobject classifier $\lT: 1 \mono \Omega$, a subobject quantifier $\omicron: 1 \mono \Psi$, and a quantizer $\kappa: \Psi \to \Omega$.

Let $p: C \times D \to \Omega$ be a predicate on a product, and let $q: C \times D \to \Psi$ be a quantity such that $p = \kappa \compL q$.
Let $\expt{p}: C \to \Omega^D$ and $\expt{q}: C \to \Psi^D$ be their exponential transposes.

Let $\beta_D: \Omega^D \to \Omega$ be a predicate, and let $\alpha_D: \Psi^D \to \Psi$ be a quantity homomorphic to $\beta_D$ via the quantizer $\kappa$, i.e., $\beta_D \compL \kappa^D = \kappa \compL \alpha_D$.

Let $b: B \mono C \defeq (\beta_D \compL \expt{p})^*\lT$ be the subobject classified by $\beta_D \compL \expt{p}$, and let $a: A \mono C \defeq (\alpha_D \compL \expt{q})^*\omicron$ be the subobject quantified by $\alpha_D \compL \expt{q}$.
\begin{equation}
\begin{tikzcd}[column sep=0, row sep=2em]
&[2em]
A
\arrow[rrrr]
\arrow[dd, "a"', monomorphism, pos=.3]
\arrow[rrdd, "\lrcorner", phantom, pos=.05]
&[2.5em]&[1em]&[3.5em]&[1.5em]
1
\arrow[dd, "\omicron", pos=.3]
\arrow[ld, leftrightarrow]
\\
B
\arrow[rrrr, crossing over]
\arrow[dd, "b"', monomorphism, pos=.3]
\arrow[rrdd, "\lrcorner", phantom, pos=.05]
&&&&
1
\\
&
C
\arrow[rr, "\expt{q}"']
\arrow[ld, "\id_C", leftrightarrow]
&&
\Psi^D
\arrow[rr, "\alpha_D"']
\arrow[ld, "\kappa^D"]
&&
\Psi
\arrow[ld, "\kappa"]
\\
C
\arrow[rr, "\expt{p}"']
&&
\Omega^D
\arrow[rr, "\beta_D"']
&&
\Omega
\arrow[from=uu, "\lT", pos=.3, crossing over]
\end{tikzcd}
\end{equation}
Then, we have
\begin{enumerate}[label=(\roman*)]
\item \label{item:exponential_quantization}
$\kappa^D \compL \expt{q} = \expt{p}$
\item \label{item:exponential_predicate_quantity}
$\kappa \compL (\alpha_D \compL \expt{q}) = \beta_D \compL \expt{p}$
\item \label{item:classified}
$(\beta_D \compL \expt{p}) \compL a = \lT_A$
\item \label{item:quantified}
$(\alpha_D \compL \expt{q}) \compL b = \omicron_B$
\end{enumerate}
\end{theorem}
\begin{pf}
\ref{item:exponential_quantization} follows from the property of exponential.

\ref{item:exponential_predicate_quantity} shows the relationship between the predicate $\beta_D \compL \expt{p}$ and the quantity $\alpha_D \compL \expt{q}$ when $\alpha_D$ is homomorphic to $\beta_D$.
\begin{flalign}
& \kappa \compL \alpha_D \compL \expt{q}
\\
={}& \beta_D \compL \kappa^D \compL \expt{q}
& \text{(homomorphism)}
\\
={}& \beta_D \compL \expt{p}
& \text{\ref{item:exponential_quantization}}
\end{flalign}

\ref{item:classified} means that $\beta_D \compL \expt{p}$ is a classifying morphism of $a$.
\begin{flalign}
& \beta_D \compL \expt{p} \compL a
\\
={}& \kappa \compL \alpha_D \compL \expt{q} \compL a
& \text{\ref{item:exponential_predicate_quantity}}
\\
={}& \kappa \compL \omicron_A 
& \text{(pullback)}
\\
={}& \lT_A 
& \text{(composition)}
\end{flalign}

\ref{item:quantified} means that $\alpha_D \compL \expt{q}$ is a quantifying morphism of $b$.
\begin{flalign}
& \kappa \compL \alpha_D \compL \expt{q} \compL b
\\
={}& \beta_D \compL \expt{p} \compL b
& \text{\ref{item:exponential_predicate_quantity}}
\\
={}& \lT_B
& \text{(pullback)}
\end{flalign}
$\alpha_D \compL \expt{q} \compL b = \omicron_B$ follows from \cref{lem:quantity}.
\end{pf}

\begin{definition}[Universal aggregator]
A \emph{universal aggregator} $\qforall_D: \Psi^D \to \Psi$ is a quantity that is homomorphic to the universal quantifier $\forall_D: \Omega^D \to \Omega$:
\begin{equation}
\begin{tikzcd}
\Psi^D
\arrow[r, "\qforall_D"]
\arrow[d, "\kappa^D"']
&
\Psi
\arrow[d, "\kappa"]
\\
\Omega^D
\arrow[r, "\forall_D"']
&
\Omega
\end{tikzcd}
\end{equation}
\end{definition}

\begin{definition}[Existential aggregator]
An \emph{existential aggregator} $\qexists_D: \Psi^D \to \Psi$ is a quantity that is homomorphic to the existential quantifier $\exists_D: \Omega^D \to \Omega$:
\begin{equation}
\begin{tikzcd}
\Psi^D
\arrow[r, "\qexists_D"]
\arrow[d, "\kappa^D"']
&
\Psi
\arrow[d, "\kappa"]
\\
\Omega^D
\arrow[r, "\exists_D"']
&
\Omega
\end{tikzcd}
\end{equation}
\end{definition}

\begin{example}
For the set $\quant$, the canonical choices of universal aggregator $\qforall_D$ and existential aggregator $\qexists_D$ are $\sup$ and $\inf$.
If the set $D$ is finite, we can also use $\fsum$ and $\mean$ as the universal aggregator.
Non-examples of universal aggregator include $\median$ and $\fmode$, which are not homomorphic to the universal quantifier.
\end{example}

Note that the universal quantifier and existential quantifier are commutative up to isomorphism:
\begin{align}
\forall_{A \times B} \iso \forall_B \compL \forall_A^B
&\iso
\forall_A \compL \forall_B^A \iso \forall_{B \times A},
\\
\exists_{A \times B} \iso \exists_B \compL \exists_A^B
&\iso
\exists_A \compL \exists_B^A \iso \exists_{B \times A}.
\end{align}
However, we are free to choose different aggregators for different objects that are not necessarily commutative.
For example, the sum of max is usually not equal to the max of sum.


\subsection{Enrichment}

Lastly, we describe the conversion based on enrichment.

First, let us define the enriching category:
\begin{definition}
Let $(\Psi, \preceq, \qcon, \qdis, \qimp)$ be an internal quantale object in $\cSet$.
We define $\Psi\mhyphen\cSet$ to be a category whose objects are tuples consisting of a set $C$, a $\Psi$-valued strict premetric $d_C$ on $C$, and universal and existential aggregators on $C$:
\begin{equation}
\textstyle
(C, d_C: C \times C \to \Psi, \qforall_C, \qexists_C: \Psi^C \to \Psi),
\end{equation}
and morphisms from $(A, d_A, \qforall_A, \qexists_A)$ to $(B, d_B, \qforall_B, \qexists_B)$ are functions $f: A \to B$.
\end{definition}

\begin{definition}
\label{def:psi_bifunctor}
A bifunctor $\boxtimes$ on $\Psi\mhyphen\cSet$ is given by the Cartesian product of sets and functions, together with the following products of strict premetrics and aggregators:
\begin{align}
d_A \boxtimes d_B:
& (A \times B) \times (A \times B) \iso (A \times A) \times (B \times B) \xto{d_A \times d_B} \Psi \times \Psi \xto{\otimes} \Psi.
\\
\textstyle
\qforall_A \boxtimes \qforall_B:
& \Psi^{A \times B} \iso (\Psi^A)^B \xto{\qforall_A^B} \Psi^B \xto{\qforall_B} \Psi,
\\
\textstyle
\qexists_A \boxtimes \qexists_B:
& \Psi^{A \times B} \iso (\Psi^A)^B \xto{\qexists_A^B} \Psi^B \xto{\qexists_B} \Psi.
\end{align}
\end{definition}

\begin{proposition}
The product $d_A \boxtimes d_B$ of strict premetrics given in \cref{def:psi_bifunctor} is again a strict premetric.
\end{proposition}

\begin{pf}
This is a result of \cref{thm:conjunction}.
Consider the following diagram:
\begin{equation}
\begin{tikzcd}[column sep=2em]
A \times B
\arrow[r, "\id_{A \times B}"]
\arrow[d, "\Delta_{A \times B}"']
\arrow[rd, "\lrcorner", phantom, very near start]
&[.5em]
A \times B
\arrow[r]
\arrow[d, "\Delta_A \times \Delta_B"']
\arrow[rd, "\lrcorner", phantom, very near start]
&[1.5em]
1
\arrow[r]
\arrow[d, "\angles{\omicron, \omicron}"']
\arrow[rd, "\lrcorner", phantom, very near start]
&
1
\arrow[d, "\omicron"]
\\
(A \times B)^2
\arrow[r, "\iso"']
&
A^2 \times B^2
\arrow[r, "d_A \times d_B"']
&
\Psi \times \Psi
\arrow[r, "\qcon"']
&
\Psi
\end{tikzcd}
\end{equation}
$\Delta_A \times \Delta_B$ is a pullback of $\omicron$ along $\qcon \compL (d_A \times d_B)$ because $\angles{\omicron, \omicron}$ is a pullback of $\omicron$ along $\qcon$ according to \cref{thm:conjunction}, $\Delta_A \times \Delta_B$ is a pullback of $\angles{\omicron, \omicron}$ along $d_A \times d_B$, and we can apply the pullback lemma.
$\Delta_A \times \Delta_B$ is isomorphic to $\Delta_{A \times B}$, which means that $\qcon \compL (d_A \times d_B)$ is a strict premetric.
\end{pf}

Based on this definition, we can show that
\begin{equation}
(d_A \boxtimes d_B) \boxtimes d_C
\iso
d_A \boxtimes (d_B \boxtimes d_C),
\end{equation}
because $\times$ and $\otimes$ are associative up to isomorphism.

Further, we have
\begin{align}
\textstyle
(\qforall_A \boxtimes \qforall_B) \boxtimes \qforall_C
&\iso
\textstyle
\qforall_A \boxtimes (\qforall_B \boxtimes \qforall_C),
\\
\textstyle
(\qexists_A \boxtimes \qexists_B) \boxtimes \qexists_C
&\iso
\textstyle
\qexists_A \boxtimes (\qexists_B \boxtimes \qexists_C),
\end{align}
because composition is associative.

The singleton $(\singleton, \omicron_{\singleton \times \singleton}, \id_\singleton, \id_\singleton)$ is the unit of the bifunctor $\boxtimes$.
In this way, $(\Psi\mhyphen\cSet, \boxtimes, \singleton)$ forms a monoidal category.

However, note that the bifunctor $\boxtimes$ is not symmetric because $\qforall_A \boxtimes \qforall_B$ and $\qexists_A \boxtimes \qexists_B$ are not necessarily symmetric, and $d_A \boxtimes d_B \iso d_B \boxtimes d_A$ if and only if the monoidal product $\otimes$ of the quantale $\Psi$ is commutative.

Since $\Psi\mhyphen\cSet$ is a monoidal category, we can consider a strict monoidal functor $F_\Psi: \cSet \to \Psi\mhyphen\cSet$, which equips products of sets with product strict premetrics and product aggregators:
\begin{align}
d_{A \times B} &\defeq d_A \boxtimes d_B,
\\
\textstyle
\qforall_{A \times B}
&\defeq
\textstyle
\qforall_A \boxtimes \qforall_B,
\\
\textstyle
\qexists_{A \times B}
&\defeq
\textstyle
\qexists_A \boxtimes \qexists_B.
\end{align}

Such a monoidal functor induces a functor from a ($\cSet$-enriched) category to a $\Psi\mhyphen\cSet$-enriched category, called the base change of enriching category.
This means that we have a systematic way to equip a set $[A, B]$ of morphisms with a strict premetric
\begin{equation}
d_{[A, B]}: [A, B] \times [A, B] \to \Psi,
\end{equation}
a universal aggregator
\begin{equation}
\textstyle
\qforall_{[A, B]}: \Psi^{[A, B]} \to \Psi,
\end{equation}
and an existential aggregator
\begin{equation}
\textstyle
\qexists_{[A, B]}: \Psi^{[A, B]} \to \Psi,
\end{equation}
which is compatible with the product.

Then, \cref{thm:main} is a special case of this enrichment.
The relationship between predicates and quantities follows from \cref{thm:operation,thm:quantifier}.


\subsection{Summary}

Finally, to accommodate readers without a background in category theory, we present the instantiated definitions and theoretical results free of categorical terminology.
The non-categorical proofs are omitted.
We will be using the following functions:
\begin{itemize}
\item the \emph{zero predicate} $\zeta: \quant \to \truth \defeq x \mapsto (x = 0)$,
\item the \emph{product} $\zeta^n: \quant^n \to \truth^n: (q_1, \dots, q_n) \mapsto (q_1 = 0, \dots, q_n = 0)$, and
\item the \emph{postcomposition} $\zeta^A: \quant^A \to \truth^A \defeq q \mapsto \zeta \compL q$ of the zero predicate.
\end{itemize}

\begin{definition}[Quantity]
A quantity $q: A \to \quant$ is \emph{homomorphic} to a predicate $p: A \to \truth$ if $p = \zeta \compL q$:
\begin{equation}
\forall a \in A.\;
(q(a) = 0) \leqv p(a).
\end{equation}
A quantity $q$ is \emph{subhomomorphic} to a predicate $p$ if $\zeta \compL q \limp p$:
\begin{equation}
\forall a \in A.\;
(q(a) = 0) \limp p(a).
\end{equation}
\end{definition}

\begin{example}
A \emph{strict premetric} $d_A: A \times A \to \quant$ is a quantity on the product set $A \times A$ homomorphic to the equality predicate $=_A: A \times A \to \truth$.
\end{example}

\begin{definition}[Quantitative operation]
Let $n \in \N$ be a natural number.
A quantitative operation $\alpha: \quant^n \to \quant$ is \emph{homomorphic} to a logical operation $\beta: \truth^n \to \truth$ via the zero predicate $\zeta$ if $\zeta \compL \alpha = \beta \compL \zeta^n$:
\begin{equation}
\forall (q_1, \dots, q_n) \in \quant^n.\;
(\alpha(q_1, \dots, q_n) = 0) \leqv \beta(q_1 = 0, \dots, q_n = 0).
\end{equation}
A quantitative operation $\alpha$ is \emph{subhomomorphic} to a logical operation $\beta$ via the zero predicate if $\zeta \compL \alpha \limp \beta \compL \zeta^n$:
\begin{equation}
\forall (q_1, \dots, q_n) \in \quant^n.\;
(\alpha(q_1, \dots, q_n) = 0) \limp \beta(q_1 = 0, \dots, q_n = 0).
\end{equation}
\end{definition}

The relationship between quantitative operations and logical operations is as follows (\cref{thm:operation}):
\begin{proposition}
Let $n \in \N$ be a natural number.
For $i \in \set{1, \dots, n}$, let $p_i: A \to \truth$ be a predicate, and let $q_i: A \to \quant$ be a quantity.
Let $p: A \to \truth^n \defeq \angles{p_1, \dots, p_n}$ be the tupling of the predicates, and let $q: A \to \quant^n \defeq \angles{q_1, \dots, q_n}$ be the tupling of the quantities.
Let $\alpha: \quant^n \to \quant$ be a quantitative operation, and let $\beta: \truth^n \to \truth$ be a logical operation.
Assume that for all $i \in \set{1, \dots, n}$, $q_i$ is homomorphic to $p_i$.
Then,
\begin{itemize}
\item if $\alpha$ is homomorphic to $\beta$, $\alpha \compL q$ homomorphic to $\beta \compL p$; and
\item if $\alpha$ is subhomomorphic to $\beta$, $\alpha \compL q$ subhomomorphic to $\beta \compL p$.
\end{itemize}
\end{proposition}

In fact, we can also show that if for all $i \in \set{1, \dots, n}$, $q_i$ is subhomomorphic to $p_i$, and $\alpha$ is subhomomorphic to $\beta$, then $\alpha \compL q$ is subhomomorphic to $\beta \compL p$.

\begin{definition}[Aggregator]
An aggregator $\alpha_A: \quant^A \to \quant$ is \emph{homomorphic} to a quantifier $\beta_A: \truth^A \to \truth$ via the zero predicate $\zeta$ if $\zeta \compL \alpha_A = \beta_A \compL \zeta^A$:
\begin{equation}
\forall q \in \quant^A.\;
\parens*{(\mathop{\alpha}_{a \in A} q(a)) = 0} \leqv \parens*{\mathop{\beta}_{a \in A} (q(a) = 0)}.
\end{equation}
\end{definition}

\begin{example}
A \emph{universal aggregator} $\qforall_A: \quant^A \to \quant$ is a function such that
\begin{equation}
\forall q \in \quant^A.\;
\parens*{(\qforall_{a \in A} q(a)) = 0} \leqv \parens*{\forall a \in A.\; q(a) = 0}.
\end{equation}
\end{example}

\begin{example}
An \emph{existential aggregator} $\qexists_A: \quant^A \to \quant$ is a function such that
\begin{equation}
\forall q \in \quant^A.\;
\parens*{(\qexists_{a \in A} q(a)) = 0} \leqv \parens*{\exists a \in A.\; q(a) = 0}.
\end{equation}
\end{example}

The relationship between aggregators and quantifiers is as follows (\cref{thm:quantifier}):
\begin{proposition}
Let $p: A \times B \to \truth$ be a predicate, and let $q: A \times B \to \quant$ be a quantity.
Let $\expt{p}: B \to \truth^A$ and $\expt{q}: B \to \quant^A$ be the exponential transposes of $p$ and $q$.
Let $\alpha_A: \quant^A \to \quant$ be an aggregator, and let $\beta_A: \truth^A \to \truth$ be a quantifier.
Then, if $q$ is homomorphic to $p$, and $\alpha$ is homomorphic to $\beta$, then $\alpha_A \compL \expt{q}$ is homomorphic to $\beta_A \compL \expt{p}$. 
\end{proposition}

We have a compositional way to assign a strict premetric and an aggregator to a product of sets:
\begin{proposition}
Let $d_A: A \times A \to \quant$ and $d_B: B \times B \to \quant$ be strict premetrics.
Then,
\begin{equation}
d_{A \times B}: (A \times B) \times (A \times B) \to \quant \defeq ((a, b), (a', b')) \mapsto d(a, a') + d(b, b')
\end{equation}
is a strict premetric on the product set $A \times B$.
\end{proposition}

\begin{proposition}
Let $\qforall_A: \quant^A \to \quant$ and $\qforall_B: \quant^B \to \quant$ be universal aggregators.
Then,
\begin{equation}
\qforall_{A \times B}: \quant^{A \times B} \to \quant \defeq q \mapsto \qforall_{b \in B} \qforall_{a \in A} q(a, b)
\end{equation}
is a universal aggregator on the product set $A \times B$.
\end{proposition}

\newpage
\section{Proofs}
\label{app:proofs}
\subsection[Product function]{\cref{prop:product}}

\begin{proof}
\begin{flalign}
&
q_\text{product}(m: Y \to Z) &
\\
={}&
\inf_{m_{1,1} \in [Y_1, Z_1]}
\inf_{m_{2,2} \in [Y_2, Z_2]}
d_{[Y, Z]}(m, m_{1,1} \times m_{2,2})
\\
={}&
\inf_{m_{1,1} \in [Y_1, Z_1]}
\inf_{m_{2,2} \in [Y_2, Z_2]}
(
d_{[Y, Z_1]}(m_1, m_{1,1} \compL p_1)
+
d_{[Y, Z_2]}(m_2, m_{2,2} \compL p_2)
)
\\
={}&
\inf_{m_{1,1} \in [Y_1, Z_1]}
d_{[Y, Z_1]}(m_1, m_{1,1} \compL p_1)
+
\inf_{m_{2,2} \in [Y_2, Z_2]}
d_{[Y, Z_2]}(m_2, m_{2,2} \compL p_2)
\\
={}&
\inf_{m_{1,1} \in [Y_1, Z_1]}
\qforall_{y \in Y}
d_{Z_1}(m_1(y), m_{1,1}(y_1))
+
\inf_{m_{2,2} \in [Y_2, Z_2]}
\qforall_{y \in Y}
d_{Z_2}(m_2(y), m_{2,2}(y_2))
\\
={}&
\nonumber
\inf_{m_{1,1} \in [Y_1, Z_1]}
\qforall_{y_1 \in Y_1}
\qforall_{y_2 \in Y_2}
d_{Z_1}(m_1(y_1, y_2), m_{1,1}(y_1))
\\
+&
\inf_{m_{2,2} \in [Y_2, Z_2]}
\qforall_{y_2 \in Y_2}
\qforall_{y_1 \in Y_1}
d_{Z_2}(m_2(y_1, y_2), m_{2,2}(y_2))
\\
\nonumber
={}&
\qforall_{y_1 \in Y_1}
\qforall_{y_2 \in Y_2}
d_{Z_1}(m_1(y_1, y_2), m_{1,1}^*(y_1))
\\
+&
\qforall_{y_2 \in Y_2}
\qforall_{y_1 \in Y_1}
d_{Z_2}(m_2(y_1, y_2), m_{2,2}^*(y_2))
,
\end{flalign}
where
\begin{flalign}
m_{1,1}^*
\defeq{}&
\arginf_{m_{1,1} \in [Y_1, Z_1]}
\qforall_{y_1 \in Y_1}
\qforall_{y_2 \in Y_2}
d_{Z_1}(m_1(y_1, y_2), m_{1,1}(y_1)) &
\\
={}&
y_1
\mapsto
\arginf_{z_1 \in Z_1}
\qforall_{y_2 \in Y_2}
d_{Z_1}(m_1(y_1, y_2), z_1),
\\
m_{2,2}^*
\defeq{}&
\arginf_{m_{2,2} \in [Y_2, Z_2]}
\qforall_{y_2 \in Y_2}
\qforall_{y_1 \in Y_1}
d_{Z_2}(m_2(y_1, y_2), m_{2,2}(y_2))
\\
={}&
y_2
\mapsto
\arginf_{z_2 \in Z_2}
\qforall_{y_1 \in Y_1}
d_{Z_2}(m_2(y_1, y_2), z_2)
.
\end{flalign}
\end{proof}

\subsection[Constant curried function]{\cref{prop:const_curry}}

\begin{proof}
\begin{flalign}
&
q_\text{const-curry}(m: Y \to Z) &
\\
={}&
q_\text{const}(\expt{m_1})
+
q_\text{const}(\expt{m_2})
\\
={}&
\qforall_{y_2 \in Y_2 \vphantom{y_2'}}
\qforall_{y_2' \in Y_2}
d_{[Y_1, Z_1]}(\expt{m_1}(y_2), \expt{m_1}(y_2'))
+
\qforall_{y_1 \in Y_1 \vphantom{y_1'}}
\qforall_{y_1' \in Y_1}
d_{[Y_2, Z_2]}(\expt{m_2}(y_1), \expt{m_2}(y_1'))
\\
\nonumber
={}&
\qforall_{y_2 \in Y_2 \vphantom{y_2'}}
\qforall_{y_2' \in Y_2}
\qforall_{y_1 \in Y_1 \vphantom{y_2'}}
d_{Z_1}(\expt{m_1}(y_2)(y_1), \expt{m_1}(y_2')(y_1))
\\
+&
\qforall_{y_1 \in Y_1 \vphantom{y_1'}}
\qforall_{y_1' \in Y_1}
\qforall_{y_2 \in Y_2 \vphantom{y_1'}}
d_{Z_2}(\expt{m_2}(y_1)(y_2), \expt{m_2}(y_1')(y_2))
\\
\nonumber
={}&
\qforall_{y_2 \in Y_2 \vphantom{y_2'}}
\qforall_{y_2' \in Y_2}
\qforall_{y_1 \in Y_1 \vphantom{y_2'}}
d_{Z_1}(m_1(y_1, y_2), m_1(y_1, y_2'))
\\
+&
\qforall_{y_1 \in Y_1 \vphantom{y_1'}}
\qforall_{y_1' \in Y_1}
\qforall_{y_2 \in Y_2 \vphantom{y_1'}}
d_{Z_2}(m_2(y_1, y_2), m_2(y_1', y_2))
\\
\nonumber
={}&
\qforall_{y_1 \in Y_1 \vphantom{y_2'}}
\qforall_{y_2 \in Y_2 \vphantom{y_2'}}
\qforall_{y_2' \in Y_2}
d_{Z_1}(m_1(y_1, y_2), m_1(y_1, y_2'))
\\
+&
\qforall_{y_2 \in Y_2 \vphantom{y_1'}}
\qforall_{y_1 \in Y_1 \vphantom{y_1'}}
\qforall_{y_1' \in Y_1}
d_{Z_2}(m_2(y_1, y_2), m_2(y_1', y_2))
.
\end{flalign}
\end{proof}

\newpage
\section{Discussions}
\label{app:discussions}
\subsection{Background}

Defining and measuring the properties of learning models is a core topic in machine learning, especially representation learning \citep{bengio2013representation}.
A proper comprehension of what constitutes good representations and how to assess their quality is important for developing suitable learning objectives and evaluation metrics.
To define these properties, many important concepts are given by \emph{equational predicates}, such as \emph{independence} of random variables, extensively used in statistical learning and causal learning \citep{hyvarinen2000independent, koller2009probabilistic, scholkopf2022statistical}, and \emph{equivariance} of learning models, reflecting the symmetries and structures of the data \citep{cohen2016group, zaheer2017deep, higgins2018towards, maron2019invariant, de2020natural, cohen2021equivariant, van2022multiagent, navon2023equivariant}.

Considerable efforts have been put into designing model architectures that perfectly satisfy specific properties, such as \emph{monotonicity} \citep{sill1997monotonic, daniels2010monotone}, \emph{invertibility} \citep{rezende2015variational, behrmann2019invertible, ishikawa2023universal}, \emph{convexity} \citep{amos2017input}, and \emph{equivariance} \citep{lee2019set, brehmer2023geometric}.
However, hard-coding multiple properties into a model by design could be challenging \citep{kohler2020equivariant}.
Hence, it is desirable to devise quantitative metrics to directly measure these properties, even if the models do not have the properties built-in \citep{goodfellow2009measuring, chen2020group, kvinge2022in}.
Ideally, these metrics should be easily computable or even differentiable, allowing us to directly optimize the properties.

\emph{Disentangled representation learning} \citep{bengio2013representation}, our main focus of this paper, is such a field where defining and measuring the desired properties are not straightforward tasks \citep{carbonneau2022measuring, zhang2023category}.
It has been suggested that disentangling the underlying explanatory factors in complex data is a promising approach for reliable, interpretable, generalizable, and data-efficient representation learning \citep{locatello2019fairness, locatello2019challenging, montero2021role, dittadi2021transfer, xu2022compositional}.
However, in contrast to the wealth of results regarding invariant and equivariant layers, the exploration of designing a \say{disentangled layer} has been relatively limited.
One reason is that disentanglement was not considered a singular property but rather a combination of several requirements.
The absence of a clear definition and appropriate metrics for disentanglement has created a gap between the learning objectives and evaluation metrics.
A new evaluation metric is often introduced along with a new representation learning method \citep{carbonneau2022measuring}, but it is usually unproven that the method can optimize the new metric, and the metric truly quantifies the alleged property \citep{higgins2017betavae, kim2018disentangling, chen2018isolating, li2020progressive}.

To formally define disentanglement, a line of research utilized group theory and representation theory \citep{cohen2014learning, cohen2015transformation, higgins2018towards}, with a focus on the \emph{direct product of groups}.
Thanks to the rich algebraic structure, it becomes possible to derive various model architectures and learning objectives from the equational requirements of the product and equivariance \citep{caselles2019symmetry, pfau2020disentangling, quessard2020learning, painter2020linear, miyato2022unsupervised, yang2022towards, tonnaer2022quantifying, keurti2023homomorphism}.
Another approach adopted a topological perspective, using concepts such as the \emph{product manifold} to define disentanglement \citep{zhou2020evaluating, fumero2021learning, zhang2021product, balabin2023disentanglement}.
However, theoretically comparing different approaches has been a challenging task.

To quantitatively measure disentanglement, \citet{ridgeway2018learning} proposed three concepts called \emph{modularity, compactness, and explicitness}, which were defined verbally but not mathematically.
\citet{eastwood2018framework} proposed similar three criteria called \emph{disentanglement, completeness, and informativeness} and corresponding evaluation metrics.
However, it was unclear what properties these metrics truly quantify.
Additionally, due to the necessity for additional training of classifiers along with hyperparameter tuning and the involvement of non-differentiable regressors such as the random forest \citep{breiman1984classification}, it is impossible to directly optimize these metrics using gradient-based optimization.
Recently, \citet{eastwood2023dcies} extended this framework with two new metrics called \emph{explicitness/ease-of-use and size} based on the functional capacity.
\citet{do2020theory} introduced metrics for \emph{informativeness, separability, independence, and interpretability} from an information-theoretic perspective, while \citet{tokui2022disentanglement} introduced a new metric in terms of \emph{uniqueness, redundancy, and synergy} based on partial information decomposition.
These metrics have been mainly used during the evaluation stage, after a model is trained with other learning objectives.


\subsection{Related work}

\paragraph{Equivariance}

The work by \citet{kvinge2022in} might be the closest to our approach in spirit.
They directly converted \emph{equivariance}, an equational predicate, to a quantitative metric and analyzed their relationship (Proposition~3.2).
In contrast, based on our proposed conversion method, we can use the following definition and metric:
\begin{definition}[Equivariant function]
Let $A$, $B$, and $C$ be sets.
A function $f: A \to B$ is \emph{equivariant} to actions (any binary functions) $\cdot_A: C \times A \to A$ and $\cdot_B: C \times B \to B$ if
\begin{align}
p_\text{equivariant}(f: A \to B)
\defeq{}&
\forall c \in C.\;
f \compL (c \cdot_A -) =_{[A, B]} (c \cdot_B -) \compL f
\\
={}&
\forall c \in C.\;
\forall a \in A.\;
f(c \cdot_A a) =_B c \cdot_B f(a),
\end{align}
which can be measured by
\begin{equation}
q_\text{equivariant}(f: A \to B)
\defeq
\qforall_{c \in C}
\qforall_{a \in A}
d_B(f(c \cdot_A a), c \cdot_B f(a)).
\end{equation}
\end{definition}

\paragraph{Calibration}

More broadly, the study of the relationship between different metrics in statistical learning is called \emph{calibration analysis} \citep{steinwart2007compare, reid2010composite, ni2019calibration, bao2020fractional, bao2020adversarial}.
Our work can be seen as an extension of the concept of the calibration to a wider range of properties defined by equational predicates.

\paragraph{Disentanglement metric}

In disentangled representation learning, metrics similar to \cref{eq:constant_currying_quantity} have been proposed by \citet{higgins2017betavae, kim2018disentangling}.
Their metrics also fix one factor and vary all others and calculate some constancy metrics (the mean pairwise distance in \citet{higgins2017betavae} and the variance in \citet{kim2018disentangling}).
However, both studies took an indirect approach, involving the training of a classifier to predict the fixed factor.
Consequently, the resulting metrics are not differentiable anymore and entangle modularity and informativeness.
In this work, we argue that it is better to measure these two properties separately.

\paragraph{Weakly supervised disentanglement}

\citet{ridgeway2018learning} proposed and investigated the similarity supervision and argued that such supervision is easy to obtain via crowdsourcing.
\citet{shu2020weakly} further studied this type of supervision based on distribution matching and referred it as match pairing.
Other weaker forms of supervision were also investigated, such as the number of changed factors \citep{locatello2020weakly} or paired data with unknown intervention \citep{brehmer2022weakly}.
Given that our theory can establish connections between logical definitions and quantitative metrics, it holds promise for deriving disentanglement metrics for various types of weak supervision based on logical inference.

\paragraph{Multi-valued logic}

Aristotelian logic assumes that every proposition is either true or false, adhering to the \emph{principle of bivalence}.
The law of Aristotelian logic can be algebraically represented on the set $\set{0, 1}$ of binary truth values \citep{boole1854investigation}, known as the two-element \emph{Boolean algebra}.
The exploration of non-Aristotelian logic involves investigating logical systems that relax or modify this strict binary valuation, allowing for a broader range of truth values and accommodating various forms of uncertainty, vagueness, or context-dependence in reasoning \citep{hajek1998metamathematics, malinowski2007many, bergmann2008introduction}.

The mathematical study of multi-valued logic can date back to the seminal work by \citeauthor{lukasiewicz1920logice} in \citeyear{lukasiewicz1920logice}, who introduced a third truth value interpreted as \say{possibility} and symbolized by $\frac12$.
\citet{lukasiewicz1920logice} examined several principles in this \emph{three-valued logic} such as the principles of identity, implication, syllogism, and contradiction, and discussed its theoretical and practical importance in indeterministic philosophy and deductive sciences.
Later, \citet{lukasiewicz1930untersuchungen} proposed \emph{propositional calculus}, a theory of propositions with values from the real interval $\inter$, which is now also commonly known as \emph{Łukasiewicz logic}.
Łukasiewicz logic involves new continuous logical connectives such as strong/weak conjunction and disjunction.

Furthermore, \citet{chang1958algebraic} studied the algebraic systems for \emph{many-valued logic}, called \emph{MV-algebras}.
\citet{chang1966continuous} then proposed \emph{continuous model theory}, also referred to as \emph{compact-valued logic} \citep{yaacov2022expressive}, where the truth values can be in arbitrary compact Hausdorff spaces and a wide variety of quantifiers was studied.
Later, \citet{yaacov2008model} studied model theory for metric structures and proposed \emph{(real-valued) continuous first-order logic} \citep{yaacov2010continuous}, where the space of truth values is a closed, bounded interval of real numbers with the order topology (e.g., $\inter$), and suggested that we only need two canonical quantifiers $\sup$ and $\inf$.
From a categorical perspective, \citet{cho2020categorical} developed categorical semantics of metric spaces and continuous logic by introducing the notion of \emph{continuous subobject classifier}, and \citet{figueroa2022topos} studied a topos of continuous logic using the notion of \emph{hyperdoctrine}.

On the other hand, from a categorical perspective, \citet{lawvere1973metric} showed that a generalized metric space, also known as a \emph{Lawvere metric space}, is a category enriched over what is now commonly called the \emph{Lawvere quantale} $(\quant, \geq, +, 0)$, i.e., the set $\quant$ of extended non-negative real numbers equipped with addition $+$ as a (semicartesian) monoidal product and truncated subtraction $\monus$ as the internal hom.
In other words, a Lawvere metric space is a set $A$ equipped with a function $d: A \times A \to \quant$ such that for all $a \in A$, we have $0 \geq d(a, a)$ or $d(a, a) = 0$ (identity), which makes $d$ a \emph{premetric}, and for all $a, b, c \in A$, we have $d(b, c) + d(a, b) \geq d(a, c)$ (composition), which means that $d$ satisfies the \emph{triangle inequality}.

Recently, \citet{mardare2016quantitative} took an equational approach to \emph{quantitative algebraic reasoning}, which was later also referred to as \emph{quantitative equational logic} \citep{mardare2021fixed}, by introducing approximate equality predicates $=_\varepsilon$ indexed by rational numbers $\varepsilon$ (i.e., $a =_\varepsilon b$ if $a$ and $b$ are at most $\varepsilon$ apart), and suggested that this approach essentially involves working with \emph{enriched Lawvere theory}.
\citet{dagnino2022logical} provided a logical ground to quantitative reasoning in the categorical language of \emph{Lawvere's doctrines} by viewing distances as equality predicates in \emph{linear logic}.
\citet{bacci2023propositional} further studied the natural deduction systems of propositional logics for the Lawvere quantale and introduced what was later called \emph{affine Lawvere logic}, including Łukasiewicz logic and Ben Yaacov's continuous propositional logic.
\citet{bacci2024polynomial} extended affine Lawvere logic to \emph{polynomial Lawvere logic} by allowing multiplication as an extra logical connective.
These studies are on propositional logic and do not involve predicates and quantifiers.
Recently, \citet{capucci2024quantifiers} studied a spectrum of quantifiers in $\quant$-valued quantitative predicate logic.

In the context of machine learning, these relatively recently developed logics have yet to prove their practical importance.
While these innovative approaches often hold theoretical promise, they need to demonstrate tangible benefits in real-world applications.
Key areas where these new logics might eventually make an impact include neuro-symbolic reasoning and logic/probabilistic programming \citep{davilagarcez2002neural, manhaeve2018deepproblog, sen2022neuro, badreddine2022logic, fagin2024foundations}, which hold promise for integrating low-level perception with high-level reasoning, improving model interpretability, enhancing training efficiency, and enabling more robust decision-making processes.
However, widespread adoption and validation through practical use cases are necessary to establish their true value and effectiveness in the machine learning landscape.

Under this background, let us contextualize our proposed methodology for deriving $\quant$-valued quantitative metrics from logical definitions.
We highlight three characteristics of our framework:
\begin{itemize}
\item We allowed not only metrics, but \emph{strict premetrics}, such as the relative entropy (Kullback--Leibler divergence) \citep{kullback1951information, perrone2023markov} widely used in machine learning, as the real-valued counterparts for the equality predicates;
\item We focused on whether the metrics are \emph{zero or not} and the (differentiable) optimization of the derived metrics, because our main goal is to guarantee that the minimizers of the derived metrics satisfy the predicate;
\item We included a wide range of real-valued quantifiers (or \emph{aggregators} in our terms) beyond $\sup$ and $\inf$, such as mean and mean square, as long as they are homomorphic to the two-valued quantifiers, because the derived metrics may have nicer properties or even analytical solutions.
\end{itemize}


\clearpage
\subsection{Implication and equivalence}

\begin{figure}
\centering
\begin{subfigure}[t]{0.24\linewidth}
\centering
\includegraphics[height=3.6cm]{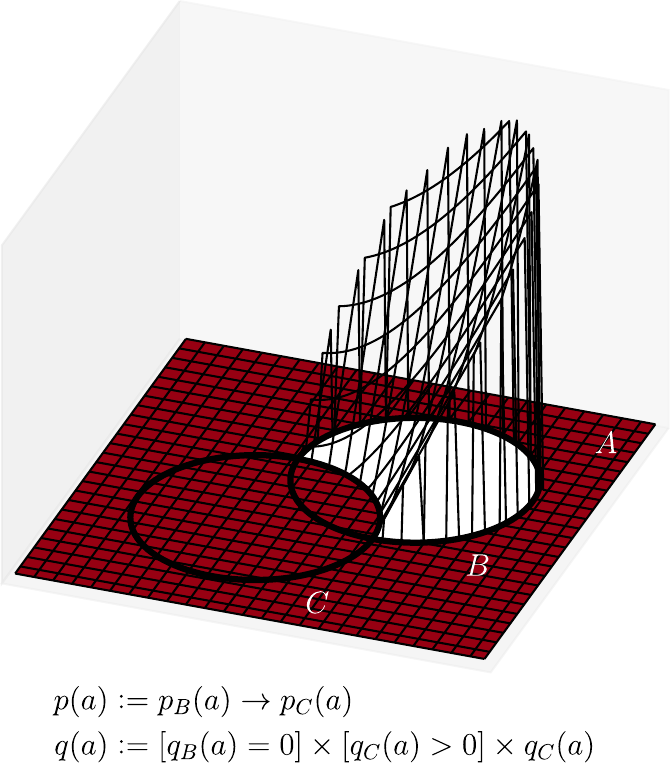}
\caption{homomorphic}
\end{subfigure}
\hfill
\begin{subfigure}[t]{0.24\linewidth}
\centering
\includegraphics[height=3.6cm]{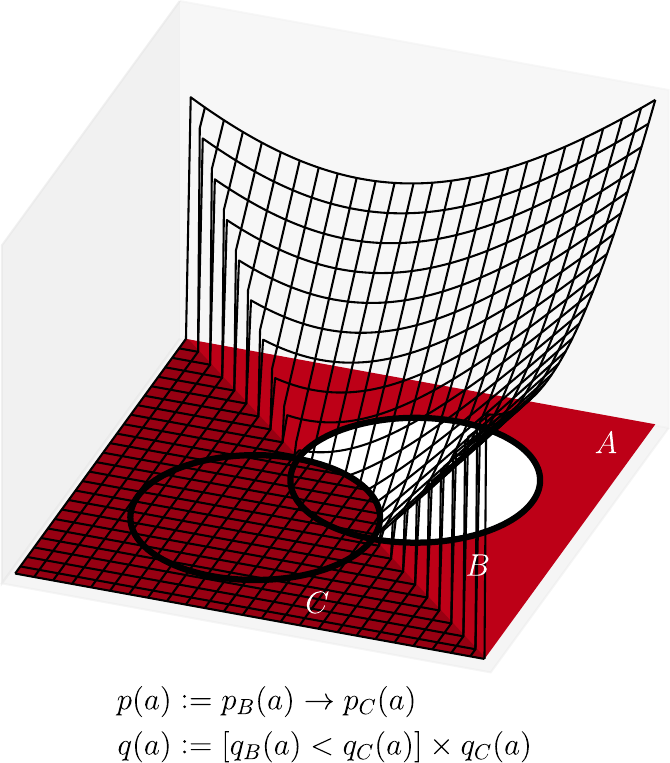}
\caption{adjoint to max}
\end{subfigure}
\hfill
\begin{subfigure}[t]{0.24\linewidth}
\centering
\includegraphics[height=3.6cm]{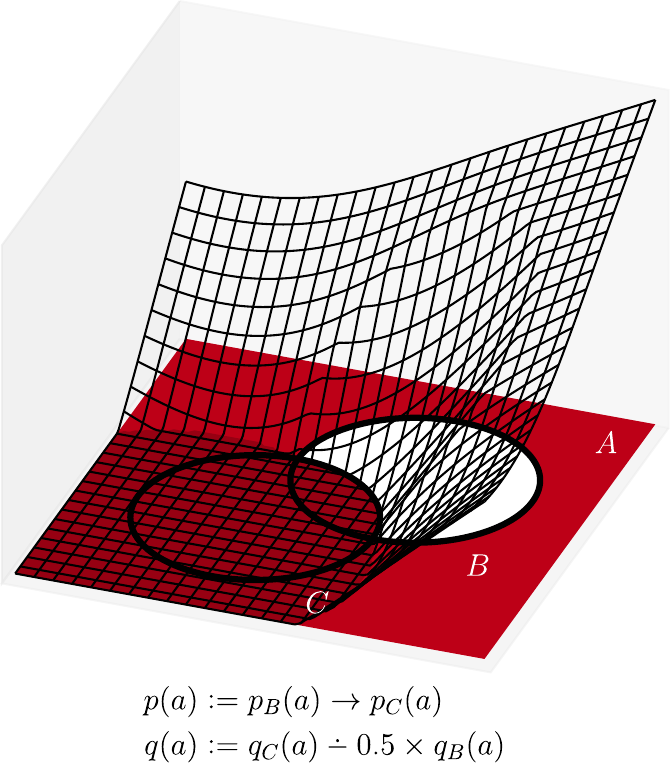}
\caption{less coverage}
\end{subfigure}
\hfill
\begin{subfigure}[t]{0.24\linewidth}
\centering
\includegraphics[height=3.6cm]{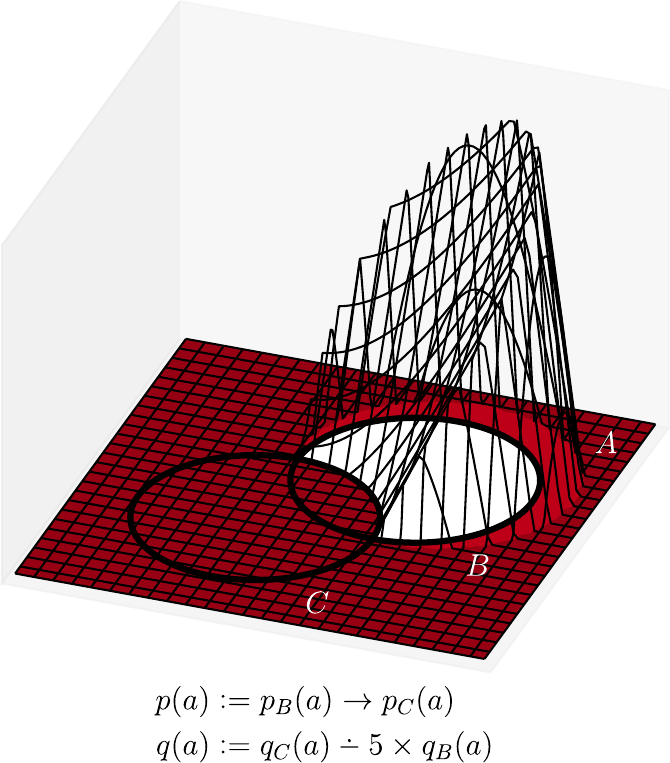}
\caption{more coverage}
\end{subfigure}

\begin{subfigure}[t]{0.24\linewidth}
\centering
\includegraphics[height=3.6cm]{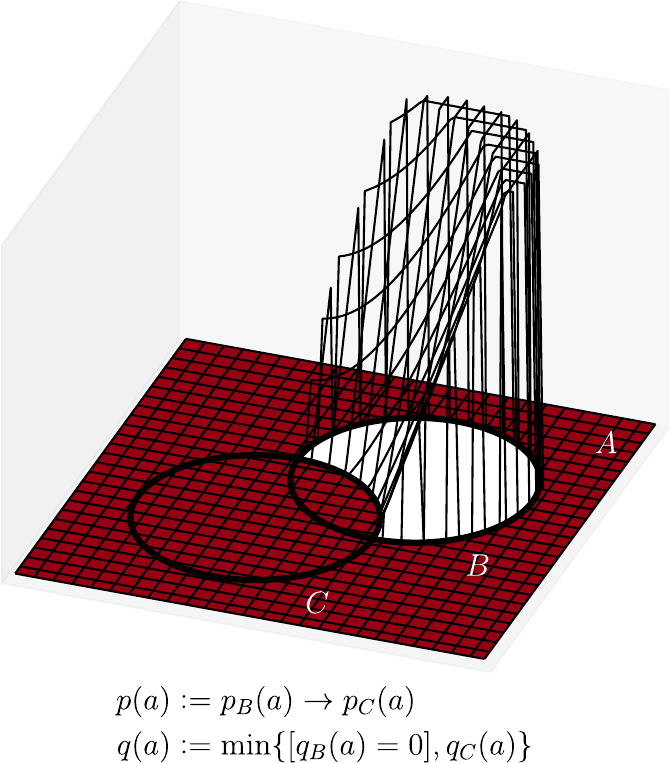}
\caption{negation, disjunction}
\end{subfigure}
\hfill
\begin{subfigure}[t]{0.24\linewidth}
\centering
\includegraphics[height=3.6cm]{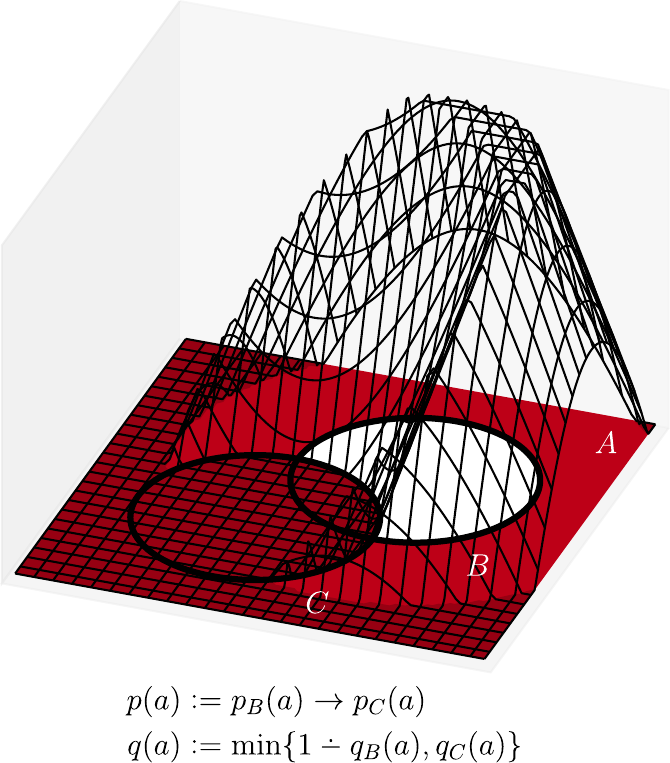}
\caption{min, hinge}
\end{subfigure}
\hfill
\begin{subfigure}[t]{0.24\linewidth}
\centering
\includegraphics[height=3.6cm]{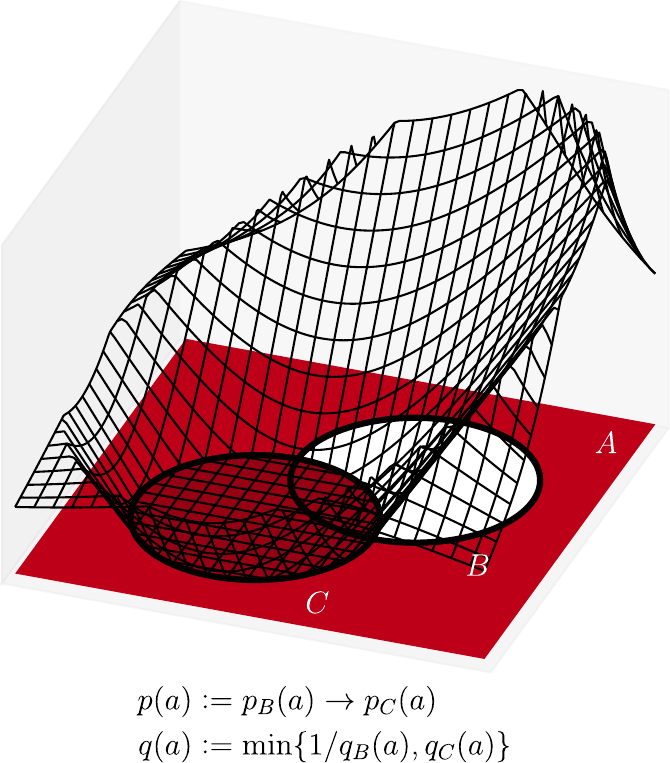}
\caption{min, reciprocal}
\end{subfigure}
\hfill
\begin{subfigure}[t]{0.24\linewidth}
\centering
\includegraphics[height=3.6cm]{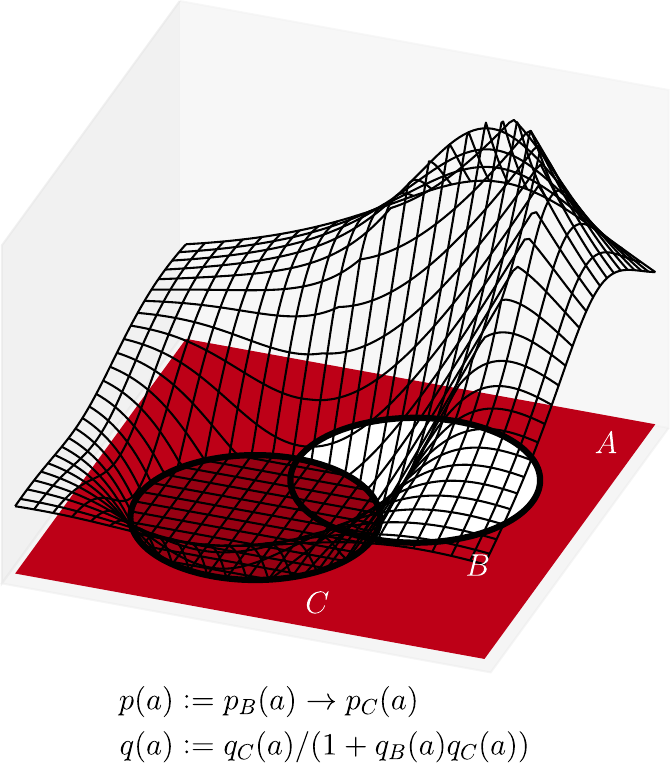}
\caption{fraction}
\end{subfigure}
\caption{Quantitative operations for implication}
\label{fig:qimp}
\end{figure}


\begin{figure}
\centering
\begin{subfigure}[t]{0.24\linewidth}
\centering
\includegraphics[height=3.6cm]{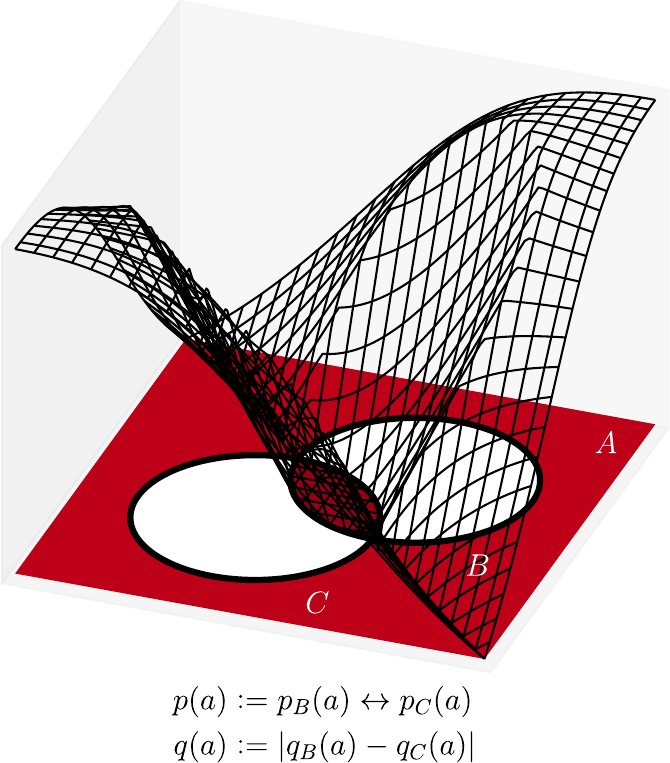}
\caption{bi-implication}
\end{subfigure}
\hfill
\begin{subfigure}[t]{0.24\linewidth}
\centering
\includegraphics[height=3.6cm]{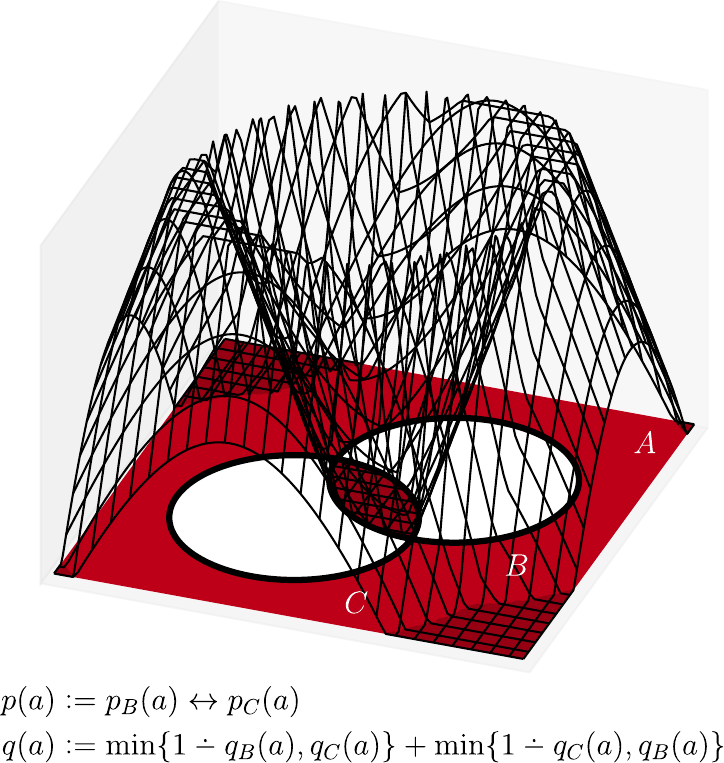}
\caption{CNF}
\end{subfigure}
\hfill
\begin{subfigure}[t]{0.24\linewidth}
\centering
\includegraphics[height=3.6cm]{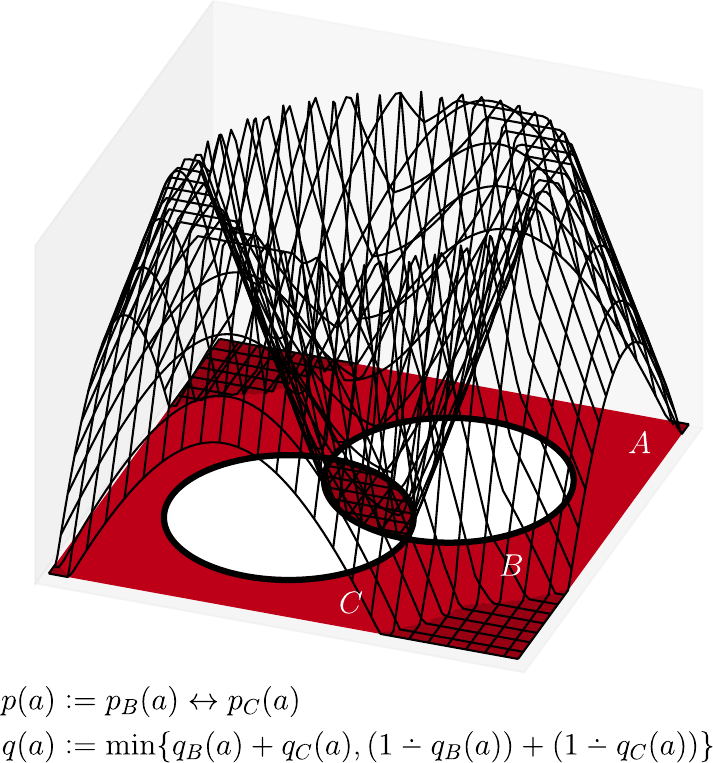}
\caption{DNF}
\end{subfigure}
\hfill
\begin{subfigure}[t]{0.24\linewidth}
\centering
\includegraphics[height=3.6cm]{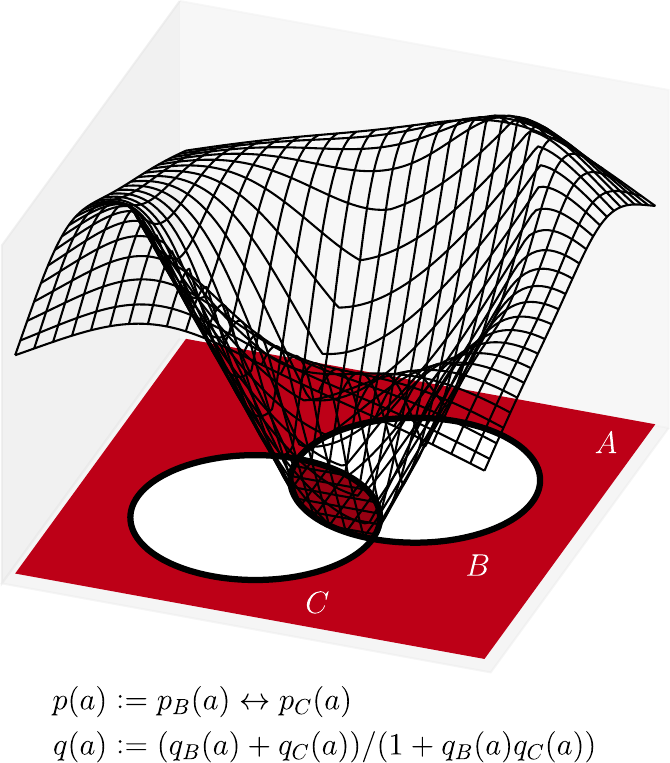}
\caption{fraction}
\end{subfigure}
\caption{Quantitative operations for equivalence}
\label{fig:qeqv}
\end{figure}


In \cref{sec:enrichment}, we only used the truncated subtraction $\monus$ as a quantitative operation for the implication $\limp$ (\cref{tab:conversion}).
In \cref{thm:main}, we noted that the implication is special because if a predicate involves the implication, then not all elements satisfying the predicate minimize the corresponding quantities (\cref{fig:venn}).
In \cref{ssec:implication}, we discussed other possible quantitative operations $\qimp$ corresponding to the implication $\limp$.

In \cref{fig:qimp}, we showed eight alternative quantitative operations for the implication.
In the first row, the first one is homomorphic to the implication, which means that its minimizers are exactly those that satisfy the predicate (\cref{eg:implication_homo});
the second one is right adjoint to the max (\cref{eg:implication_max});
and the other two are variants of the truncated subtraction (\cref{eg:implication_sum}), which have more or less coverage.
In the second row, we used the logical equivalence between $a \limp b$ and $\lnot a \ldis b$ to define quantitative operations for the implication using quantitative operations for the negation and disjunction.
For example, we can use the hinge function $1 \monus n$ and $\min$, which lead to $\min\set{1 \monus a, b}$, or the reciprocal function $\frac1n$ and $\frac{ab}{a + b}$, which lead to $\frac{\frac1a b}{\frac1a + b} = \frac{b}{1 + ab}$.

Similarly, we can use logically equivalent expressions of the logical equivalence $a \leqv b$, such as $(a \limp b) \lcon (b \limp a)$ (bi-implication), $(\lnot a \ldis b) \lcon (\lnot b \ldis a)$ (conjunctive normal form (CNF)), and $(a \lcon b) \ldis (\lnot a \lcon \lnot b)$ (disjunctive normal form (DNF)), to derive quantitative operations for the equivalence, shown in \cref{fig:qeqv}.

Note that a quantitative operation homomorphic to the implication cannot be continuous everywhere, which is undesirable for gradient-based optimization.
For example, the following quantity also measures the injectivity of a function $m: Y \to Z$:
\begin{align}
&
\qforall_{y \in Y \vphantom{y'}}
\qforall_{y' \in Y}
[d_Z(m(y), m(y')) = 0] \times [d_Y(y, y') > 0] \times d_Y(y, y')
\\
={}&
\qforall_{y \in Y \vphantom{y'}}
\qforall_{y' \in Y}
[m(y) =_Z m(y')] \times [y \neq_Y y'] \times d_Y(y, y').
\end{align}

This quantity aggregates distances between pairs of different inputs mapped to the same outputs.
However, unlike $q_\text{injective}$ introduced in \cref{ssec:informativeness}, it is not differentiable with respect to the function $m: Y \to Z$.
Thus, we cannot use it to improve the injectivity of a function by gradient descent.


\begin{figure}
\centering
\begin{subfigure}[t]{0.49\linewidth}
\centering
\includegraphics[width=\linewidth]{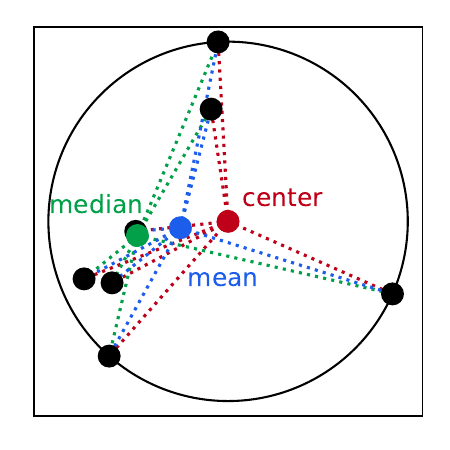}
\caption{central point}
\end{subfigure}
\hfill
\begin{subfigure}[t]{0.49\linewidth}
\centering
\includegraphics[width=\linewidth]{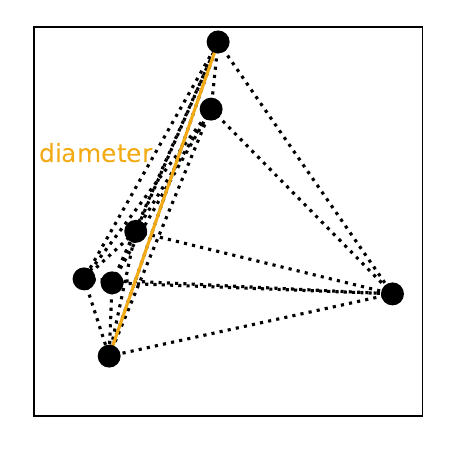}
\caption{pairwise distance}
\end{subfigure}
\caption[Constancy metrics]{%
Two approaches for measuring the \emph{constancy} of a set in $\R^2$:
(a) finding a central point, such as the center of the smallest bounding sphere, the geometric median, or the mean, and then measuring the dispersion around this point; and
(b) aggregating pairwise distances between points.
}
\label{fig:constancy}
\end{figure}


\subsection{Constant function}

Note that there are two logically equivalent definitions of a constant function (to a non-empty set).
One is based on the equality between all pairs, as in \cref{def:const}.
The other is based on the constant output value (see \cref{fig:constancy}):
\begin{definition}[Constant function with value]
A function $f: A \to B$ is a \emph{constant function} with value $b \in B$ if
\begin{equation}
p_\text{const-v}(f: A \to B)
\defeq
\exists b \in B.\;
\forall a \in A.\;
(f(a) =_B b),
\end{equation}
which can be measured by
\begin{equation}
q_\text{const-v}(f: A \to B)
\defeq
\inf_{b \in B}
\qforall_{a \in A}
d_B(f(a), b).
\end{equation}
\end{definition}

This quantity $q_\text{const-v}$ finds a central point in the codomain that best approximates all the outputs of a function, which is similar to the approach we discussed in \cref{ssec:product_approximation}.
In fact, we can prove that if we use $q_\text{const-v}$ in $q_\text{const-curry}$, we will end up with the same quantity $q_\text{product}$.


\begin{figure}
\centering
\begin{subfigure}[t]{0.49\linewidth}
\centering
\includegraphics[width=\linewidth]{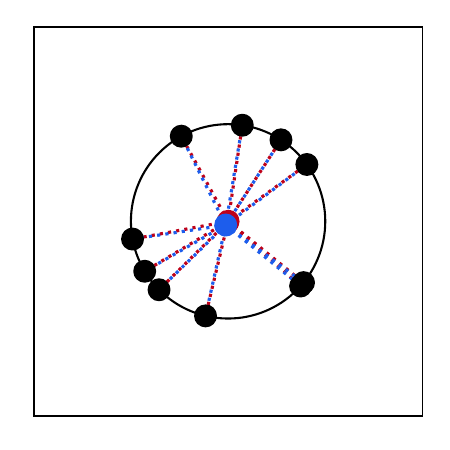}
\caption{small radius, large variance}
\end{subfigure}
\hfill
\begin{subfigure}[t]{0.49\linewidth}
\centering
\includegraphics[width=\linewidth]{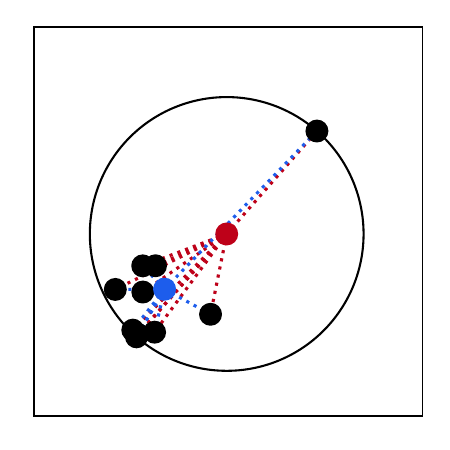}
\caption{large radius, small variance}
\end{subfigure}
\caption[Radius and variance]{Metrics may rank imperfect representations differently.}
\label{fig:radius_variance}
\end{figure}

\begin{figure}
\centering
\includegraphics[width=\linewidth]{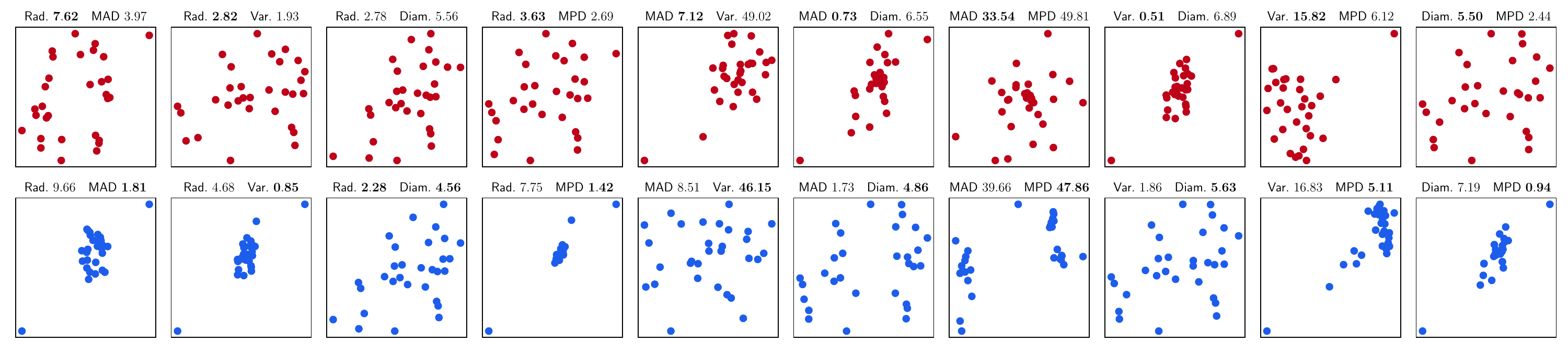}
\caption[Rank of imperfect representations]{%
For a pair of constancy metrics (each column), we can find two sets of points in $\R^2$ ranked differently by these metrics, except for the radius and diameter, because for a subset $A_0$ in a set $A$, we have
$
\inf_{a_0 \in A}   \sup_{a \in A_0} d_A(a_0, a)
\leq
\inf_{a_0 \in A_0} \sup_{a \in A_0} d_A(a_0, a)
\leq
\sup_{a_0 \in A_0} \sup_{a \in A_0} d_A(a_0, a)
$.}
\label{fig:rank}
\end{figure}


\newpage
\subsection{Rank of imperfect representations}
\label{ssec:rank}

It is worth noting that \cref{thm:main} only guarantees that the minimizers of different quantitative metrics derived from the same logical definition are the same, but imperfect representations, whose evaluation results are non-zero, may be ranked differently by different metrics.

For example, \cref{fig:radius_variance} illustrates two constancy metrics, the radius of the smallest bounding sphere and the variance, on two sets of points in $\R^2$, where one set has a small radius but a large variance, while the other has a large radius but a small variance.
More examples are presented in \cref{fig:rank}, and such results can also be observed in \cref{tab:main}.
This difference can lead to differences in risk preferences, sensitivity to outliers, and learning dynamics when these metrics are used as learning objectives.
Further investigation of the characteristics of these metrics for imperfect representations is left for future work.


\clearpage
\subsection{Implementation}
\label{ssec:implementation}

Thanks to advanced indexing (e.g., NumPy \citep{numpy} and PyTorch \citep{pytorch}) and analytical solutions to some optimization problems (e.g., \cref{eq:variance}), some of the proposed metrics can be easily implemented, even as Python one-liners.

For example, the following function implements a family of modularity metrics:

\begin{lstlisting}[language=Python, frame=single]
def q_product(y: np.ndarray, z: np.ndarray, aggregate, deviation):  
    return np.sum([aggregate([deviation(zi[yi == yv]) for yv in np.unique(yi)]) for yi, zi in zip(y, z)])
\end{lstlisting}

Here, \lstinline{y} and \lstinline{z} are NumPy arrays of shape \lstinline{(factor, index)};
\lstinline{aggregate} can be \lstinline{max}, \lstinline{mean}, or \lstinline{sum}; 
\lstinline{deviation} can be a function calculating the radius of the smallest bounding sphere,\footnote{\url{https://github.com/marmakoide/miniball} (MIT License) \citep{welzl1991smallest}} mean absolute deviation around the geometric median,\footnote{\url{https://github.com/krishnap25/geom_median} (GNU General Public License, Version 3 (GPLv3)) \citep{pillutla2022robust}} variance, diameter, or mean pairwise distance.
Please note, however, that the deviation function can be computationally expensive, depending on the dimension of the codes.


\subsection{Limitations}
\label{ssec:limitations}

Lastly, we discuss several aspects that are not covered in this work and potential directions for future research.

\paragraph{Function equality}

A collection of input-out pairs $\set{(x_i, y_i)}_{i=1}^n \in (X \times Y)^n$ may not define a \emph{function} $g: X \to Y$ for two reasons:
First, the set of all inputs $X_0 \defeq \set{x_i}_{i=0}^n$ is unlikely to enumerate all possible inputs (i.e., $X_0 \subsetneq X$), especially when the cardinality of the domain $X$ is infinite (e.g., $\R$), so the data may only define a \emph{partial function} $g: X \rightharpoonup Y$ or a function from a smaller domain $g_0: X_0 \to Y$.
Second, the inputs may not be distinct, e.g., when an input is given multiple labels by different annotators, so the data may define a \emph{multi-valued function}.
The extension from functions to relations or stochastic maps is an important future direction of our work.

\paragraph{Partial combinations}

A more general issue is learning and evaluating disentangled representations given only a subset of all combinations of factors, which is common when dealing with a large number of factors \citep{trauble2021disentangled, montero2021role, montero2022lost, roth2023disentanglement}.
It is crucial to evaluate and justify whether a metric computed on partial combinations of factors is a reliable proxy for the performance of the model on unseen combinations.

\paragraph{Unknown projections}

Another common scenario is when the extracted representation is not properly aligned with the underlying factors.
For example, a model may extract a three-dimensional representation $z \in \R^3$ for two factors $y \in \inter^2$, and it can project to $((z_1, z_2), z_3)$ or $(z_1, (z_2, z_3))$. 
How can we determine which is better, without enumerating all possible projections?
Finding the optimal assignment \citep{mahon2023correcting} and correcting a pre-trained model post hoc \citep{trauble2021disentangled} based on the proposed metrics are interesting future directions.


\subsection{Broader impact}
\label{ssec:impact}

This paper focuses on the theoretical aspects of disentangled representation learning, and we do not foresee any immediate negative societal consequences.
However, we would acknowledge that disentanglement is closely related to data-efficiency and fairness, potentially sparking discussions on ethical considerations.
Besides, the application of category theory may facilitate the transfer and integration of knowledge across disciplines, fostering closer connections between various fields of study, even beyond the machine learning community.

\newpage
\section{Experiments}
\label{app:experiments}
In this section, we provide the detailed data configuration used in \cref{sec:experiments} and further experimental results.


\subsection{Synthetic data}

\begin{figure}
\centering
\begin{subfigure}[t]{0.24\linewidth}
\centering
\includegraphics[width=\linewidth]{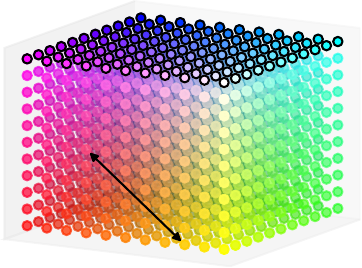}
\caption{true factors $Y$}
\label{sfig:factor}
\end{subfigure}
\hfill
\begin{subfigure}[t]{0.24\linewidth}
\centering
\includegraphics[width=\linewidth]{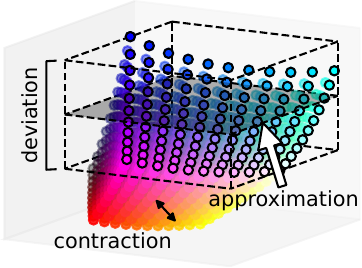}
\caption{entangled codes $Z$}
\label{sfig:code}
\end{subfigure}
\hfill
\begin{subfigure}[t]{0.24\linewidth}
\centering
\includegraphics[width=\linewidth]{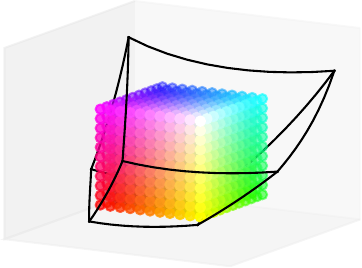}
\caption{product approximation of an encoder $m: Y \to Z$}
\label{sfig:product}
\end{subfigure}
\hfill
\begin{subfigure}[t]{0.24\linewidth}
\centering
\includegraphics[width=\linewidth]{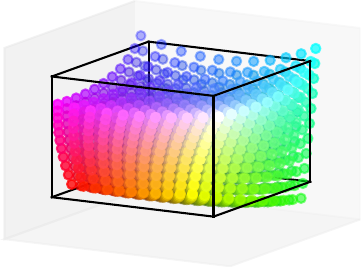}
\caption{linear approximation of its retraction $h: Z \to Y$}
\label{sfig:retraction}
\end{subfigure}
\caption[Illustration of synthetic data]{%
(a) a set of \emph{factors} $Y$ represented by the RGB color model;
(b) a set of entangled \emph{codes} $Z$ extracted by an encoder $m: Y \to Z$;
(c) a \emph{product} function approximation; and
(d) a linear approximation of the \emph{retraction} $h: Z \to Y$ of the encoder.
}
\label{fig:demo}
\end{figure}

We used a simple synthetic setup to simulate entanglement of factors and common failures patterns.
Concretely, we used a Cartesian product $Y \defeq \set{0, 0.1, \dots, 1}^3$ of three sets as the underlying factors (\cref{sfig:factor}).
We used a random rotation matrix $R$ to entangle factors and componentwise exponential as a non-linear transformation.
We composited this procedure twice and used an affine transformation to normalize the outputs (\cref{sfig:code}).
That is, we used the following function as the data generating process:
\begin{equation}
g: Y \to X: y \mapsto a \cdot \exp(R \cdot \exp(R \cdot y)) + b.
\end{equation}
Note that this function is injective but not a product or linear.

We used the following functions as the function $m: Y \to Z$:
\begin{equation*}
\begin{array}{lrcl}
\text{entanglement} & (y_1, y_2, y_3) & \mapsto & g(y_1, y_2, y_3) \\
\text{rotation}     & (y_1, y_2, y_3) & \mapsto & R \cdot (y_1, y_2, y_3) \\
\text{duplicate}    & (y_1, y_2, y_3) & \mapsto & ((y_1, y_2, y_3), (y_1, y_2, y_3), y_3) \\
\text{complement}   & (y_1, y_2, y_3) & \mapsto & ((y_2, y_3), (y_1, y_3), (y_1, y_2)) \\
\text{misalignment} & (y_1, y_2, y_3) & \mapsto & (y_2, y_3, y_1) \\
\text{redundancy}   & (y_1, y_2, y_3) & \mapsto & ((y_1, -y_1), y_2, y_3) \\
\text{contraction}  & (y_1, y_2, y_3) & \mapsto & 0.01 \times (y_1, y_2, y_3) \\
\text{nonlinear}    & (y_1, y_2, y_3) & \mapsto & (y_1^2, y_2^2, y_3^2) \\
\text{constant}     & (y_1, y_2, y_3) & \mapsto & (0, 0, 0) \\
\end{array}
\end{equation*}
The rotation operation entangles factors but can be (linearly) inverted.
The duplicate encoder has a modular decoder (projections), but itself is not modular.
The redundancy encoder is both modular and informative, but not all codes can be decoded.
The constant encoder is perfectly modular but not informative.


\begin{figure}
\centering
\includegraphics[width=\linewidth]{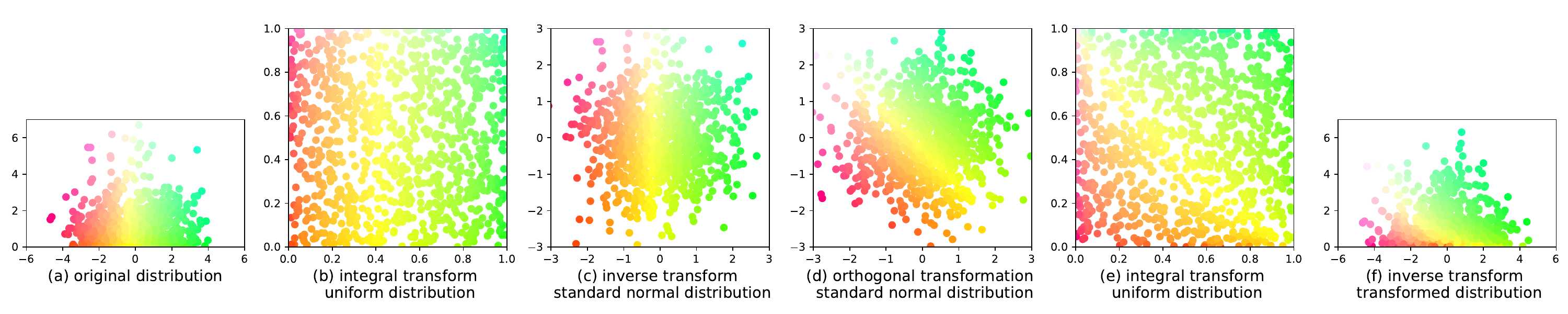}
\caption[Entanglement of a distribution]{Entangling a multivariate distribution via probability integral transform, inverse transform, and orthogonal transformation of standard normal distribution \citep[Theorem 1]{locatello2019challenging}.}
\label{fig:entangle}
\end{figure}


\subsection{Weakly supervised modularity metrics}

\begin{table}[t]
\centering
\caption{Supervised modularity metrics}
\label{tab:strong}
\begin{tabular}{l rrrrr}
\toprule
& \multicolumn{3}{c}{Product approx.}
& \multicolumn{2}{c}{Constancy}
\\
\cmidrule(lr){2-4}
\cmidrule(lr){5-6}
& Rad. & MAD & Var. & Diam. & MPD
\\
\midrule
entanglement & $0.44$ & $0.75$ & $0.96$ & $0.19$ & $0.82$ \\
rotation     & $0.22$ & $0.51$ & $0.80$ & $0.05$ & $0.64$ \\
duplicate    & $0.24$ & $0.43$ & $0.67$ & $0.06$ & $0.56$ \\
complement   & $0.12$ & $0.28$ & $0.55$ & $0.01$ & $0.42$ \\
misalignment & $0.22$ & $0.44$ & $0.74$ & $0.05$ & $0.58$ \\
random       & $0.22$ & $0.48$ & $0.78$ & $0.05$ & $0.61$ \\
\bottomrule
\end{tabular}
\end{table}
\begin{table}[t]
\centering
\caption{Weakly supervised modularity metrics}
\label{tab:weak}
\begin{tabular}{l rrrrr}
\toprule
& \multicolumn{3}{c}{Product approx.}
& \multicolumn{2}{c}{Constancy}
\\
\cmidrule(lr){2-4}
\cmidrule(lr){5-6}
& Rad. & MAD & Var. & Diam. & MPD
\\
\midrule
entanglement & $0.50$ & $0.77$ & $0.96$ & $0.26$ & $0.84$ \\
rotation     & $0.24$ & $0.54$ & $0.83$ & $0.06$ & $0.68$ \\
duplicate    & $0.28$ & $0.46$ & $0.71$ & $0.07$ & $0.60$ \\
complement   & $0.14$ & $0.32$ & $0.59$ & $0.02$ & $0.47$ \\
misalignment & $0.22$ & $0.48$ & $0.77$ & $0.05$ & $0.62$ \\
random       & $0.24$ & $0.52$ & $0.81$ & $0.06$ & $0.64$ \\
\bottomrule
\end{tabular}
\end{table}

We briefly comment on the possibility of employing weak supervision for measuring disentangled representations.

For supervised disentanglement metrics we discussed in \cref{sec:metrics}, the necessary data consists of observation-factor pairs $(x, y)$, representing a generator $g: Y \to X$.
In order to evaluate an encoder $f: X \to Z$, we compose it with a generator and study the properties of the composition $m: Y \to Z \defeq f \compL g$.

It is worth noting that $p_\text{product}$ and $p_\text{injective}$ are equational predicates, which means that they are \emph{invariant to bijections}.
Similarly, $q_\text{product}$ and $q_\text{injective}$ are \emph{invariant to isometries}.
This implies that the exact values of the factors are not important; we only need to know if two factors are equal or not.
Hence, we only need weak supervision of the form $(x, x', y_i =_{Y_i} y'_i)$ so that we can construct some equivalence classes of factors, and we can still calculate or approximate \cref{eq:variance,eq:diameter}.
Note that this type of supervision has been partially investigated by \citet{ridgeway2018learning, shu2020weakly}.

For example, suppose we have an object A (e.g., red circle) and an object B (e.g., red triangle), and all we know is that objects A and B have the same color.
Based on such weak information, we can still construct an equivalence class containing objects with the same color as object A.
Then, we can regularize an encoder $f: X \to Z$ by minimizing the variance of the color representations over this equivalence class (\cref{eq:variance}).
In this way, the modularity of the encoder can be improved.
The challenge arises when there is noise or only partial combinations, which is an interesting future work direction.
In such cases, we may need to use semi-supervised clustering to group the data \citep{wagstaff2001constrained, basu2002semi, bilenko2004integrating}.

To validate this idea, we conducted experiments where we only used a random sample of pairs and their similarities.
\cref{tab:strong} is an excerpt of \cref{tab:main}, showing only the proposed modularity metrics, and \cref{tab:weak} shows these metrics calculated using only similarity supervision.
We reported the mean values of $10$ random samples of pairs, and the variances are negligible.
Comparing \cref{tab:strong,tab:weak}, we can observe that weakly supervised metrics may overestimate imperfect representations, but they can still maintain the ranks.
This observation suggests the potential utility of employing weak supervision for both learning and evaluating disentangled representations using the proposed metrics.


\subsection{Evaluation of existing models on image datasets}

\begin{table}[t]
\centering
\caption{Supervised disentanglement metrics on image datasets}
\label{tab:vae}
\begin{adjustbox}{width=\linewidth}
\begin{tabular}{l rrrrr rrrrr rrrrrr}
\toprule
& \multicolumn{5}{c}{Modularity} & \multicolumn{5}{c}{Informativeness} & \multicolumn{6}{c}{Existing metrics}
\\
\cmidrule(lr){2-6}
\cmidrule(lr){7-11}
\cmidrule(lr){12-17}
& \multicolumn{3}{c}{Product approx.} & \multicolumn{2}{c}{Constancy}
& \multicolumn{3}{c}{Retraction approx.} & \multicolumn{2}{c}{Contraction}
& \multicolumn{2}{c}{Pair} & \multicolumn{1}{c}{Info.} & \multicolumn{3}{c}{Regressor}
\\
\cmidrule(lr){2-4}
\cmidrule(lr){5-6}
\cmidrule(lr){7-9}
\cmidrule(lr){10-11}
\cmidrule(lr){12-13}
\cmidrule(lr){14-14}
\cmidrule(lr){15-17}
& Rad. & MAD & Var. & Diam. & MPD 
& ME & MAE & MSE & Max & Mean
& Beta$^a$ & Factor$^b$ & MIG$^c$ & Dis.$^d$ & Com.$^d$ & Info.$^d$
\\
\midrule
\multicolumn{10}{l}{\cars}\\
\midrule
VAE
& $0.27$ & $0.76$ & $0.95$ & $0.07$ & $0.83$ & $0.44$ & $0.82$ & $0.94$ & $0.21$ & $0.75$
& $0.90$ & $0.22$ & $0.02$ & $0.07$ & $0.05$ & $0.54$ \\
$\beta$-VAE
& $0.26$ & $0.76$ & $0.95$ & $0.07$ & $0.82$ & $0.42$ & $0.82$ & $0.94$ & $0.20$ & $0.74$
& $0.90$ & $0.21$ & $0.01$ & $0.13$ & $0.10$ & $0.54$ \\
FactorVAE
& $0.24$ & $0.75$ & $0.95$ & $0.06$ & $0.82$ & $0.35$ & $0.82$ & $0.94$ & $0.21$ & $0.74$
& $0.89$ & $0.20$ & $0.03$ & $0.11$ & $0.08$ & $0.54$ \\
$\beta$-TCVAE
& $0.28$ & $0.77$ & $0.95$ & $0.08$ & $0.83$ & $0.43$ & $0.82$ & $0.94$ & $0.21$ & $0.74$
& $0.90$ & $0.21$ & $0.02$ & $0.14$ & $0.11$ & $0.59$ \\
\midrule
\multicolumn{10}{l}{\dsprites}\\
\midrule
VAE
& $0.24$ & $0.64$ & $0.94$ & $0.06$ & $0.74$ & $0.37$ & $0.82$ & $0.94$ & $0.18$ & $0.68$
& $0.54$ & $0.26$ & $0.09$ & $0.16$ & $0.15$ & $0.40$ \\
$\beta$-VAE
& $0.14$ & $0.64$ & $0.93$ & $0.02$ & $0.73$ & $0.41$ & $0.83$ & $0.94$ & $0.20$ & $0.70$
& $0.58$ & $0.29$ & $0.13$ & $0.20$ & $0.24$ & $0.44$ \\
FactorVAE
& $0.18$ & $0.63$ & $0.93$ & $0.03$ & $0.73$ & $0.38$ & $0.82$ & $0.94$ & $0.17$ & $0.67$
& $0.48$ & $0.26$ & $0.13$ & $0.20$ & $0.23$ & $0.36$ \\
$\beta$-TCVAE
& $0.22$ & $0.64$ & $0.93$ & $0.05$ & $0.73$ & $0.42$ & $0.83$ & $0.94$ & $0.21$ & $0.70$
& $0.56$ & $0.27$ & $0.18$ & $0.29$ & $0.31$ & $0.59$ \\
\midrule
\multicolumn{10}{l}{\shapes}\\
\midrule
VAE
& $0.25$ & $0.76$ & $0.96$ & $0.06$ & $0.82$ & $0.39$ & $0.88$ & $0.96$ & $0.20$ & $0.78$
& $0.99$ & $0.94$ & $0.19$ & $0.35$ & $0.29$ & $0.76$ \\
$\beta$-VAE
& $0.20$ & $0.72$ & $0.95$ & $0.04$ & $0.79$ & $0.39$ & $0.84$ & $0.94$ & $0.19$ & $0.69$
& $0.86$ & $0.77$ & $0.26$ & $0.69$ & $0.62$ & $0.99$ \\
FactorVAE
& $0.24$ & $0.69$ & $0.96$ & $0.06$ & $0.78$ & $0.34$ & $0.82$ & $0.93$ & $0.16$ & $0.64$
& $0.83$ & $0.57$ & $0.15$ & $0.40$ & $0.40$ & $0.84$ \\
$\beta$-TCVAE
& $0.25$ & $0.69$ & $0.94$ & $0.06$ & $0.77$ & $0.32$ & $0.82$ & $0.93$ & $0.18$ & $0.63$
& $0.76$ & $0.50$ & $0.11$ & $0.68$ & $0.57$ & $0.98$ \\
\midrule
\multicolumn{10}{l}{\mpi}\\
\midrule
VAE
& $0.04$ & $0.56$ & $0.91$ & $0.00$ & $0.66$ & $0.30$ & $0.76$ & $0.89$ & $0.12$ & $0.46$
& $0.48$ & $0.11$ & $0.12$ & $0.29$ & $0.33$ & $0.64$ \\
$\beta$-VAE
& $0.02$ & $0.84$ & $0.97$ & $0.00$ & $0.87$ & $0.28$ & $0.75$ & $0.89$ & $0.10$ & $0.41$
& $0.39$ & $0.11$ & $0.07$ & $0.15$ & $0.18$ & $0.47$ \\
FactorVAE
& $0.09$ & $0.60$ & $0.93$ & $0.01$ & $0.70$ & $0.27$ & $0.75$ & $0.89$ & $0.11$ & $0.43$
& $0.47$ & $0.09$ & $0.12$ & $0.26$ & $0.30$ & $0.60$ \\
$\beta$-TCVAE
& $0.07$ & $0.65$ & $0.95$ & $0.01$ & $0.74$ & $0.28$ & $0.76$ & $0.89$ & $0.12$ & $0.46$
& $0.45$ & $0.07$ & $0.12$ & $0.24$ & $0.31$ & $0.57$ \\
\bottomrule
\end{tabular}
\end{adjustbox}
\raggedright
\footnotesize{
$^a$ \citep{higgins2017betavae}
$^b$ \citep{kim2018disentangling}
$^c$ \citep{chen2018isolating}
$^d$ \citep{eastwood2018framework}
}
\end{table}

We also report the results of several widely used unsupervised disentangled representation learning methods (VAE \citep{kingma2014auto}, $\beta$-VAE \citep{higgins2017betavae}, FactorVAE \citep{kim2018disentangling}, and $\beta$-TCVAE \citep{chen2018isolating}) evaluated on four image datasets (\cars, \dsprites, \shapes, and \mpi) in \cref{tab:vae}.

We used a public PyTorch implementation \citep{pytorch} of these methods and used the same encoder/decoder architecture with the default hyperparameters described in \citet{locatello2019challenging} for all methods for a fair comparison.
We used linear projection to find the most informative representations for each factor.
The experiments were conducted on a NVIDIA Tesla V100 GPU.

Before analyzing these results, it is important to note that the evaluation of these learning models is \emph{not} meant to be a proof of the correctness of the proposed metrics, since we cannot tell whether a bad result is due to the insufficiency of a learning method, to the quality of the datasets, or to the problem of the evaluation, if we have no theoretical guarantee for the metrics.
We can trust the results of the proposed metrics because the properties of their minimizers are guaranteed by \cref{thm:main}.

From \cref{tab:vae} we can observe that the considered learning methods do not exhibit significant difference in terms of modularity and informativeness.
This result supports the theoretical finding of \citet{locatello2019challenging} that unsupervised learning of disentangled representations by matching the distributions of observations is fundamentally impossible (see also \cref{fig:entangle}) as well as their empirical finding that there is no evidence that learning disentangled representations in an unsupervised manner is reliable.


\subsection{Kendall tau distance between metrics}

\begin{table}[t]
\centering
\caption{Average Kendall tau rank distances bewteen disentanglement metrics}
\label{tab:kendalltau}
\begin{adjustbox}{width=\linewidth}
\begin{tabular}{l rrrrr rrrrr rrrrrr}
\toprule
& \multicolumn{5}{c}{Modularity} & \multicolumn{5}{c}{Informativeness} & \multicolumn{6}{c}{Existing metrics}
\\
\cmidrule(lr){2-6}
\cmidrule(lr){7-11}
\cmidrule(lr){12-17}
& \multicolumn{3}{c}{Product approx.} & \multicolumn{2}{c}{Constancy}
& \multicolumn{3}{c}{Retraction approx.} & \multicolumn{2}{c}{Contraction}
& \multicolumn{2}{c}{Pair} & \multicolumn{1}{c}{Info.} & \multicolumn{3}{c}{Regressor}
\\
\cmidrule(lr){2-4}
\cmidrule(lr){5-6}
\cmidrule(lr){7-9}
\cmidrule(lr){10-11}
\cmidrule(lr){12-13}
\cmidrule(lr){14-14}
\cmidrule(lr){15-17}
& Rad. & MAD & Var. & Diam. & MPD 
& ME & MAE & MSE & Max & Mean
& Beta$^a$ & Factor$^b$ & MIG$^c$ & Dis.$^d$ & Com.$^d$ & Info.$^d$
\\
\midrule
Rad.    & \shade $ 1.00$ & \shade $ 0.08$ & \shade $ 0.25$ & \shade $ 1.00$ & \shade $ 0.33$ & $-0.08$ & $ 0.08$ & $ 0.17$ & $ 0.08$ & $ 0.17$ & $-0.25$ & $ 0.17$ & $ 0.00$ & $ 0.00$ & $ 0.17$ & $-0.08$ \\
MAD     & \shade $ 0.08$ & \shade $ 1.00$ & \shade $ 0.50$ & \shade $ 0.08$ & \shade $ 0.75$ & $ 0.33$ & $ 0.67$ & $ 0.58$ & $ 0.00$ & $ 0.42$ & $-0.50$ & $-0.58$ & $-0.25$ & $ 0.25$ & $ 0.08$ & $-0.00$ \\
Var.    & \shade $ 0.25$ & \shade $ 0.50$ & \shade $ 1.00$ & \shade $ 0.25$ & \shade $ 0.75$ & $ 0.33$ & $ 0.33$ & $ 0.42$ & $-0.17$ & $ 0.25$ & $-0.00$ & $-0.25$ & $-0.08$ & $ 0.58$ & $ 0.42$ & $ 0.33$ \\
Diam.   & \shade $ 1.00$ & \shade $ 0.08$ & \shade $ 0.25$ & \shade $ 1.00$ & \shade $ 0.33$ & $-0.08$ & $ 0.08$ & $ 0.17$ & $ 0.08$ & $ 0.17$ & $-0.25$ & $ 0.17$ & $ 0.00$ & $ 0.00$ & $ 0.17$ & $-0.08$ \\
MPD     & \shade $ 0.33$ & \shade $ 0.75$ & \shade $ 0.75$ & \shade $ 0.33$ & \shade $ 1.00$ & $ 0.25$ & $ 0.42$ & $ 0.50$ & $-0.08$ & $ 0.33$ & $-0.25$ & $-0.33$ & $-0.17$ & $ 0.33$ & $ 0.17$ & $ 0.08$ \\
\midrule
ME      & $-0.08$ & $ 0.33$ & $ 0.33$ & $-0.08$ & $ 0.25$ & \shade $ 1.00$ & \shade $ 0.50$ & \shade $ 0.58$ & \shade $ 0.33$ & \shade $ 0.42$ & $-0.17$ & $-0.25$ & $-0.42$ & $ 0.08$ & $-0.08$ & $ 0.00$ \\
MAE     & $ 0.08$ & $ 0.67$ & $ 0.33$ & $ 0.08$ & $ 0.42$ & \shade $ 0.50$ & \shade $ 1.00$ & \shade $ 0.92$ & \shade $ 0.17$ & \shade $ 0.75$ & $-0.67$ & $-0.58$ & $-0.25$ & $ 0.08$ & $-0.08$ & $-0.17$ \\
MSE     & $ 0.17$ & $ 0.58$ & $ 0.42$ & $ 0.17$ & $ 0.50$ & \shade $ 0.58$ & \shade $ 0.92$ & \shade $ 1.00$ & \shade $ 0.25$ & \shade $ 0.83$ & $-0.58$ & $-0.50$ & $-0.33$ & $ 0.00$ & $-0.17$ & $-0.25$ \\
Max     & $ 0.08$ & $ 0.00$ & $-0.17$ & $ 0.08$ & $-0.08$ & \shade $ 0.33$ & \shade $ 0.17$ & \shade $ 0.25$ & \shade $ 1.00$ & \shade $ 0.42$ & $-0.33$ & $ 0.08$ & $-0.42$ & $-0.25$ & $-0.42$ & $-0.33$ \\
Mean    & $ 0.17$ & $ 0.42$ & $ 0.25$ & $ 0.17$ & $ 0.33$ & \shade $ 0.42$ & \shade $ 0.75$ & \shade $ 0.83$ & \shade $ 0.42$ & \shade $ 1.00$ & $-0.75$ & $-0.50$ & $-0.33$ & $-0.17$ & $-0.33$ & $-0.42$ \\
\midrule
Beta    & $-0.25$ & $-0.50$ & $-0.00$ & $-0.25$ & $-0.25$ & $-0.17$ & $-0.67$ & $-0.58$ & $-0.33$ & $-0.75$ & \shade $ 1.00$ & \shade $ 0.58$ & $ 0.25$ & $ 0.25$ & $ 0.25$ & $ 0.50$ \\
Factor  & $ 0.17$ & $-0.58$ & $-0.25$ & $ 0.17$ & $-0.33$ & $-0.25$ & $-0.58$ & $-0.50$ & $ 0.08$ & $-0.50$ & \shade $ 0.58$ & \shade $ 1.00$ & $-0.17$ & $-0.17$ & $-0.17$ & $ 0.08$ \\
MIG     & $ 0.00$ & $-0.25$ & $-0.08$ & $ 0.00$ & $-0.17$ & $-0.42$ & $-0.25$ & $-0.33$ & $-0.42$ & $-0.33$ & $ 0.25$ & $-0.17$ & \shade $ 1.00$ & $ 0.17$ & $ 0.33$ & $ 0.08$ \\
Dis.    & $ 0.00$ & $ 0.25$ & $ 0.58$ & $ 0.00$ & $ 0.33$ & $ 0.08$ & $ 0.08$ & $ 0.00$ & $-0.25$ & $-0.17$ & $ 0.25$ & $-0.17$ & $ 0.17$ & \shade $ 1.00$ & \shade $ 0.83$ & \shade $ 0.75$ \\
Com.    & $ 0.17$ & $ 0.08$ & $ 0.42$ & $ 0.17$ & $ 0.17$ & $-0.08$ & $-0.08$ & $-0.17$ & $-0.42$ & $-0.33$ & $ 0.25$ & $-0.17$ & $ 0.33$ & \shade $ 0.83$ & \shade $ 1.00$ & \shade $ 0.75$ \\
Info.   & $-0.08$ & $-0.00$ & $ 0.33$ & $-0.08$ & $ 0.08$ & $ 0.00$ & $-0.17$ & $-0.25$ & $-0.33$ & $-0.42$ & $ 0.50$ & $ 0.08$ & $ 0.08$ & \shade $ 0.75$ & \shade $ 0.75$ & \shade $ 1.00$ \\
\bottomrule
\end{tabular}
\end{adjustbox}
\raggedright
\footnotesize{
$^a$ \citep{higgins2017betavae}
$^b$ \citep{kim2018disentangling}
$^c$ \citep{chen2018isolating}
$^d$ \citep{eastwood2018framework}
}
\end{table}

To analyze the relationship between these metrics, we report the Kendall tau distance \citep{kendall1938new, scipy} averaged over experimental settings in \cref{tab:kendalltau}.
The Kendall tau distance is a correlation measure for ordinal data valued in $[-1, 1]$ which counts the number of pairwise disagreements between two ranking lists.
Values close to $1$ indicate strong agreement, and values close to $-1$ indicate strong disagreement.

From \cref{tab:kendalltau} we can observe that even though different metrics derived from the same logical definition may rank imperfect representations differently (see also \cref{fig:rank}), they still have positive correlations with each other, indicating that they measure the same property.
The metrics proposed by \citet{higgins2017betavae} and \citet{kim2018disentangling} have the highest correlations with each other (except for themselves), and we hypothesize that this is because they are both based on the pairwise distance approach.
The DCI disentanglement metric \citep{eastwood2018framework} weakly agrees with the modularity metrics.
However, the DCI informativeness metric \citep{eastwood2018framework} weakly disagrees with the informativeness metrics.
It is possible that this is because of the different regressors (\lstinline{sklearn.ensemble.GradientBoostingClassifier}, \lstinline{sklearn.linear_model.LinearRegression}, and \lstinline{sklearn.linear_model.QuantileRegressor} \citep{sklearn}) used in predicting factors from codes, showing that random seeds and hyperparameters of the metrics may matter more than the models when additional predictors need to be trained to evaluate the learning methods.
This result indicates the advantage of $q_\text{injective}$ over $q_\text{retractable}$.

However, it is important to note the limitations of these experimental results.
Since the representations were learned from data and not fully controlled, it is possible that such results are due to the choices of datasets, learning algorithms, hyperparameters, and optimization errors.
A high rank correlation coefficient between two metrics in this specific setting cannot guarantee that these metrics always measure the same property, or that they rank imperfect representations similarly in other settings.
To gain a deeper understanding of these metrics, it is preferable to analyze their minimizers theoretically (\cref{thm:main}) or test them in a fully controlled environment (\cref{sec:experiments}).


\subsection{Computation time of metrics}

\begin{table}[t]
\centering
\caption{Computation time (seconds) of supervised disentanglement metrics on image datasets}
\label{tab:time}
\begin{adjustbox}{width=\linewidth}
\begin{tabular}{l rrrrr rrrrr rrrrrr}
\toprule
& \multicolumn{5}{c}{Modularity} & \multicolumn{5}{c}{Informativeness} & \multicolumn{4}{c}{Existing metrics}
\\
\cmidrule(lr){2-6}
\cmidrule(lr){7-11}
\cmidrule(lr){12-15}
& \multicolumn{3}{c}{Product approx.} & \multicolumn{2}{c}{Constancy}
& \multicolumn{3}{c}{Retraction approx.} & \multicolumn{2}{c}{Contraction}
& \multicolumn{2}{c}{Pair} & \multicolumn{1}{c}{Info.} & \multicolumn{1}{c}{Regressor}
\\
\cmidrule(lr){2-4}
\cmidrule(lr){5-6}
\cmidrule(lr){7-9}
\cmidrule(lr){10-11}
\cmidrule(lr){12-13}
\cmidrule(lr){14-14}
\cmidrule(lr){15-15}
& Rad. & MAD & Var. & Diam. & MPD 
& ME & MAE & MSE & Max & Mean
& Beta$^a$ & Factor$^b$ & MIG$^c$ & DCI$^d$
\\
\midrule
\cars
& $0.35$ & $2.22$ & $0.01$ & $0.03$ & $0.03$ & $0.14$ & $2.11$ & $0.00$ & $1.09$ & $1.12$ & $4.12$ & $3.77$
& $2.46$ & $896.99$ \\
\dsprites
& $0.47$ & $4.04$ & $0.00$ & $0.03$ & $0.03$ & $0.51$ & $3.84$ & $0.00$ & $1.36$ & $1.37$ & $7.00$ & $6.37$
& $11.51$ & $353.04$ \\
\shapes
& $0.60$ & $6.02$ & $0.00$ & $0.03$ & $0.03$ & $0.66$ & $4.77$ & $0.00$ & $1.67$ & $1.52$ & $4.85$ & $4.65$
& $10.16$ & $169.11$ \\
\mpi
& $0.86$ & $9.30$ & $0.00$ & $0.09$ & $0.09$ & $1.61$ & $5.98$ & $0.00$ & $1.89$ & $1.94$ & $9.54$ & $8.60$
& $21.70$ & $310.38$ \\
\bottomrule
\end{tabular}
\end{adjustbox}
\raggedright
\footnotesize{
$^a$ \citep{higgins2017betavae}
$^b$ \citep{kim2018disentangling}
$^c$ \citep{chen2018isolating}
$^d$ \citep{eastwood2018framework}
}
\end{table}

Finally, we report the computation time of the considered metrics in \cref{tab:time} to support our claim that the proposed metrics are much faster than those that require training additional predictors and hyperparameter tuning.
We can see that, in an extreme case, the calculation of the DCI metrics \citep{eastwood2018framework} using \lstinline{GradientBoostingClassifier} \citep{sklearn} takes around 15 minutes, while other metrics can be calculated within seconds.
This computation time may be acceptable if the metrics are only used in the evaluation phase, but it is not feasible to use them as learning objectives even in derivative-free optimization.


\newpage
\subsection{Factor-wise modularity metrics}

An advantage of the proposed modularity metrics is that we can even evaluate each factor separately, which is impossible for those metrics that entangle modularity and informativeness.
The results were reported in \cref{tab:cars,tab:dsprites,tab:shapes,tab:mpi}.
We found that some learning methods may outperform others on one factor but underperform on others, and different modularity metrics may rank learning methods differently (see also \cref{ssec:rank}).

For example, on {\mpi} (\cref{tab:mpi}), $\beta$-VAE \citep{higgins2017betavae} has the highest scores (radius, MAD, variance, diameter, and MPD) on the horizontal and vertical axis factors, but the lowest scores (radius and diameter) on the object size, camera height, and background color factors.
However, as measured by the MAD and MPD, it still has the highest scores on these factors.
This means that $\beta$-VAE may generally encode the object size, camera height, and background color well compared to other considered methods, but has a small number of outliers.
We believe that such fine-grained evaluation can guide the design of learning objectives, data collection, and further refinement of trained representation learning models.

\begin{table}[t]
\centering
\caption{Factor-wise modularity metrics on \cars}
\label{tab:cars}
\begin{tabular}{l ccccc}
\toprule
& \multicolumn{3}{c}{Product approx.}
& \multicolumn{2}{c}{Constancy}
\\
\cmidrule(lr){2-4}
\cmidrule(lr){5-6}
& Rad. & MAD & Var. & Diam. & MPD
\\
\midrule
\multicolumn{6}{l}{Elevation (4)}\\
\midrule
VAE &
$0.49$ & $0.87$ & $0.97$ & $0.24$ & $0.90$ \\
$\beta$-VAE &
$0.52$ & $0.86$ & $0.97$ & $0.27$ & $0.90$ \\
FactorVAE &
$0.50$ & $0.86$ & $0.97$ & $0.25$ & $0.90$ \\
$\beta$-TCVAE &
$0.50$ & $0.87$ & $0.97$ & $0.25$ & $0.90$ \\
\midrule
\multicolumn{6}{l}{Azimuth (24)}\\
\midrule
VAE &
$0.62$ & $0.91$ & $0.99$ & $0.39$ & $0.94$ \\
$\beta$-VAE &
$0.60$ & $0.91$ & $0.98$ & $0.37$ & $0.93$ \\
FactorVAE &
$0.57$ & $0.90$ & $0.98$ & $0.32$ & $0.93$ \\
$\beta$-TCVAE &
$0.65$ & $0.91$ & $0.99$ & $0.42$ & $0.94$ \\
\midrule
\multicolumn{6}{l}{Object (183)}\\
\midrule
VAE &
$0.87$ & $0.97$ & $1.00$ & $0.75$ & $0.98$ \\
$\beta$-VAE &
$0.84$ & $0.97$ & $1.00$ & $0.71$ & $0.98$ \\
FactorVAE &
$0.85$ & $0.97$ & $1.00$ & $0.72$ & $0.98$ \\
$\beta$-TCVAE &
$0.86$ & $0.97$ & $1.00$ & $0.75$ & $0.98$ \\
\bottomrule
\end{tabular}
\end{table}
\begin{table}[t]
\centering
\caption{Factor-wise modularity metrics on \dsprites}
\label{tab:dsprites}
\begin{tabular}{l rrrrr}
\toprule
& \multicolumn{3}{c}{Product approx.}
& \multicolumn{2}{c}{Constancy}
\\
\cmidrule(lr){2-4}
\cmidrule(lr){5-6}
& Rad. & MAD & Var. & Diam. & MPD
\\
\midrule
\multicolumn{6}{l}{Shape (3)}\\
\midrule
VAE &
$0.74$ & $0.89$ & $0.98$ & $0.55$ & $0.92$ \\
$\beta$-VAE &
$0.70$ & $0.88$ & $0.98$ & $0.50$ & $0.92$ \\
FactorVAE &
$0.73$ & $0.89$ & $0.98$ & $0.53$ & $0.92$ \\
$\beta$-TCVAE &
$0.72$ & $0.88$ & $0.98$ & $0.53$ & $0.92$ \\
\midrule
\multicolumn{6}{l}{Scale (6)}\\
\midrule
VAE &
$0.71$ & $0.90$ & $0.98$ & $0.51$ & $0.93$ \\
$\beta$-VAE &
$0.55$ & $0.88$ & $0.98$ & $0.30$ & $0.92$ \\
FactorVAE &
$0.67$ & $0.90$ & $0.98$ & $0.45$ & $0.93$ \\
$\beta$-TCVAE &
$0.77$ & $0.90$ & $0.98$ & $0.59$ & $0.93$ \\
\midrule
\multicolumn{6}{l}{Orientation (40)}\\
\midrule
VAE &
$0.92$ & $0.97$ & $1.00$ & $0.84$ & $0.98$ \\
$\beta$-VAE &
$0.86$ & $0.98$ & $1.00$ & $0.74$ & $0.98$ \\
FactorVAE &
$0.95$ & $0.98$ & $1.00$ & $0.90$ & $0.99$ \\
$\beta$-TCVAE &
$0.88$ & $0.97$ & $1.00$ & $0.77$ & $0.98$ \\
\midrule
\multicolumn{6}{l}{Position X (32)}\\
\midrule
VAE &
$0.71$ & $0.90$ & $0.98$ & $0.51$ & $0.93$ \\
$\beta$-VAE &
$0.66$ & $0.92$ & $0.99$ & $0.43$ & $0.95$ \\
FactorVAE &
$0.62$ & $0.89$ & $0.98$ & $0.39$ & $0.92$ \\
$\beta$-TCVAE &
$0.67$ & $0.91$ & $0.99$ & $0.44$ & $0.94$ \\
\midrule
\multicolumn{6}{l}{Position Y (32)}\\
\midrule
VAE &
$0.68$ & $0.91$ & $0.99$ & $0.47$ & $0.94$ \\
$\beta$-VAE &
$0.66$ & $0.92$ & $0.99$ & $0.44$ & $0.94$ \\
FactorVAE &
$0.64$ & $0.90$ & $0.98$ & $0.41$ & $0.93$ \\
$\beta$-TCVAE &
$0.67$ & $0.90$ & $0.98$ & $0.45$ & $0.93$ \\
\bottomrule
\end{tabular}
\end{table}
\begin{table}[t]
\centering
\caption{Factor-wise modularity metrics on \shapes}
\label{tab:shapes}
\begin{tabular}{l rrrrr}
\toprule
& \multicolumn{3}{c}{Product approx.}
& \multicolumn{2}{c}{Constancy}
\\
\cmidrule(lr){2-4}
\cmidrule(lr){5-6}
& Rad. & MAD & Var. & Diam. & MPD
\\
\midrule
\multicolumn{6}{l}{Floor hue (10)}\\
\midrule
VAE &
$0.85$ & $0.97$ & $1.00$ & $0.72$ & $0.98$ \\
$\beta$-VAE &
$0.87$ & $0.97$ & $1.00$ & $0.75$ & $0.98$ \\
FactorVAE &
$0.75$ & $0.93$ & $0.99$ & $0.57$ & $0.95$ \\
$\beta$-TCVAE &
$0.99$ & $1.00$ & $1.00$ & $0.97$ & $1.00$ \\
\midrule
\multicolumn{6}{l}{Wall hue (10)}\\
\midrule
VAE &
$0.85$ & $0.97$ & $1.00$ & $0.73$ & $0.98$ \\
$\beta$-VAE &
$0.94$ & $0.99$ & $1.00$ & $0.87$ & $0.99$ \\
FactorVAE &
$0.78$ & $0.94$ & $0.99$ & $0.60$ & $0.96$ \\
$\beta$-TCVAE &
$0.82$ & $0.94$ & $0.99$ & $0.68$ & $0.96$ \\
\midrule
\multicolumn{6}{l}{Object hue (10)}\\
\midrule
VAE &
$0.77$ & $0.95$ & $0.99$ & $0.59$ & $0.96$ \\
$\beta$-VAE &
$0.75$ & $0.92$ & $0.99$ & $0.56$ & $0.95$ \\
FactorVAE &
$0.75$ & $0.93$ & $0.99$ & $0.56$ & $0.95$ \\
$\beta$-TCVAE &
$0.77$ & $0.92$ & $0.99$ & $0.59$ & $0.95$ \\
\midrule
\multicolumn{6}{l}{Scale (8)}\\
\midrule
VAE &
$0.80$ & $0.95$ & $0.99$ & $0.65$ & $0.96$ \\
$\beta$-VAE &
$0.56$ & $0.91$ & $0.98$ & $0.32$ & $0.93$ \\
FactorVAE &
$0.79$ & $0.93$ & $0.99$ & $0.63$ & $0.95$ \\
$\beta$-TCVAE &
$0.80$ & $0.95$ & $1.00$ & $0.63$ & $0.97$ \\
\midrule
\multicolumn{6}{l}{Shape (4)}\\
\midrule
VAE &
$0.59$ & $0.91$ & $0.98$ & $0.35$ & $0.93$ \\
$\beta$-VAE &
$0.62$ & $0.91$ & $0.98$ & $0.38$ & $0.93$ \\
FactorVAE &
$0.77$ & $0.93$ & $0.99$ & $0.60$ & $0.95$ \\
$\beta$-TCVAE &
$0.69$ & $0.92$ & $0.98$ & $0.47$ & $0.94$ \\
\midrule
\multicolumn{6}{l}{Orientation (15)}\\
\midrule
VAE &
$0.95$ & $0.99$ & $1.00$ & $0.91$ & $0.99$ \\
$\beta$-VAE &
$0.94$ & $0.99$ & $1.00$ & $0.89$ & $0.99$ \\
FactorVAE &
$0.89$ & $0.98$ & $1.00$ & $0.80$ & $0.99$ \\
$\beta$-TCVAE &
$0.72$ & $0.91$ & $0.98$ & $0.52$ & $0.94$ \\
\bottomrule
\end{tabular}
\end{table}
\begin{table}[t]
\centering
\caption{Factor-wise modularity metrics on \mpi}
\label{tab:mpi}
\begin{tabular}{l rrrrr}
\toprule
& \multicolumn{3}{c}{Product approx.}
& \multicolumn{2}{c}{Constancy}
\\
\cmidrule(lr){2-4}
\cmidrule(lr){5-6}
& Rad. & MAD & Var. & Diam. & MPD
\\
\midrule
\multicolumn{6}{l}{Object color (6)}\\
\midrule
VAE &
$0.52$ & $0.92$ & $0.99$ & $0.27$ & $0.94$ \\
$\beta$-VAE &
$0.51$ & $0.97$ & $0.99$ & $0.26$ & $0.97$ \\
FactorVAE &
$0.66$ & $0.94$ & $0.99$ & $0.43$ & $0.96$ \\
$\beta$-TCVAE &
$0.57$ & $0.92$ & $0.99$ & $0.33$ & $0.94$ \\
\midrule
\multicolumn{6}{l}{Object shape (6)}\\
\midrule
VAE &
$0.88$ & $0.97$ & $1.00$ & $0.77$ & $0.98$ \\
$\beta$-VAE &
$0.94$ & $1.00$ & $1.00$ & $0.89$ & $1.00$ \\
FactorVAE &
$0.93$ & $0.99$ & $1.00$ & $0.87$ & $0.99$ \\
$\beta$-TCVAE &
$0.95$ & $1.00$ & $1.00$ & $0.91$ & $1.00$ \\
\midrule
\multicolumn{6}{l}{Object size (2)}\\
\midrule
VAE &
$0.70$ & $0.93$ & $0.99$ & $0.49$ & $0.95$ \\
$\beta$-VAE &
$0.63$ & $0.98$ & $1.00$ & $0.40$ & $0.98$ \\
FactorVAE &
$0.68$ & $0.94$ & $0.99$ & $0.46$ & $0.95$ \\
$\beta$-TCVAE &
$0.75$ & $0.96$ & $1.00$ & $0.56$ & $0.97$ \\
\midrule
\multicolumn{6}{l}{Camera height (3)}\\
\midrule
VAE &
$0.48$ & $0.87$ & $0.97$ & $0.23$ & $0.90$ \\
$\beta$-VAE &
$0.34$ & $0.95$ & $0.99$ & $0.11$ & $0.96$ \\
FactorVAE &
$0.58$ & $0.85$ & $0.96$ & $0.34$ & $0.90$ \\
$\beta$-TCVAE &
$0.69$ & $0.91$ & $0.99$ & $0.47$ & $0.94$ \\
\midrule
\multicolumn{6}{l}{Background color (3)}\\
\midrule
VAE &
$0.60$ & $0.92$ & $0.99$ & $0.36$ & $0.94$ \\
$\beta$-VAE &
$0.34$ & $0.96$ & $0.99$ & $0.11$ & $0.97$ \\
FactorVAE &
$0.76$ & $0.94$ & $0.99$ & $0.58$ & $0.96$ \\
$\beta$-TCVAE &
$0.59$ & $0.93$ & $0.99$ & $0.34$ & $0.95$ \\
\midrule
\multicolumn{6}{l}{Horizontal axis (40)}\\
\midrule
VAE &
$0.72$ & $0.93$ & $0.99$ & $0.52$ & $0.95$ \\
$\beta$-VAE &
$0.74$ & $0.99$ & $1.00$ & $0.55$ & $0.99$ \\
FactorVAE &
$0.69$ & $0.92$ & $0.99$ & $0.47$ & $0.94$ \\
$\beta$-TCVAE &
$0.73$ & $0.94$ & $0.99$ & $0.54$ & $0.96$ \\
\midrule
\multicolumn{6}{l}{Vertical axis (40)}\\
\midrule
VAE &
$0.67$ & $0.91$ & $0.99$ & $0.45$ & $0.94$ \\
$\beta$-VAE &
$0.75$ & $0.99$ & $1.00$ & $0.56$ & $0.99$ \\
FactorVAE &
$0.73$ & $0.93$ & $0.99$ & $0.53$ & $0.95$ \\
$\beta$-TCVAE &
$0.60$ & $0.93$ & $0.99$ & $0.36$ & $0.95$ \\
\bottomrule
\end{tabular}
\end{table}

\end{document}